\PassOptionsToPackage{dvipsnames,table}{xcolor}
\documentclass[10pt,letterpaper]{reallab-fairmeta}

\usepackage{amsmath,amsfonts,bm}

\def\eqref#1{equation~\ref{#1}}

\def\1{\bm{1}}

\DeclareMathAlphabet{\mathsfit}{\encodingdefault}{\sfdefault}{m}{sl}
\SetMathAlphabet{\mathsfit}{bold}{\encodingdefault}{\sfdefault}{bx}{n}

\usepackage{iftex}
\ifPDFTeX
  \usepackage[utf8]{inputenc}
  \usepackage[T1]{fontenc}
  \usepackage[nointegrals]{wasysym}
  \DeclareUnicodeCharacter{266B}{\twonotes}
\else
  \PackageError{DRBenchPro arXiv}{Use pdfLaTeX for this manuscript}{Select pdfLaTeX as the compiler.}
\fi

\usepackage{lmodern}
\usepackage{amsmath}
\usepackage{amsfonts}
\usepackage{nicefrac}
\usepackage{graphicx}
\usepackage{multicol}
\usepackage{multirow}
\usepackage{xspace}
\usepackage{wrapfig}
\usepackage{needspace}
\usepackage{float}
\usepackage{enumitem}
\usepackage{xcolor}
\usepackage{colortbl}
\usepackage{caption}
\usepackage{algorithm}
\usepackage{algorithmic}
\usepackage{subcaption}
\usepackage{booktabs}
\usepackage{pifont}
\usepackage{tabularx}
\usepackage{hyperref}
\usepackage{url}

\usepackage{tikz}
\definecolor{exRule}{HTML}{D6E0E4}
\definecolor{exTableStripe}{HTML}{F0F3F6}
\newlength{\ExReportUnit}
\newcommand{\ExArrow}{\ensuremath{\rightarrow}}

\newcolumntype{R}{>{\raggedleft\arraybackslash}X}
\definecolor{lightblue}{RGB}{220,235,250}
\definecolor{lightgray}{gray}{0.91}
\definecolor{drsystemshade}{RGB}{238,237,255}
\definecolor{nativesearchshade}{RGB}{255,245,234}
\definecolor{localframeworkshade}{RGB}{235,251,235}
\definecolor{tablefull}{RGB}{0,128,96}
\colorlet{tablepartial}{blue!70!black}
\definecolor{tablenone}{RGB}{213,94,0}
\newcommand{\fullmark}{\textcolor{tablefull}{\ding{51}}}
\newcommand{\partialmark}{%
  {\textcolor{tablepartial}{\ding{51}}%
   \kern-1.1ex
   \textcolor{tablepartial}{\raisebox{.7ex}{\rotatebox[origin=c]{125}{--}}}}}
\newcommand{\nomark}{\textcolor{tablenone}{\ding{55}}}

\newcommand{\syl}[1]{{\color{black}#1}}

\newcommand{\finishwraptable}{%
  \par
  \ifnum\value{WF@wrappedlines}>1\relax
    \begingroup
    \dimen0=\baselineskip
    \multiply\dimen0 by \numexpr\value{WF@wrappedlines}-1\relax
    \vskip\dimen0
    \endgroup
  \fi
  \WFclear
}

\newenvironment{appendixtable}[2][H]{%
  \begingroup
  \setlength{\intextsep}{10pt}%
  \begin{table}[#1]
  \centering
  \begin{minipage}{#2}
  \centering
  \setlength{\parskip}{0pt}%
  \captionsetup{position=top,skip=6pt}%
}{%
  \end{minipage}
  \end{table}
  \endgroup
}

\definecolor{reallab-red}{RGB}{248,191,199}
\definecolor{reallab-blue}{RGB}{59,161,255}
\definecolor{reallab-green}{RGB}{157,207,127}
\definecolor{reallab-yellow}{RGB}{249,180,13}

\usepackage[most,skins,theorems]{tcolorbox}
\usepackage{listings}
\newtcolorbox{promptbox}[1][]{
  enhanced, breakable,
  colback=gray!1,
  colframe=gray!60,
  coltitle=black,
  boxrule=2pt,
  arc=10pt,
  left=6pt, right=6pt, top=6pt, bottom=6pt,
  title={#1}, fonttitle=\bfseries,
  attach boxed title to top left={yshift*=-3mm},
  boxed title style={colback=gray!10}
}

\tcbset{
  aibox/.style={
    width=\linewidth,
    breakable,
    top=8pt,
    bottom=4pt,
    colback=reallab-red!15,
    colframe=reallab-red,
    colbacktitle=reallab-red!90!black,
    enhanced,
    center,
    fonttitle=\bfseries,
    attach boxed title to top left={yshift=-0.1in,xshift=0.15in},
    boxed title style={boxrule=0pt,colframe=white,},
  }
}
\newtcblisting{AIbox}[2][]{
  aibox,
  title=#2,
  listing only,
  listing engine=listings,
  listing options={
    language={},
    basicstyle=\ttfamily\scriptsize,
    columns=fullflexible,
    keepspaces=true,
    breaklines=true,
    breakatwhitespace=false,
    showstringspaces=false,
    numbers=none,
    frame=none,
    backgroundcolor=\color{reallab-red!15},
    aboveskip=0pt,
    belowskip=0pt
  },
  #1
}

\usepackage{amsthm}
\theoremstyle{plain}

\theoremstyle{definition}

\theoremstyle{remark}

\newcommand{\benchmarkname}{DRBench Pro\xspace}

\setcitestyle{authoryear,round,citesep={;},aysep={,},yysep={;}}

\title{
\benchmarkname: A Knowledge-Centric Benchmark for Professional Deep Research Agents
}

\author[1,*]{Han~Huang}
\author[1,*]{Yuchen~Yan}
\author[1,2]{Xingyu~Wu}
\author[1]{Yijun~Li}
\author[2]{Chao~Deng}
\author[2]{Jie~Liu}
\author[2]{Jin~Ma}
\author[1,\dagger]{Yongliang~Shen}
\affiliation[1]{Zhejiang University}
\affiliation[2]{Tencent}
\contribution[*]{Equal contribution}
\contribution[\dagger]{Corresponding author}

\date{October 13, 2026}
\metadata[Code]{\url{https://github.com/ZJU-REAL/DRBenchPro}}
\metadata[Project Page]{\url{https://zjureal.com/DRBenchPro/}}
\correspondence{\email{\{hanhuang, yanyuchen, syl\}@zju.edu.cn}}

\abstract{

Deep research agents gather information from multiple sources and synthesize it into reports on complex, open-ended questions. Existing benchmarks assess report quality along several dimensions, but lack a dedicated knowledge-centric evaluation of which research directions a report covers and how deeply it explores them. We introduce \textbf{\benchmarkname}, a benchmark that makes structured knowledge coverage the primary evaluation target. Our automated pipeline builds local knowledge trees from Wikipedia and uses sampled subgraphs to generate research queries and evidence-grounded rubrics. Each rubric corresponds to a relation in the tree, preserving the structure of the knowledge being evaluated. Agents receive only the research query. This structure allows us to measure \emph{breadth} across major research directions and \emph{depth} along related knowledge chains, with their geometric mean giving \emph{overall coverage}. The benchmark includes 144 cases across 24 domains and 44,102 rubrics. Evaluating 16 deep research agents reveals different strengths in breadth and depth, with the best-performing agent reaching 57.04\% overall coverage. Rubric judgments agree closely with human annotations, and agent rankings remain stable across the tested metric configurations.
}

\begin{document}
\maketitle

\begin{figure}[b]
\centering
\includegraphics[width=\textwidth]{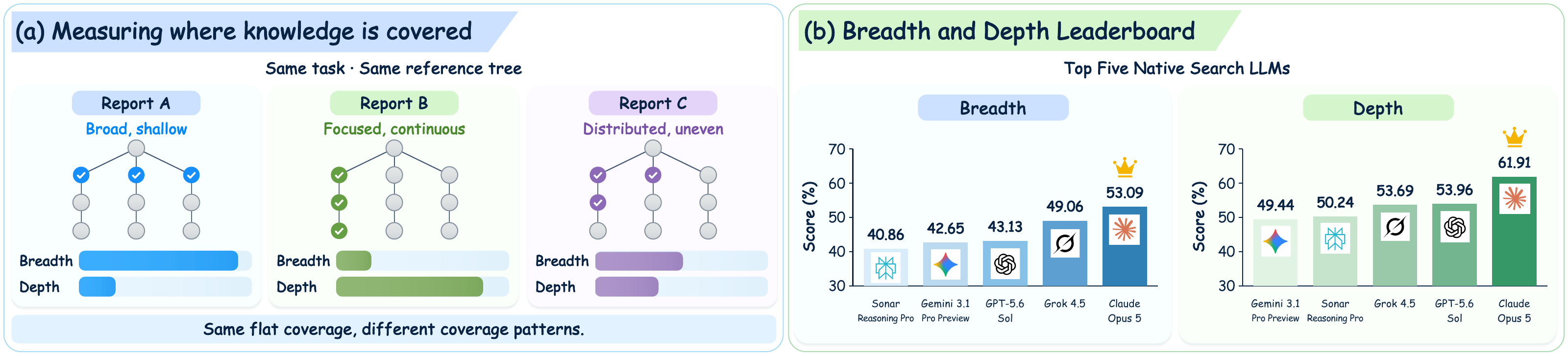}
\caption{\textbf{Structured knowledge coverage in \benchmarkname.} (a) Reports with the same flat coverage can exhibit different coverage patterns over a shared reference tree. (b) Overview of five native search LLMs' performance on \benchmarkname.}
\label{fig:teaser}
\end{figure}

\section{Introduction}
\label{sec:intro}

Deep research agents address complex, open-ended information needs through iterative search, evidence selection, and long-form synthesis. They can investigate a topic across multiple sources and produce research reports that connect relevant facts, explain relationships, and organize findings into a coherent account. Unlike short-form question answering, this task requires an agent to decide which research directions deserve investigation, what evidence to gather, and which related knowledge to pursue further. A comprehensive report therefore depends on both covering the major directions relevant to a question and following chains of related knowledge within those directions.

Evaluating deep research is challenging because open-ended questions admit multiple valid reports, while a single reference answer rarely captures all relevant knowledge. Factuality checks can identify unsupported statements but cannot penalize omitted knowledge~\citep{min2023factscore,wei2024long,song2024veriscore}. Existing benchmarks therefore use expert reports, reference facts, and task-specific rubrics to define expected content and assess factuality, comprehensiveness, citation quality, and writing~\citep{du2026deepresearch,sharma2026researchrubrics,li2026deepresearch}. Recent approaches further reduce manual effort by generating evaluation criteria from Wikipedia, research graphs, or structured task representations~\citep{wang2026wiki,wang2026deepresearcheval,dong2026super,wang2026simple}, providing explicit criteria for assessing report coverage.

However, existing benchmarks lack a dedicated knowledge-centric evaluation of how broadly and deeply reports cover a structured reference space. Knowledge coverage is typically one component of broader assessments that also consider formatting, organization, and writing style\syl{~\citep{du2026deepresearch,wang2026liveresearchbench}}. Expert references and manually authored rubrics further limit expansion across domains, while individually specifying or checking detailed criteria makes rubric coverage costly to scale within each task. \syl{More fundamentally,} even automatically generated criteria lose their relationships under flat aggregation, obscuring whether reports cover more research directions or develop selected directions more deeply. 
Recent tree- and DAG-based approaches preserve structure for supervision, task completion, or evidence traceability~\citep{zhu2026deeprubric,wang2026simple,sun2026hievidr,gao2026dr}. They do not, however, explicitly treat structured knowledge coverage as their primary evaluation target. Specifically, coverage scores use topology-derived weights that are fixed in advance for each task and do not adapt to interactions among knowledge items covered in a report. Yet the value of a knowledge item also depends on whether it connects to related knowledge that is already covered in the report. Complete-path checks establish whether conclusions have supporting evidence, but do not provide a graded measure of how far coverage extends along a knowledge chain when coverage is partial or interrupted by missing intermediate relations. These limitations leave both the breadth of coverage across research directions and the depth of coverage along related knowledge chains insufficiently characterized (Table~\ref{tab:benchmark-comparison}).

\begin{table}[t]
\captionsetup{skip=2pt}

\caption{\textbf{Comparison with representative deep research benchmarks.} \fullmark: fully addressed; \partialmark: partially or indirectly addressed; \nomark: absent.}
\label{tab:benchmark-comparison}
\centering
\scriptsize
\setlength{\tabcolsep}{2.4pt}
\renewcommand{\arraystretch}{1.0}
\begin{tabularx}{\textwidth}{@{}>{\raggedright\arraybackslash}X*{9}{>{\centering\arraybackslash}c}>{\columncolor{lightgray}\centering\arraybackslash}c@{}}
\toprule
\textbf{Comparison Dimension}
& \textbf{DRB}
& \textbf{DRB-II}
& \textbf{ResRub.}
& \textbf{WLC}
& \textbf{DREval}
& \textbf{SuperRes.}
& \textbf{SQA$\rightarrow$DR}
& \textbf{HiEviDR}
& \textbf{DR-Arena}
& \textbf{Ours} \\
\midrule
\textbf{Automated Construction}
& \nomark & \partialmark & \nomark & \partialmark & \fullmark & \partialmark & \fullmark & \partialmark & \fullmark & \fullmark \\
\textbf{Structured Reference Space}
& \nomark & \nomark & \nomark & \nomark & \nomark & \fullmark & \fullmark & \fullmark & \fullmark & \fullmark \\
\textbf{Structure-Aligned Units}
& \nomark & \nomark & \nomark & \nomark & \nomark & \partialmark & \fullmark & \fullmark & \partialmark & \fullmark \\

\textbf{Topology-Derived Breadth--Depth}
& \nomark & \nomark & \nomark & \nomark & \nomark & \partialmark & \partialmark & \partialmark & \partialmark & \fullmark \\
\textbf{Coverage as Primary Target}
& \partialmark & \partialmark & \partialmark & \partialmark & \partialmark & \partialmark & \partialmark & \partialmark & \partialmark & \fullmark \\
\bottomrule
\end{tabularx}
\end{table}

To fill this gap, we introduce \textbf{\benchmarkname}, a knowledge-centric benchmark that treats structured knowledge coverage as its primary evaluation target. Our automated pipeline builds a local entity--relation tree through multi-hop expansion from a Wikipedia root entity and samples coherent subgraphs defining distinct research scopes. Each subgraph is used to generate both a root-centered research query and a rubric set. Every retained edge defines a rubric grounded in Wikipedia evidence, preserving its structural position in the reference tree. This \textit{shared construction aligns task scope with evaluation units} and makes each judgment traceable to its originating relation and structural position.

We further propose topology-aware metrics that aggregate edge-level judgments bottom-up over the reference tree. \emph{Breadth} measures how comprehensively a report covers major research directions, allowing descendant coverage to partially compensate for an omitted relation. \emph{Depth} measures how far reports develop related knowledge chains, with descendant credit conditioned on whether connecting relations are covered. It rewards continuity while retaining discounted credit beyond missing intermediate relations, providing graded scores for incomplete chains. Subtrees are ranked by their realized coverage in each report to emphasize well-developed chains. Their geometric mean yields \emph{overall coverage}. Evaluating 16 deep research agents reveals substantially interleaved breadth and depth rankings, exposing coverage differences that a single aggregate score can conceal (Figure~\ref{fig:teaser}). The strongest evaluated agent, Claude Opus 5, achieves 57.04\% overall coverage, followed by Grok 4.5 at 50.98\%, leaving substantial room for improvement even among frontier agents.
\syl{Rubric judgments agree with human labels at a Macro-F1 of 0.893, and agent rankings remain stable under perturbations of all metric parameters.}

Our main contributions are as follows:

\begin{itemize}[topsep=4pt,itemsep=2pt,leftmargin=*]

\item We introduce \benchmarkname, a knowledge-centric benchmark dedicated to structured knowledge coverage in deep research reports, comprising 144 cases across 24 domains and 44,102 evidence-grounded rubrics\syl{, built by an automated pipeline that jointly generates each research query and its hidden rubrics from the same Wikipedia-derived subgraph while preserving the evidence provenance and structural position of every rubric}.

\item We propose topology-aware breadth and depth metrics, together with overall coverage, to distinguish coverage across research directions from progression along related knowledge chains, supported by human-consistency and stability analyses.

\item We evaluate 16 deep research agents and identify distinct coverage strengths and remaining gaps. We also provide a local research framework with a fixed Wikipedia snapshot, standardized tools, and shared workflow and interaction budgets for controlled evaluation of backbone LLMs.

\end{itemize}

\section{Related Work}
\label{sec:related}

\paragraph{Deep research evaluation.}
Deep research benchmarks evaluate reports through task-specific criteria covering comprehensiveness, factuality, and citation quality~\citep{du2026deepresearch,sharma2026researchrubrics,han2025deer,zhong2026draco,wang2026liveresearchbench,avraham2026dream}. Expected content is drawn from expert reports, Wikipedia, generated criteria, or research graphs, with varying degrees of automation in task and rubric construction~\citep{li2026deepresearch,wang2026wiki,wang2026deepresearcheval,dong2026super}. For survey-based reports, reference-literature recall and statement veracity measure source coverage and factual accuracy~\citep{li2025reportbench}. Complementary evaluations examine information gathering through broad exploration, passage-retrieval completeness, and exhaustive collection~\citep{java2026characterizing,kim2026seekergym,wong2026widesearch,gupta2026deepsearchqa}.

\paragraph{Factuality and content coverage.}
Factuality and coverage address complementary questions: whether reported claims are supported, and whether relevant information is missing. Claim-level verification and citation audits check generated statements against evidence~\citep{min2023factscore,wei2024long,song2024veriscore,gao2023enabling,venkit2025deeptrace}, while joint analysis of retrieval and generation helps locate failures in systems that use retrieved evidence~\citep{ru2024ragchecker}. Coverage evaluation instead identifies omissions by comparing outputs with reference content units and key facts~\citep{nenkova2004evaluating,liu2023revisiting,song2024finesure,lee2024unisumeval}, testing whether they answer reference questions~\citep{deutsch2021towards,scialom2021questeval,zhang2024qapyramid,tan2024proxyqa}, or checking for salient aspects and insights across documents~\citep{samarinas2025beyond,pradeep2025great,laban2024summary}. Automated rubric generation and judgment calibration support the construction and application of these evaluation criteria~\citep{rao2026autorubric,hashemi2024llm}.

\paragraph{Structured construction and evaluation.}
Graphs and trees organize multi-hop questions~\citep{welbl2018constructing,yang2018hotpotqa,ho2020constructing,trivedi2022musique,zhu2024fanoutqa} and synthetic web-search tasks~\citep{li2025websailor,lu2025deepdive,zhang2025infoagent,tao2025webleaper,tao2026webshaper,li2026websailor}. For deep research, they connect queries, evidence, and rubrics to support reinforcement learning supervision and propagation of rubric weights~\citep{zhu2026deeprubric,wang2026simple}. Structured evidence also enables tracing across research stages and depth and width tests for system comparison~\citep{sun2026hievidr,gao2026dr}. Using structure to construct tasks or trace evidence is distinct from retaining it to aggregate report coverage. Our benchmark makes structured knowledge coverage the primary evaluation target: each rubric retains its position in an entity--relation tree during aggregation, yielding separate breadth and depth measures for coverage across research directions and along related knowledge chains. Table~\ref{tab:benchmark-comparison} summarizes these distinctions.

\section{Benchmark Construction and Evaluation Protocol}
\label{sec:method}

\begin{figure}[t]
\centering
\includegraphics[width=\textwidth]{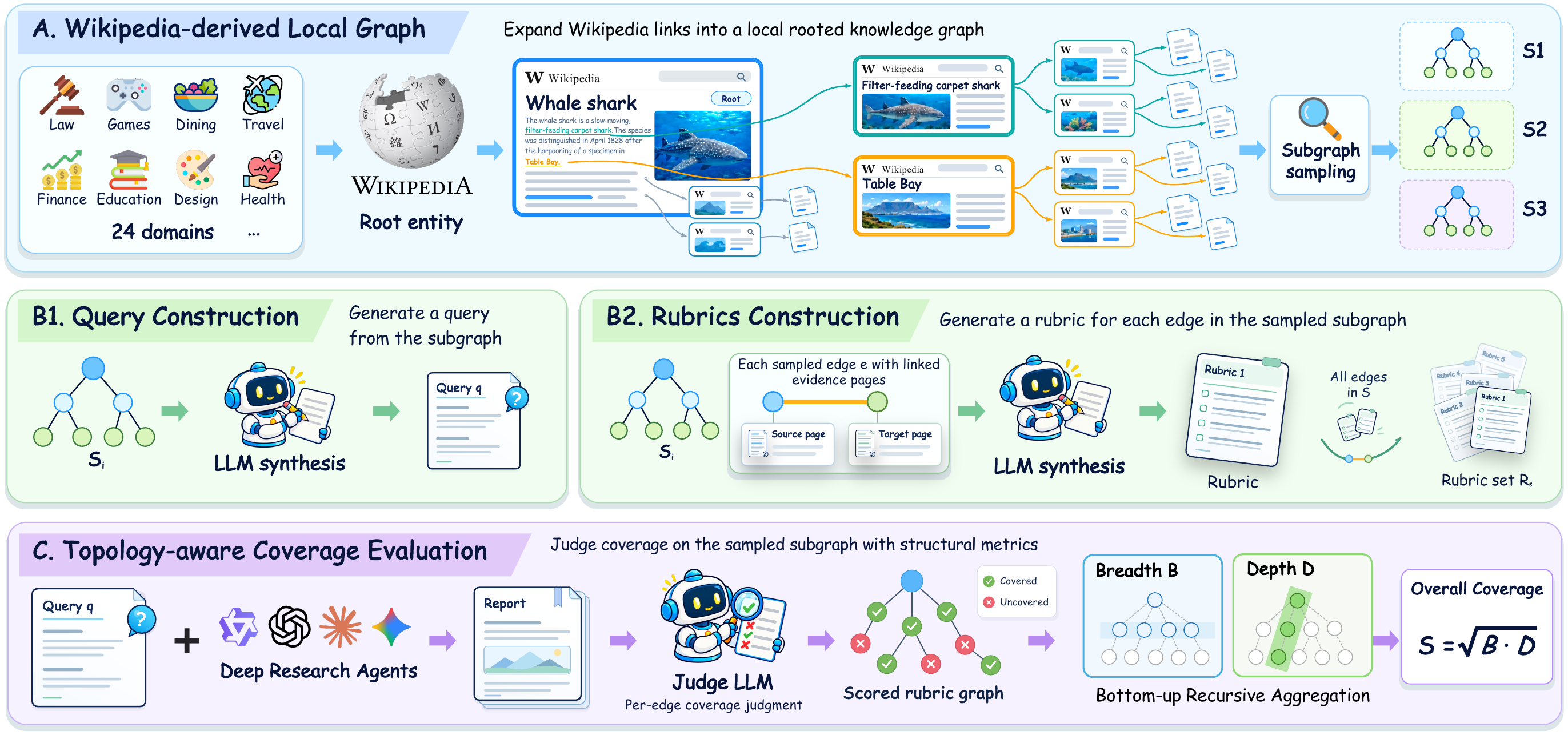}

\caption{\textbf{Overview of \benchmarkname.} \textbf{(A)} We construct a Wikipedia-derived local graph around a selected root entity and sample subgraphs by retaining selected children of the root and their complete descendant subtrees together with the root. \textbf{(B1)} Each sampled subgraph defines a research query. \textbf{(B2)} Each sampled edge is converted into an edge-grounded rubric, forming the hidden rubric set. \textbf{(C)} An evaluated deep research agent answers the query with a report; a judge scores the report against the rubric set, and the edge-level scores are mapped back to the local knowledge tree and aggregated bottom-up into breadth and depth.}
\label{fig:method-overview}
\end{figure}

Measuring knowledge coverage requires a reference that specifies both what a report could cover and how that knowledge is organized. We introduce \textbf{\benchmarkname}, which uses Wikipedia-derived knowledge trees to define research tasks and their evaluation criteria within a shared reference space. The tree structure separates major research directions from chains of related knowledge within them, enabling distinct measures of breadth and depth. Figure~\ref{fig:method-overview} illustrates the construction and evaluation pipeline. The prompts used for benchmark construction and evaluation are provided in Appendix~\ref{app:prompt-templates}.

\subsection{Wikipedia-derived Local Graph Construction}
\label{sec:local-graph}

A reference space for knowledge coverage needs to connect research topics with supporting evidence. Wikipedia provides linked entities whose relationships can be examined in their source context. To span diverse domains, we adopt all 24 topic categories from WebOrganizer~\citep{wettig2025organize} and select two root entities per category. Each root has a sufficiently detailed, reliably parseable article and enough outgoing links to support expansion.

Starting from each root, we expand Wikipedia links breadth-first. Nodes represent entities, and edges represent relationships supported by link anchors and surrounding text. We organize the retained expansion as a rooted tree, with one parent per non-root node and a unique path from the root. This structure gives each relation an identifiable position for subsequent rubric construction and coverage aggregation. Distance from the root is measured by the number of links along this path, and expansion stops at three hops.

To provide diverse candidate research directions while bounding the size of deeper expansions, we retain 50 children of the root and at most five children of each other node. Invalid or duplicate candidates are filtered before Claude Opus 4.8 selects neighbors for their relevance to the current entity and usefulness for research tasks. The resulting 48 local knowledge trees contain approximately 1.5K nodes each on average. Appendix~\ref{app:benchmark-statistics} reports detailed statistics, and Appendix~\ref{app:sample-size-stability} analyzes the choice of two roots per domain. Model-call costs for construction are reported in Appendix~\ref{app:model-call-costs}.

\subsection{Subgraph Sampling}
\label{sec:subgraph-sampling}

A full knowledge tree spans more material than a single research request can reasonably target. We therefore sample smaller subgraphs, each defining the scope of one research case. A branch consists of one child of the root and all its descendants: it represents a research direction, while paths within it capture chains of related knowledge. Each case retains the root and 10 complete branches, preserving the knowledge structure within the selected directions.

To make each scope coherent, Claude Opus 4.8 jointly selects three groups of branches from each tree, using the candidate entities' page summaries, relations to the root, and descendant information. Branches within a group are selected to provide complementary perspectives on a common research task; the three groups do not share root branches. Retaining complete branches preserves the deeper relations needed to evaluate coverage within each direction. This yields three cases per root and 144 cases across the 24 domains.

\subsection{Query and Rubric Construction}
\label{sec:query-rubric}

The research request and its evaluation criteria should refer to the same knowledge scope. We therefore use Claude Opus 4.8 to generate both the query and the rubrics from each sampled subgraph. For query generation, the model receives the root entity and the sampled knowledge scope, and is instructed to frame a natural research request around their broader themes. The prompt discourages listing hidden entities or restating the subgraph as a checklist, so the agent is asked to investigate a topic rather than reproduce an explicit answer outline.

To make the expected coverage traceable to evidence, we create one rubric per retained relation edge. Each rubric asks whether the report covers the relationship between its source and target entities. The generator uses only the supplied evidence: page summaries, the anchor sentence, relevant paragraphs, and available backlink context from the two pages. This ties each scoring requirement to documented content rather than an unsupported model-generated expectation.

Each rubric retains its edge's position in the reference tree, allowing its judgment to be aggregated with those of related rubrics. We also record its hop distance as the distance from the root to the edge's target node for structural diagnostics. These annotations are kept outside the rubric text. Evaluated agents receive only the query; the reference tree and rubrics remain hidden.

\subsection{Topology-aware Knowledge Coverage Evaluation}
\label{sec:coverage-evaluation}

Counting satisfied rubrics alone can give similar scores to reports with different coverage patterns. We therefore retain the reference tree when aggregating judgments, distinguishing coverage across research directions from coverage along related knowledge chains. An LLM judge assigns each hidden rubric one of three verdicts: \emph{Satisfied}, \emph{Partially Satisfied}, or \emph{Not Satisfied}. For coverage metric computation, we map \emph{Satisfied} to $1$ (covered) and both \emph{Partially Satisfied} and \emph{Not Satisfied} to $0$ (uncovered). These binary values are then aggregated over the reference tree.

\paragraph{Breadth.}
Breadth measures how comprehensively a report covers the selected research directions. We give these directions equal weight so that a branch with more rubrics does not dominate the score. Within a branch, deeper relations can demonstrate coverage of a direction even when a shallower relation is missing. We preserve each relation’s direct score as a weighted component and use the mean breadth of its child subtrees to account for descendant coverage. Averaging the resulting branch scores gives case-level breadth.

\paragraph{Depth.}
Depth measures coverage along related knowledge chains in the reference tree. A report can cover one chain deeply without covering every branch equally, so we give greater weight to child subtrees with higher depth scores while retaining smaller contributions from the others. Each relation contributes its own score and regulates how much descendant coverage is credited: an uncovered intermediate relation reduces that credit without discarding it. This rewards continuity along knowledge chains. We apply the same rank-based weighting across the selected research directions to obtain case-level depth. Appendix~\ref{app:chain-textual-organization} analyzes the textual organization of supporting evidence for consecutive satisfied rubrics.

\paragraph{Overall coverage.}
We combine case-level breadth $B_c$ and depth $D_c$ through their geometric mean,
\begin{equation}
S_c=\sqrt{B_cD_c},
\label{eq:overall-coverage-main}
\end{equation}
so weak coverage in either dimension lowers the combined score. Each of the three metrics is then averaged separately across the 144 cases.

Both metrics are computed from the leaves toward the root, starting with the direct rubric scores and recursively combining them with descendant coverage. Appendix~\ref{app:coverage-metric-details} provides the complete procedure in Algorithm~\ref{alg:coverage}, together with the equations, parameter definitions, and boundary conventions.

\section{Evaluation}
\label{sec:experiments}

\subsection{Evaluation Setup}
\label{sec:experimental-setup}

\paragraph{Evaluation methods.}
We evaluate 16 deep research agents and group them into three categories according to how search and research orchestration are provided.

\paragraph{(1) Deep Research Systems.}
These are end-to-end research agents explicitly designed for autonomous, long-horizon investigation. They natively combine planning, iterative evidence gathering, and long-form synthesis. We evaluate o3 Deep Research, Perplexity Sonar Deep Research, and Qwen Deep Research.

\paragraph{(2) Native Search LLMs.}
These models integrate provider-native web search directly into generation without relying on a separate Deep Research product workflow. We evaluate Grok 4.5, Claude Opus 5, GPT-5.6 Sol, Gemini 3.1 Pro Preview, Sonar Reasoning Pro, and Qwen3.8 Max with native web search enabled.

\paragraph{(3) Backbone LLMs under the local research framework.}
As a controlled and reproducible evaluation setting, we implement the local research framework backed by the fixed English Wikipedia dump \texttt{enwiki-20260701-v1}. The framework exposes \textsc{Search} and \textsc{Open} operations over the same snapshot, disables provider-native web search, and keeps the research workflow and budget fixed across backbone LLMs. Its purpose is to provide a common search service into which different backbone LLMs can be integrated, rather than to improve their benchmark scores. This design removes temporal drift in the retrieval corpus, reduces variation from external search infrastructure, and retains auditable evidence records. We evaluate Gemma 4 31B, DeepSeek V4 Pro, DeepSeek V4 Flash, Qwen3.8 Max, Qwen3.7 Plus, GLM 5.2, and MiniMax M3 as backbone LLMs. The execution protocol and formal configuration of the local research framework are detailed in Appendix~\ref{app:local-research-framework}. Appendix~\ref{app:odr-supplementary-evaluation} evaluates LangChain Open Deep Research with DeepSeek V4 Flash on a small sample.

\paragraph{Implementation details.}
The evaluated reports were collected during the periods listed in Appendix~\ref{app:evaluation-timeframes}. We use Gemini 2.5 Pro as the judge for the formal evaluation. It assigns each rubric one of three verdicts: \emph{Satisfied}, \emph{Partially Satisfied}, or \emph{Not Satisfied}. We retain these original judgments and convert them into binary rubric scores of $0$ or $1$ for the metrics in Section~\ref{sec:coverage-evaluation}. Appendix~\ref{app:model-call-costs} reports the recorded model-call costs for report generation and judge evaluation.

\subsection{Evaluation Results}
\label{sec:results}

\begin{table}[t]

\caption{Main results on \benchmarkname (\%). \textbf{Bold} and \underline{underlined} values indicate the best and second-best scores across all agents, respectively, based on unrounded values.}
\label{tab:main-results}
\centering
\small
\setlength{\tabcolsep}{6pt}
\begin{tabularx}{\textwidth}{>{\raggedright\arraybackslash}Xccc}
\toprule
\textbf{Deep Research Agent} & \textbf{Overall Coverage} & \textbf{Breadth} & \textbf{Depth} \\
\midrule
\rowcolor{drsystemshade}\multicolumn{4}{c}{\textbf{(1) Deep Research Systems}} \\
o3 Deep Research                  & 49.45 & 45.77 & \underline{54.32} \\
Perplexity Sonar Deep Research    & 46.06 & 40.00 & 54.08 \\
Qwen Deep Research                & 35.16 & 29.33 & 43.37 \\
\midrule
\rowcolor{nativesearchshade}\multicolumn{4}{c}{\textbf{(2) Native Search LLMs}} \\
Claude Opus 5                     & \textbf{57.04} & \textbf{53.09} & \textbf{61.91} \\
Grok 4.5                          & \underline{50.98} & \underline{49.06} & 53.69 \\
GPT-5.6 Sol                       & 47.80 & 43.13 & 53.96 \\
Gemini 3.1 Pro Preview            & 45.50 & 42.65 & 49.44 \\
Sonar Reasoning Pro               & 44.76 & 40.86 & 50.24 \\
Qwen3.8 Max                       & 36.53 & 31.08 & 44.09 \\
\midrule
\rowcolor{localframeworkshade}\multicolumn{4}{c}{\textbf{(3) Backbone LLMs under the local research framework}} \\
Qwen3.8 Max                       & 47.94 & 46.62 & 50.37 \\
MiniMax M3                        & 45.34 & 44.34 & 47.55 \\
GLM 5.2                           & 45.34 & 44.61 & 47.09 \\
DeepSeek V4 Flash                 & 42.63 & 41.42 & 45.04 \\
DeepSeek V4 Pro                   & 39.22 & 36.96 & 43.03 \\
Qwen3.7 Plus                      & 36.10 & 35.37 & 37.94 \\
Gemma 4 31B                       & 32.00 & 30.43 & 34.82 \\
\bottomrule
\end{tabularx}
\end{table}

\begin{figure}[t]
\centering
\includegraphics[width=\textwidth]{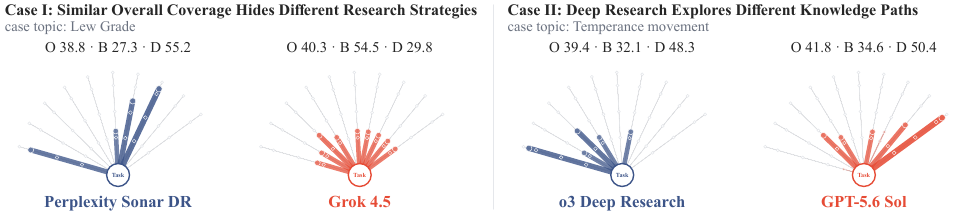}
\caption{\textbf{Case-level coverage profiles.} O/B/D: overall coverage/breadth/depth (\%). Gray rays mark shared directions; colored ray width/opacity and length encode branch breadth and depth, respectively.}
\label{fig:case-study-main}
\end{figure}

\paragraph{Overall performance.}

Table~\ref{tab:main-results} shows that Claude Opus 5 achieves the highest overall coverage (57.04\%) and ranks first in both breadth (53.09\%) and depth (61.91\%), yielding strong performance in both dimensions. Grok 4.5 and o3 Deep Research follow, with relatively stronger breadth and depth, respectively. High overall coverage therefore requires both comprehensive coverage across major directions and progress along related knowledge chains. Appendix~\ref{app:additional-prior-benchmarks} lists selected results from prior benchmarks for context.

\paragraph{Interleaved breadth--depth profiles.}

The agent orderings differ substantially between breadth and depth, revealing distinct research strengths and showing that research workflows can produce markedly different coverage profiles. The clearest example is Qwen3.8 Max: under the local research framework, its breadth and depth scores are 15.54 and 6.28 percentage points higher, respectively, than under its Native Search configuration, with the difference concentrated primarily in breadth. The two configurations of the same-named model therefore exhibit different knowledge-exploration patterns: the local research framework configuration covers major research directions more comprehensively, whereas the depth difference is comparatively limited. A similar contrast appears within the Perplexity product line: Perplexity Sonar Deep Research attains higher depth but slightly lower breadth than Sonar Reasoning Pro. Together, these results show that breadth and depth expose agent specializations and research-strategy differences that overall coverage alone cannot reveal.

Figure~\ref{fig:case-study-main} presents two representative cases. For Lew Grade, similar overall coverage masks a breadth--depth trade-off. For Temperance, agents can deepen their investigations along different knowledge paths. Appendix~\ref{app:case-studies} presents the case studies in greater detail. Appendix~\ref{app:example-showcase} provides a full report with rubric judgments and highlighted evidence.

\paragraph{Coverage under the local research framework.}

Under the local research framework, Qwen3.8 Max achieves the highest overall coverage (47.94\%). Several backbone LLMs remain competitive under a limited budget. Among the leading backbones, breadth scores are already relatively close, whereas differences in depth further separate their overall coverage. Because these models share the same knowledge source, workflow, tool-use constraints, and budget, the remaining differences reflect backbone-dependent behavior in task decomposition, tool use, evidence selection, and report synthesis. Appendix~\ref{app:trajectory-local} analyzes the agents' research trajectories. Appendices~\ref{app:target-length-sensitivity} and~\ref{app:retrieval-report-retention} analyze length sensitivity and retrieval-to-report knowledge retention.

\subsection{Coverage Profiles across Domains}
\label{sec:domain-coverage-profiles}

\begin{figure}[t]
\centering
\includegraphics[width=\textwidth]{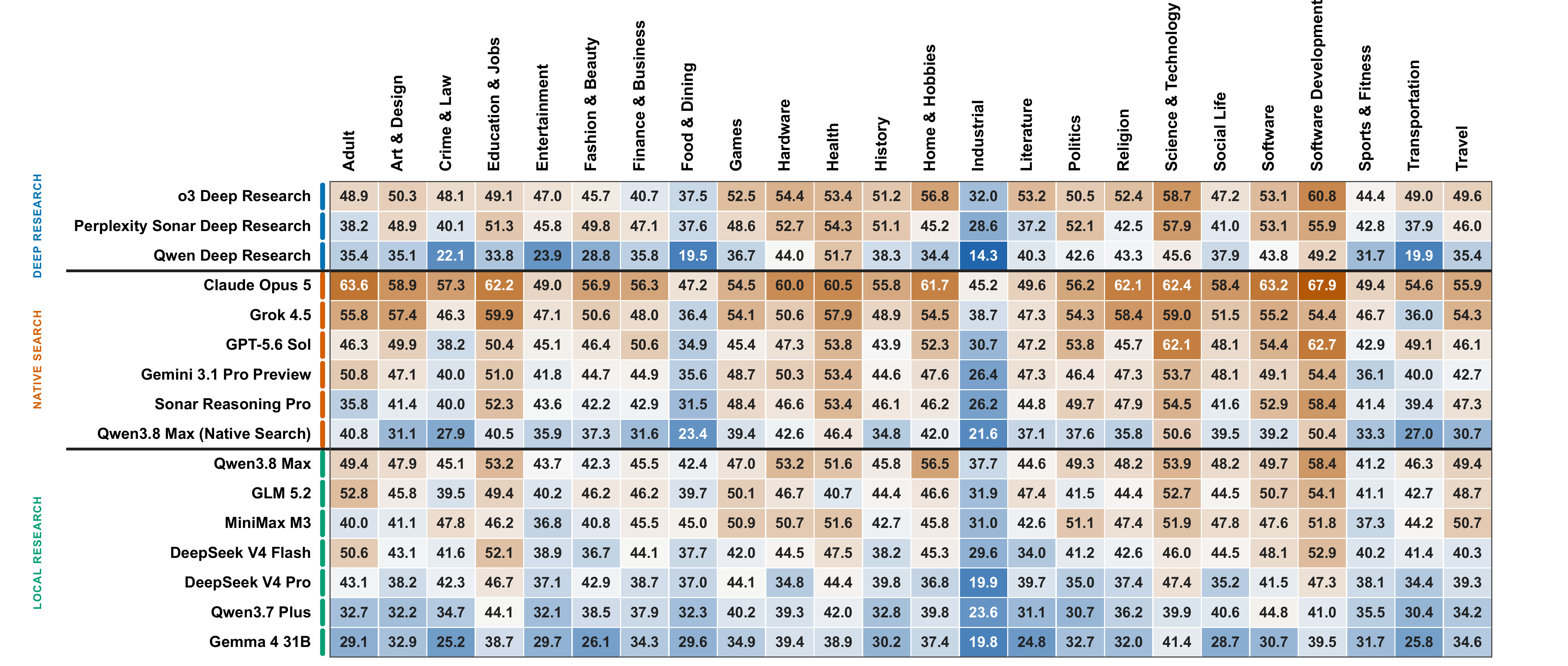}

\caption{\textbf{Overall coverage across domains.} Values are percentages. Orange indicates higher scores, while blue indicates lower scores.}
\label{fig:domain-overall-coverage}
\end{figure}

Figure~\ref{fig:domain-overall-coverage} shows overall coverage across 24 domains, with each agent--domain score averaged over six cases. Across agents, coverage is highest in Software Development (53.69\%), Science \& Technology (52.36\%), and Health (50.10\%), and lowest in Industrial (28.58\%) and Food \& Dining (35.45\%).

Claude Opus 5 leads in 23 domains, and o3 Deep Research leads in Literature. Appendix~\ref{app:domain-coverage-profiles} provides the corresponding breadth and depth profiles.

\Needspace{5\baselineskip}
\subsection{Research Task Quality and Leakage}
\label{sec:task-quality-leakage}

Because each research query is generated from a sampled subgraph, we manually review all 144 queries using three criteria: research value, public answerability, and alignment with their assigned domains (Table~\ref{tab:task-quality-main}). All are rated Good on the first two criteria; 141 are Good and three Medium on domain match.

The three Medium cases arise from natural cross-domain overlap, as subgraph sampling produces topic-specific research scopes that are not necessarily confined to a single domain; the queries nevertheless remain coherent, publicly answerable, and sufficiently aligned with their assigned domains for benchmark evaluation.

We also check whether queries reveal the hidden entity information used for evaluation. For each of 24 sampled cases (one per domain), we compare its query with every non-root entity title in its subgraph, yielding 7,385 query--entity comparisons, and manually review literal matches. The 34 generic-term overlaps arise because some subgraph titles, or components of those titles, are ordinary words or phrases that naturally occur in research queries and therefore do not unambiguously identify the corresponding entities. These overlaps are excluded from the leakage count. 8 entity-title mentions are confirmed as leaks, with none at the first hop (Table~\ref{tab:task-leakage-main}). Therefore, the low leakage rate is unlikely to materially affect subsequent report generation or benchmark evaluation.

\begingroup
\setlength{\intextsep}{10pt}
\begin{table}[t]
\centering
\small
\setlength{\parskip}{0pt}
\captionsetup{position=top,skip=6pt}
\begin{minipage}[t]{0.44\textwidth}
\centering
\caption{\textbf{Research-task quality.} 
The three criteria are Research Value, Public Answerability, and Query--Domain Match, as defined in Appendix~\ref{app:query-quality}.
}
\label{tab:task-quality-main}
\setlength{\tabcolsep}{3.5pt}
\begin{tabular}{@{}lccc@{}}
\toprule
Criterion & Poor & Medium & Good \\
\midrule
Research value & 0 & 0 & 144 \\
Public answerability & 0 & 0 & 144 \\
Query--domain match & 0 & 3 & 141 \\
\bottomrule
\end{tabular}
\end{minipage}\hfill
\begin{minipage}[t]{0.55\textwidth}
\centering
\caption{\textbf{Lexical leakage by hop distance.}}
\label{tab:task-leakage-main}
\setlength{\tabcolsep}{2.7pt}
\begin{tabular}{@{}lccc@{}}
\toprule
\shortstack{Hop\\distance} & \shortstack{Inspected\\entities} & \shortstack{Generic-term\\matches} & \shortstack{Confirmed leaks,\\$n$ (\%)} \\
\midrule
$h(v)=1$ & 240 & 3 & 0 (0\%) \\
$h(v)=2$ & 1,200 & 9 & 5 (0.417\%) \\
$h(v)=3$ & 5,945 & 22 & 3 (0.050\%) \\
\midrule
Total & 7,385 & 34 & 8 (0.108\%) \\
\bottomrule
\end{tabular}
\end{minipage}
\end{table}
\endgroup

\FloatBarrier

\Needspace{29\baselineskip}
\subsection{Human Consistency}
\label{sec:human-judge-consistency}

\begingroup
\paragraph{Validating the design of evaluation metrics.}
At the report level, we compare our final metrics with independent expert assessments of 72 reports from six agents across 12 cases.

\newsavebox{\metricvalidationtablebox}
\sbox{\metricvalidationtablebox}{%
\begin{tabular}{@{}ll@{\hspace{6pt}}c@{\hspace{6pt}}c@{}}
\toprule
Dimension & Scoring scheme & $\tau_b\uparrow$ & $\rho_s\uparrow$ \\
\midrule
Overall & \textbf{Final metrics} & \textbf{0.751} & \textbf{0.833} \\
 & Level-balanced coverage & 0.615 & 0.718 \\
 & Flat coverage & 0.508 & 0.597 \\
\midrule
Breadth & \textbf{Final metrics} & \textbf{0.696} & \textbf{0.777} \\
 & Branch coverage & 0.533 & 0.592 \\
 & w/o descendant fusion & 0.643 & 0.714 \\
\midrule
Depth & \textbf{Final metrics} & \textbf{0.727} & \textbf{0.813} \\
 & Within-path pair coverage & 0.577 & 0.664 \\
 & w/o continuity discount & 0.668 & 0.761 \\
\bottomrule
\end{tabular}%
}
\par\addvspace{\baselineskip}
\begin{wraptable}{r}{\wd\metricvalidationtablebox}
\centering
\caption{Agreement with expert assessments. Kendall's $\tau_b$ and Spearman's $\rho_s$ are computed within each case and averaged across cases.}
\label{tab:metric-expert-agreement}
\usebox{\metricvalidationtablebox}
\end{wraptable}

\strut Three experts independently rate each report's breadth, depth, and overall quality on a 1--10 scale, and their mean ratings serve as the human reference (Appendix~\ref{app:report-level-annotation}). Holding the evaluated reports and their corresponding rubric judgments fixed, we compare these metrics with flat and structural coverage baselines and conduct component ablations to examine the contributions of key design choices in our final metrics.

As shown in Table~\ref{tab:metric-expert-agreement}, our final metrics exhibit strong agreement with expert assessments, with mean Kendall correlations of 0.696--0.751 and Spearman correlations of 0.777--0.833 across the three dimensions, supporting their effectiveness in reflecting expert judgments. They also achieve the highest agreement among the evaluated baselines and ablations under both statistics. Ablations further support the contributions of key design components in our final metrics: local-descendant coverage fusion and chain continuity reward, with additional baseline and ablation results and complete scoring definitions provided in Appendix~\ref{app:metric-validation}.
\par
\endgroup
\par\addvspace{\baselineskip}

\newsavebox{\humanjudgetablebox}
\sbox{\humanjudgetablebox}{%
\small
\begin{tabular}{@{}lcc@{}}
\toprule
\multicolumn{1}{c}{\multirow{2}{*}{Judge}} & \multicolumn{2}{c}{Macro-F1} \\
\cmidrule(lr){2-3}
 & Binary & Ternary \\
\midrule
Gemini 2.5 Pro  & \textbf{0.893} & \textbf{0.806} \\
Qwen 3.8 Max    & 0.840 & 0.718 \\
DeepSeek V4 Pro & 0.865 & 0.630 \\
\bottomrule
\end{tabular}%
}
\Needspace{11\baselineskip}
\begin{wraptable}{r}{\wd\humanjudgetablebox}
\centering
\small
\caption{\textbf{Human--LLM agreement.}}
\label{tab:human-judge-macro-f1}
\usebox{\humanjudgetablebox}
\par\addvspace{\baselineskip}
\end{wraptable}

At the rubric level, we compare three LLM judges with human-reviewed labels on 3,695 rubric judgments from 12 reports, each representing a different domain. Gemini 2.5 Pro, Qwen 3.8 Max, and DeepSeek V4 Pro assign \emph{Satisfied}, \emph{Partially Satisfied}, or \emph{Not Satisfied} verdicts. For binary agreement, we map \emph{Satisfied} to $1$ (covered) and the other two verdicts to $0$ (uncovered) for both human and judge labels. We compute Macro-F1 separately for each report and average it across the 12 reports. Table~\ref{tab:human-judge-macro-f1} presents the results for both binary and ternary scoring. Appendix~\ref{app:rubric-level-annotation} details the human annotation protocol.

\paragraph{Binary versus ternary scoring.}
In our benchmark, binary scoring yields higher mean Macro-F1 than ternary scoring for all three judges. This finding aligns with the trend reported by ResearchRubrics~\citep{sharma2026researchrubrics}. We therefore adopt binary rubric scores for metric computation in our benchmark.

\paragraph{Judge-model comparison.}
Under binary scoring, Gemini 2.5 Pro has the highest agreement with the human-reviewed labels among the three judges. We therefore retain Gemini 2.5 Pro as the judge and convert its ternary outputs to binary scores for the final evaluation.

\par\addvspace{2\baselineskip}
\WFclear

\Needspace{24\baselineskip}
\subsection{Metric Sensitivity}
\label{sec:metric-sensitivity}

\newsavebox{\metricsensitivitytablebox}
\sbox{\metricsensitivitytablebox}{%
\footnotesize
\setlength{\tabcolsep}{2pt}
\begin{tabular}{@{}lccccccc@{}}
\toprule
\multicolumn{5}{c}{Parameter configuration} & \multicolumn{3}{c}{Ranking stability} \\
\cmidrule(lr){1-5}\cmidrule(lr){6-8}
Variant & $\alpha$ & $\eta$ & $\rho$ & $\beta$ & Spearman $r_s$ & Exact & Practical \\
\midrule
Default        & 0.25 & 0.75 & 0.50 & 0.60 & 1.000/1.000 & 0/0 & 0/0 \\
$\alpha$-Low   & \textbf{0.10} & 0.75 & 0.50 & 0.60 & 1.000/0.979 & 0/5 & 0/2 \\
$\alpha$-High  & \textbf{0.40} & 0.75 & 0.50 & 0.60 & 1.000/0.976 & 0/6 & 0/1 \\
$\eta$-Low     & 0.25 & \textbf{0.50} & 0.50 & 0.60 & 1.000/0.997 & 0/1 & 0/0 \\
$\eta$-High    & 0.25 & \textbf{0.90} & 0.50 & 0.60 & 1.000/1.000 & 0/0 & 0/0 \\
$\rho$-Low     & 0.25 & 0.75 & \textbf{0.40} & 0.60 & 1.000/0.994 & 0/2 & 0/0 \\
$\rho$-High    & 0.25 & 0.75 & \textbf{0.60} & 0.60 & 1.000/0.985 & 0/4 & 0/0 \\
$\beta$-Low    & 0.25 & 0.75 & 0.50 & \textbf{0.50} & 1.000/1.000 & 0/0 & 0/0 \\
$\beta$-High   & 0.25 & 0.75 & 0.50 & \textbf{0.75} & 0.997/1.000 & 1/0 & 0/0 \\
\bottomrule
\end{tabular}%
}
\begin{wraptable}{r}{\wd\metricsensitivitytablebox}
\centering
\caption{\textbf{Sensitivity of agent rankings to metric hyperparameters.} Spearman correlations compare each ranking with the default; exact and practical reversals are counted over 120 agent pairs. All paired values are breadth/depth ($B/D$), with breadth first.}
\label{tab:metric-sensitivity-main}
\usebox{\metricsensitivitytablebox}
\end{wraptable}

We test ranking sensitivity by varying aggregation parameter and recomputing each deep research agent's mean breadth and depth over all 144 cases. Table~\ref{tab:metric-sensitivity-main} compares the resulting rankings of 16 agents with the default configuration. Exact reversals count agent pairs whose ordering changes; practical reversals additionally require the absolute score difference between the two agents to exceed 0.5 percentage points under both the default configuration and the tested parameter configuration.

Rankings remain stable across the tested perturbations: Spearman correlations are at least 0.997 for breadth and 0.976 for depth. Depth is most sensitive to $\alpha$: raising it from 0.25 to 0.40 causes one practical reversal, whereas lowering it to 0.10 causes two. This reflects $\alpha$'s balance between a relation's direct coverage and its descendants' coverage. A large $\alpha$ suppresses descendant evidence during bottom-up propagation; a small one gives little independent credit to a covered non-leaf relation when its descendants are uncovered. Neither extreme matches our intended depth definition. All other perturbations cause no practical reversals, supporting stable agent comparisons within the tested ranges.

\subsection{Construction-Model Sensitivity}
\label{sec:construction-model-sensitivity}
\begingroup
\setlength{\intextsep}{12pt plus 2pt minus 2pt}
While several related benchmarks, such as DeepResearchEval~\citep{wang2026deepresearcheval} and DeepResearch Bench II~\citep{li2026deepresearch}, assess robustness across judge models, we additionally examine potential sensitivity to the model used to construct the tasks and rubrics. To this end, we reconstructed reference graphs, queries, and rubrics with Kimi K2.6. On 10 sampled cases, we compare Claude Opus 5 with three Native Search LLMs that performed relatively close to Claude in the original evaluation. We compare the performance of the same four models on the Kimi-constructed benchmark and the original benchmark constructed by Claude Opus 4.8, using corresponding roots.

\begin{table}[H]
\centering
\caption{\textbf{Sensitivity to the benchmark construction model.} Ranks are computed among the four models within each case and averaged across cases. Positive gaps favor Claude.}
\label{tab:construction-model-sensitivity}
\small
\setlength{\tabcolsep}{4pt}
\begin{tabular}{@{}lcccccc@{}}
\toprule
\mbox{\textbf{Evaluated model}} & \multicolumn{2}{c}{\textbf{Mean Overall (\%)}} & \multicolumn{2}{c}{\textbf{Mean rank}} & \multicolumn{2}{c}{\textbf{Claude's standardized gap}} \\
\cmidrule(lr){2-3}\cmidrule(lr){4-5}\cmidrule(lr){6-7}
 & Original & Kimi-built & Original & Kimi-built & Original & Kimi-built \\
\midrule
Claude Opus 5 & 56.84 & 40.20 & 1.2 & 1.8 & --- & --- \\
Grok 4.5 & 51.25 & 37.63 & 2.5 & 1.9 & +0.869 & +0.336 \\
GPT-5.6 Sol & 46.65 & 34.21 & 3.0 & 2.7 & +1.546 & +0.807 \\
Sonar Reasoning Pro & 44.09 & 25.84 & 3.3 & 3.6 & +1.944 & +1.804 \\
\bottomrule
\end{tabular}
\end{table}

To quantify Claude's relative advantage over each opponent, for each benchmark version, we compute the mean standardized gap across the 10 cases as \(\mathrm{Gap}_m=\frac{1}{10}\sum_{i=1}^{10}\frac{S_{i,C}-S_{i,m}}{\sigma_i}\), where \(S_{i,C}\) and \(S_{i,m}\) are the Overall scores of Claude and opponent \(m\) on case \(i\), and \(\sigma_i\) is the population standard deviation of the four model scores on that case.

As shown in Table~\ref{tab:construction-model-sensitivity}, under Kimi-based construction, Claude retains the highest mean overall score and appreciable positive standardized gaps against all three opponents in the sampled cases, although its relative advantage narrows. Taken together, these findings suggest that Claude's observed advantage can persist when the benchmark is constructed by a different model family, although the choice of construction model may still play a role.
\par
\endgroup
\FloatBarrier

\section{Conclusion}
\label{sec:conclusion}

We introduce \textbf{\benchmarkname}, a knowledge-centric benchmark for professional deep research agents. From each sampled multi-hop Wikipedia-derived entity subgraph, it jointly generates a research query and traceable edge-grounded rubrics, giving the case, knowledge scope, and evaluation criteria a shared structural foundation. Topology-aware aggregation captures breadth across major research directions and depth along related knowledge chains, with their geometric mean yielding overall coverage. Evaluation across 16 deep research agents reveals interleaved breadth--depth profiles; with the knowledge source, workflow, tool interface, and interaction budget fixed, the local research framework further distinguishes the coverage behavior of different backbone LLMs. Overall, \textbf{\benchmarkname} uses graph topology in both benchmark construction and metric formulation, providing a scalable, structure-aware foundation for evaluating knowledge coverage in deep research reports.

\bibliography{main,custom}
\bibliographystyle{main}

\clearpage
\appendix

\renewcommand{\thetable}{\arabic{table}}
\renewcommand{\thefigure}{\arabic{figure}}

\Needspace{48\baselineskip}
\section{Benchmark Statistics}
\label{app:benchmark-statistics}

For the structural statistics below, $h(v)$ denotes the number of edges from the root to node $v$. For a rubric $r_e$ associated with an edge $e=(u,v)$, its hop distance is $\delta(r_e)=h(v)$.

Our \benchmarkname benchmark is domain-balanced and contains 24 topical domains, 48 root entities, and 144 research cases. Each domain contributes two root entities, and each root yields three case subgraphs, giving six cases per domain. Table~\ref{tab:benchmark-root-entities} lists the two root entities selected for each domain. As summarized in Table~\ref{tab:benchmark-statistics}, each of the three case subgraphs sampled from a root entity selects 10 of that root's 50 children and retains their complete branches. The three sets of selected children are mutually disjoint, maintaining distinct case scopes.

\begin{appendixtable}[H]{\textwidth}
\centering
\small
\caption{\textbf{\benchmarkname statistics.} Computed over 48 local graphs and 144 sampled subgraphs.}
\label{tab:benchmark-statistics}
\begin{tabularx}{\columnwidth}{@{}Xc@{}}
\toprule
\multicolumn{2}{@{}l}{\textbf{Panel A. Benchmark composition}} \\
\midrule
Statistic & Value \\
\midrule
Domains & 24 \\
Root entities & 48 \\
Cases & 144 \\
Cases per domain & 6 \\
Cases per root & 3 \\
Total rubrics & 44,102 \\
\bottomrule
\end{tabularx}
\vspace{0.8em}

\begin{tabularx}{\columnwidth}{@{}XXcc@{}}
\toprule
\multicolumn{4}{@{}l}{\textbf{Panel B. Graph and case structure}} \\
\midrule
Graph scope & Statistic & Mean / value & Observed range \\
\midrule
Complete local graph & Edges & 1,528.6 & 1,365--1,550 \\
Sampled case subgraph & Edges/rubrics & 306.3 & 236--310 \\
Sampled case subgraph & Root children & 10 & fixed \\
Sampled case subgraph & Maximum hop distance & 3 & fixed \\
\bottomrule
\end{tabularx}
\vspace{0.8em}

\begin{tabularx}{\columnwidth}{@{}Xccc@{}}
\toprule
\multicolumn{4}{@{}l}{\textbf{Panel C. Rubric distribution by hop distance}} \\
\midrule
Rubric hop distance $\delta(r_e)$ & Rubrics & Raw share & Mean/case \\
\midrule
$\delta=1$: root $\rightarrow$ first hop & 1,440 & 3.27\% & 10.0 \\
$\delta=2$: first $\rightarrow$ second hop & 7,184 & 16.29\% & 49.9 \\
$\delta=3$: second $\rightarrow$ third hop & 35,478 & 80.45\% & 246.4 \\
\bottomrule
\end{tabularx}
\end{appendixtable}

Rubric counts are uneven across hop distances, with most rubrics occurring at greater hop distances. Our topology-aware scheme uses normalized aggregation of child-subtree values at each recursive step. Rubric contributions therefore depend on coverage, positions in the tree, and the aggregation rules; larger rubric counts at greater hop distances alone do not imply greater aggregate weight.

\begin{table}[t]
\centering
\small
\renewcommand{\arraystretch}{1.00}
\caption{\textbf{Root entities in \benchmarkname benchmark.} The table lists the two selected root entities from each of the 24 domains, for a total of 48 root entities.}
\label{tab:benchmark-root-entities}
\begin{tabularx}{\columnwidth}{@{}>{\raggedright\arraybackslash}p{0.25\columnwidth}>{\centering\arraybackslash}X>{\centering\arraybackslash}X@{}}
\toprule
Domain & Root entity 1 & Root entity 2 \\
\midrule
Adult & Fanny Hill & American burlesque \\
Art \& Design & Black Mountain College & Constructivism (art) \\
Crime \& Law & Guildford Four and Maguire Seven & Lindbergh kidnapping \\
Education \& Jobs & Highlander Research and Education Center & Educational Testing Service \\
Entertainment & The Second City & Lew Grade \\
Fashion \& Beauty & Beauty pageant & Mod (subculture) \\
Finance \& Business & John Law (economist) & Grameen Bank \\
Food \& Dining & Horn \& Hardart & Culinary Institute of America \\
Games & Roguelike & Games Workshop \\
Hardware & Xerox Alto & Cray-1 \\
Health & Ignaz Semmelweis & John Snow \\
History & Taiping Rebellion & Bandung Conference \\
Home \& Hobbies & Royal Horticultural Society & Amateur radio \\
Industrial & Birmingham Small Arms Company & Saltaire \\
Literature & Hugo Award & New Journalism \\
Politics & Non-Aligned Movement & Treaty of Waitangi \\
Religion & Albigensian Crusade & Guru Granth Sahib \\
Science \& Technology & International Geophysical Year & Cavendish Laboratory \\
Social Life & Mardi Gras in New Orleans & Temperance movement \\
Software & Apache Hadoop & OpenVMS \\
Software Development & Agile software development & GNU Compiler Collection \\
Sports \& Fitness & Kabaddi & Bandy \\
Transportation & Cunard Line & Romney, Hythe and Dymchurch Railway \\
Travel & Periplus of the Erythraean Sea & Via Francigena \\
\bottomrule
\end{tabularx}
\end{table}

\section{Metric Details}
\label{app:coverage-metric-details}

This section provides the complete coverage computation for Section~\ref{sec:coverage-evaluation}. Algorithm~\ref{alg:coverage} summarizes the procedure; the following subsections define its notation, recursive equations, and parameters.

\begin{algorithm}[!t]
\caption{Bottom-up recursive evaluation of structured knowledge coverage}
\label{alg:coverage}
\small
\renewcommand{\algorithmicrequire}{\textbf{Input:}}
\renewcommand{\algorithmicensure}{\textbf{Output:}}
\begin{algorithmic}[1]
\REQUIRE Rooted tree $T$ with root $s$ and child sets $\operatorname{Ch}(v)$; binary incoming-edge scores $x(v)\in\{0,1\}$ for $v\ne s$
\REQUIRE Parameters $\beta=0.60$, $\alpha=0.25$, $\eta=0.75$, $\rho=0.50$
\ENSURE Case-level breadth $B_c$, depth $D_c$, and overall coverage $S_c$
\item[]\vspace{0.2em}\hspace*{-\leftmargin}\textbf{Main procedure}
\STATE $\mathcal A\gets\operatorname{Ch}(s)$ \hfill {\footnotesize\textit{Selected research directions}}
\STATE \textbf{for} each $b\in\mathcal A$ \textbf{do} $(B(b),D(b))\gets\operatorname{Evaluate}(b)$
\STATE $B_c\gets\frac{1}{|\mathcal A|}\sum_{b\in\mathcal A}B(b)$; \quad $D_c\gets\operatorname{RankAgg}\bigl((D(b))_{b\in\mathcal A},\rho\bigr)$
\STATE $S_c\gets\sqrt{B_cD_c}$; \quad \textbf{return} $(B_c,D_c,S_c)$
\item[]\vspace{0.3em}\hspace*{-\leftmargin}\textbf{Recursive evaluation}
\STATE \textbf{function} $\operatorname{Evaluate}(v)$, $v\ne s$:
\STATE \hspace{1em}$C\gets\operatorname{Ch}(v)$
\STATE \hspace{1em}\textbf{if} $C=\emptyset$, \textbf{return} $(x(v),x(v))$ \hfill {\footnotesize\textit{Leaf initialization}}
\STATE \hspace{1em}\textbf{for} each $w\in C$ \textbf{do} $(B(w),D(w))\gets\operatorname{Evaluate}(w)$
\STATE \hspace{1em}$B(v)\gets \beta x(v)+(1-\beta)\,\frac{1}{|C|}\sum_{w\in C}B(w)$ \hfill {\footnotesize\textit{Local-descendant coverage fusion}}
\STATE \hspace{1em}$d\gets\operatorname{RankAgg}\bigl((D(w))_{w\in C},\rho\bigr)$
\STATE \hspace{1em}$D(v)\gets\alpha x(v)+(1-\alpha)d\bigl[\eta+(1-\eta)x(v)\bigr]$ \hfill {\footnotesize\textit{Chain continuity reward}}
\STATE \hspace{1em}\textbf{return} $(B(v),D(v))$
\STATE \textbf{end function}
\item[]\vspace{0.3em}\hspace*{-\leftmargin}\textit{Helper function}
\STATE \textbf{function} $\operatorname{RankAgg}(Y,\rho)$: \hfill {\footnotesize\textit{Rank-decayed aggregation}}
\STATE \hspace{1em}\textbf{if} $Y$ is empty, \textbf{return} $0$
\STATE \hspace{1em}Sort $Y$ as $y_{(1)}\geq\cdots\geq y_{(n)}$, where $n=|Y|$
\STATE \hspace{1em}\textbf{return} $\bigl(\sum_{i=1}^{n}\rho^{i-1}y_{(i)}\bigr)\,/\,\bigl(\sum_{i=1}^{n}\rho^{i-1}\bigr)$
\STATE \textbf{end function}
\end{algorithmic}
\end{algorithm}

\subsection{Notation and Recursive Evaluation}

For a case $c$, let $T=(V,E)$ be its reference tree, with root $s$ and child sets $\operatorname{Ch}(v)$. The children of the root, $\mathcal A=\operatorname{Ch}(s)$, represent the selected research directions. Each non-root node $v$ has a binary incoming-edge rubric score $x(v)\in\{0,1\}$: $1$ (covered) for a \emph{Satisfied} judge verdict and $0$ (uncovered) for a \emph{Partially Satisfied} or \emph{Not Satisfied} verdict. The root has no incoming-edge rubric.

We associate two recursive values with each non-root node: $B(v)$ for breadth and $D(v)$ for depth. Both recursions begin at every actual leaf, including those at shallower levels:
\begin{equation}
B(v)=D(v)=x(v),\qquad \operatorname{Ch}(v)=\emptyset.
\label{eq:coverage-leaf-initialization}
\end{equation}

Algorithm~\ref{alg:coverage} invokes $\operatorname{Evaluate}$ on each child of the root. Each call recursively evaluates the child subtrees before computing and returning the current node's breadth and depth values, so scores are aggregated from children to parents following a postorder traversal.

\subsection{Breadth: Local-descendant Coverage Fusion}

Breadth measures how comprehensively a report covers the selected research directions. A covered relation provides local evidence but does not alone establish coverage of its descendant subtopics. We therefore compute breadth as a weighted average of the current relation’s coverage and the mean breadth of its child subtrees, allowing coverage of descendant subtopics to contribute recursively to the current direction’s breadth.

For a non-leaf, non-root node $v$, let $m_B(v)$ be the mean of $B(w)$ over $w\in\operatorname{Ch}(v)$. We define
\begin{equation}
B(v)=\beta x(v)+(1-\beta)m_B(v).
\label{eq:breadth-recursion}
\end{equation}
Here, $0<\beta<1$ weights the current relation’s coverage, while $1-\beta$ weights the mean breadth of its child subtrees. Covering a non-leaf relation alone therefore does not yield full breadth credit for its subtree. Conversely, covered descendants can still contribute when the current relation is uncovered, partially compensating for the current relation’s lack of coverage. Their contribution propagates upward through the recursion to the branch-level breadth score.

Case-level breadth is
\begin{equation}
B_c=\frac{1}{|\mathcal A|}\sum_{b\in\mathcal A}B(b).
\label{eq:breadth-case}
\end{equation}
This gives equal weight to the selected directions while retaining the descendant coverage within each branch.

\subsection{Depth: Knowledge-Chain Aggregation and Continuity}

Depth measures how far a report follows chains of related knowledge within selected research directions. Deeper coverage along one knowledge chain can demonstrate substantial depth; coverage along additional chains provides further evidence but should have diminishing influence. We capture this objective through rank-decayed aggregation. For child-subtree values sorted as $d_{(1)}\geq\cdots\geq d_{(n)}$, define
\begin{equation}
\operatorname{RankAgg}_{\rho}(d_1,\ldots,d_n)
=\frac{\sum_{i=1}^{n}\rho^{i-1}d_{(i)}}{\sum_{i=1}^{n}\rho^{i-1}}.
\label{eq:rank-route-aggregation}
\end{equation}
For $0<\rho<1$, higher-valued subtrees receive greater weight, while the others retain a contribution.

Depth should also reward continuous coverage along a knowledge chain, where successive parent--child relations are covered. Descendant evidence deserves greater credit when the report also covers the relation linking the current node to its parent, preserving coverage continuity at that link. When this relation is uncovered, it creates a gap in the chain's coverage; evidence farther along the chain should receive reduced credit without being discarded entirely. The current relation must also contribute independently. For a non-leaf node, let $d(v)$ be the rank-decayed aggregation of its children's depth values. We combine these requirements as
\begin{equation}
D(v)=\alpha x(v)+(1-\alpha)d(v)[\eta+(1-\eta)x(v)].
\label{eq:depth-recursion}
\end{equation}
The term $\alpha x(v)$ credits the current relation directly. The factor $\eta+(1-\eta)x(v)$ rewards continuity: with binary scores, it equals one when $x(v)=1$ and $\eta$ when $x(v)=0$. This favors continuously covered knowledge chains while preserving limited credit for descendant evidence beyond a coverage gap.

After recursive evaluation, case-level depth applies the same operator across the selected directions:
\begin{equation}
D_c=\operatorname{RankAgg}_{\rho}\bigl((D(b))_{b\in\mathcal A}\bigr).
\label{eq:depth-case}
\end{equation}

Breadth therefore weights directions equally, whereas depth emphasizes those with deeper coverage along related knowledge chains.

\subsection{Metric Parameters and Final Aggregation}

Table~\ref{tab:coverage-parameters} lists the fixed defaults used across cases and agents; Section~\ref{sec:metric-sensitivity} examines ranking sensitivity to these choices.

\begin{appendixtable}[H]{\textwidth}
\small
\caption{\textbf{Coverage metric parameters.}}
\label{tab:coverage-parameters}
\begin{tabularx}{\linewidth}{@{}lcX@{}}
\toprule
Parameter & Default & Role \\
\midrule

$\beta$ & $0.60$ & Weight assigned to the current relation’s coverage in the breadth recursion at non-leaf nodes; $1-\beta$ weights the mean breadth of its child subtrees. \\

$\alpha$ & $0.25$ & Weight of the direct current-edge contribution in depth at non-leaf nodes. \\

$\eta$ & $0.75$ & Retention factor applied to descendant depth when the current edge is uncovered. \\

$\rho$ & $0.50$ & Ratio between successive rank weights in depth aggregation. \\
\bottomrule
\end{tabularx}
\end{appendixtable}

We treat breadth and depth as equally important and compute case-level overall coverage as $S_c=\sqrt{B_cD_c}$, making weak coverage in either dimension consequential. For each deep research agent, we report the mean case-level breadth, depth, and overall coverage across all 144 benchmark cases.

\Needspace{9\baselineskip}
\section{Benchmark Validation}
\label{app:benchmark-validation}

\subsection{Benchmark Size Selection}
\label{app:sample-size-stability}

In fact, we initially constructed 720 research cases, covering 10 root entities in each of 24 domains and three cases per root. We then evaluated o3 Deep Research, Perplexity Sonar Deep Research, and Qwen Deep Research on all 720 cases. To determine whether a smaller benchmark could maintain stable results, we compare different numbers of sampled roots per domain. Table~\ref{tab:sample-size-stability} summarizes score stability across sample sizes.

\newsavebox{\samplestabilitytablebox}
\sbox{\samplestabilitytablebox}{%
\footnotesize
\setlength{\tabcolsep}{3pt}%
\begin{tabular}{@{}ccccc@{}}
\toprule
$k$ & Cases & MAE $\downarrow$ & SD $\downarrow$ & Within $\pm2$ pp $\uparrow$ \\
\midrule
1 & 72  & 1.481 & 1.610 & 75.0\% \\
\textbf{2} & \textbf{144} & \textbf{0.768} & \textbf{0.926} & \textbf{96.7\%} \\
3 & 216 & 0.570 & 0.740 & 98.3\% \\
4 & 288 & 0.458 & 0.533 & 100.0\% \\
5 & 360 & 0.350 & 0.448 & 100.0\% \\
\bottomrule
\end{tabular}%
}
\Needspace{9\baselineskip}
\begin{wraptable}{r}{\wd\samplestabilitytablebox}
\centering
\caption{\textbf{Sample-size stability of the overall score.} MAE and mean SD are in percentage points.}
\label{tab:sample-size-stability}
\usebox{\samplestabilitytablebox}
\end{wraptable}

We use $k$ to denote the number of roots sampled from each domain. One sample is constructed by selecting $k$ roots from each of the 24 domains and retaining all three cases under every selected root. For each value of $k$, we independently construct 20 such samples. We then compare the three systems' overall scores on these samples with their results on all 720 cases.

The overall score for each case is the geometric mean of its breadth and depth scores, and a system's overall score is averaged across the selected cases. For each $k$, the three systems and 20 samples produce 60 scores. MAE measures their average absolute difference from the corresponding 720-case scores. Mean SD averages the standard deviations of the 20 scores for each system. Within $\pm2$ pp reports the proportion of the 60 scores that differ from their corresponding 720-case scores by no more than two percentage points.

At $k=2$, MAE decreases from 1.481 to 0.768 percentage points, 96.7\% of the sampled scores remain within two percentage points of the complete-pool results, and all 20 samples preserve the system ordering. Larger samples provide only modest further improvements. We therefore select two roots per domain and use the resulting 144-case benchmark.

\subsection{Research Query Quality Criteria}
\label{app:query-quality}

\paragraph{Research Value.}
This criterion asks whether the query warrants a research report. A strong query calls for substantive investigation, retrieval, and synthesis across relevant evidence; a request answerable through a short factual lookup has limited research value.

\paragraph{Public Answerability.}
This criterion asks whether reliable, publicly available sources can, in principle, support an answer to the query. The answer need not appear in a single source, but the requested findings must be verifiable without private information or unsupported speculation.

\paragraph{Query--Domain Match.}
This criterion asks whether the query's central topic and requested analysis fit its assigned domain. The domain is shown alongside the query during review. Cross-domain connections are compatible with a match when the assigned domain remains central; a merely incidental connection is insufficient.

\subsection{Additional Results from Prior Benchmarks}
\label{app:additional-prior-benchmarks}

Table~\ref{tab:prior-benchmark-results} compiles selected results from prior benchmarks, focusing on model families represented in Table~\ref{tab:main-results} and evaluation dimensions with definitions similar to ours. These results serve as an external reference for our evaluation, with each row labeled by its operating setting in the source study. Setting~1 (Deep Research product) denotes an end-to-end research product or system that integrates research planning, iterative retrieval, and report generation. Setting~2 (native search / search tools) denotes a model using native search or search tools during generation. Setting~3 (external framework + backbone) denotes a backbone model embedded in an external retrieval or research framework that organizes tool calls and the research workflow. These labels describe the source experiments; the same model can use different settings across studies. In HiEviDR~\citep{sun2026hievidr}, $3_{\mathrm{R}}$ and $3_{\mathrm{D}}$ denote the RAG and Deep Research workflows, respectively. For ResearchRubrics~\citep{sharma2026researchrubrics}, we report binary compliance on tasks with high conceptual breadth or deep logical nesting; DR-Arena's Depth/Width values describe the composition of failure types, rather than standalone ability scores~\citep{gao2026dr}.

\begin{table}[t]
\centering
\begingroup
\newcommand{\priorbenchmarkpanel}[2]{\par\noindent\colorbox{black!7}{\parbox{\dimexpr\linewidth-2\fboxsep\relax}{\strut\textbf{\hypersetup{hidelinks}#1}\hfill{\fontsize{7}{8}\selectfont #2}}}\par\vspace{2pt}}
\captionsetup{font=footnotesize,skip=5pt}\caption{\textbf{Selected results from prior benchmarks.} Slash-separated scores follow the listed metric order; score scales are specific to each benchmark and metric.}
\label{tab:prior-benchmark-results}
\noindent{\raggedright\fontsize{8}{9.5}\selectfont\textbf{Source setting:} 1 = Deep Research product; 2 = native search / search tools; 3 = external framework + backbone. For HiEviDR, 3\textsubscript{R} = RAG and 3\textsubscript{D} = Deep Research.\par}
\vspace{7pt}
\noindent
\begin{minipage}[t]{0.485\textwidth}\vspace{0pt}\fontsize{8}{9.4}\selectfont
\noindent\textbf{Setting / Model or system}\hfill\textbf{Score(s)}\par\vspace{2pt}\hrule\vspace{4pt}
\priorbenchmarkpanel{\href{https://arxiv.org/pdf/2506.11763v1}{(a) DeepResearch Bench}}{Table 1}
{\fontsize{7.5}{8.5}\selectfont\itshape RACE Comp. $\uparrow$ / RACE Depth $\uparrow$\par}\vspace{2pt}
\setlength{\tabcolsep}{1.5pt}\renewcommand{\arraystretch}{1.06}
\begin{tabularx}{\linewidth}{@{}>{\color{black!65}}c>{\raggedright\arraybackslash}Xc@{}}
1 & OpenAI DR & 46.87\,/\,45.25 \\
1 & Perplexity DR & 40.69\,/\,39.39 \\
1 & Grok Deeper Search & 37.97\,/\,35.37 \\
1 & Gemini-2.5-Pro DR & 48.53\,/\,48.50 \\
2 & Perplexity-Sonar-Reasoning-Pro(high) & 37.38\,/\,36.11 \\
2 & Gemini-2.5-Pro-Grounding & 34.06\,/\,29.79 \\
\end{tabularx}
\par\vspace{4pt}
\priorbenchmarkpanel{\href{https://arxiv.org/pdf/2601.08536v3}{(b) DeepResearch Bench II}}{Table 3}
{\fontsize{7.5}{8.5}\selectfont\itshape InfoRecall $\uparrow$\par}\vspace{2pt}
\setlength{\tabcolsep}{1.5pt}\renewcommand{\arraystretch}{1.06}
\begin{tabularx}{\linewidth}{@{}>{\color{black!65}}c>{\raggedright\arraybackslash}Xc@{}}
1 & OpenAI-GPT-o3 DR & 39.98 \\
1 & Gemini-3-Pro DR & 39.09 \\
1 & Gemini-2.5-Pro DR & 34.91 \\
1 & Qwen3-Max DR & 34.18 \\
1 & Grok Deep Search & 33.52 \\
1 & Perplexity Research & 33.05 \\
\end{tabularx}
\par\vspace{4pt}
\priorbenchmarkpanel{\href{https://arxiv.org/pdf/2511.07685v1}{(c) ResearchRubrics}}{Figure 12}
{\fontsize{7.5}{8.5}\selectfont\itshape High conceptual breadth $\uparrow$ / Deep logical nesting $\uparrow$\par}\vspace{2pt}
\setlength{\tabcolsep}{1.5pt}\renewcommand{\arraystretch}{1.06}
\begin{tabularx}{\linewidth}{@{}>{\color{black!65}}c>{\raggedright\arraybackslash}Xc@{}}
1 & Gemini DR & 56.5\%\,/\,59.6\% \\
1 & ChatGPT DR & 53.1\%\,/\,57.5\% \\
1 & Perplexity DR & 50.8\%\,/\,47.5\% \\
\end{tabularx}
\par\vspace{4pt}
\priorbenchmarkpanel{\href{https://arxiv.org/html/2602.01590}{(d) Wiki Live Challenge}}{Table 1}
{\fontsize{7.5}{8.5}\selectfont\itshape Broad $\uparrow$ / Cov. Wiki $\uparrow$\par}\vspace{2pt}
\setlength{\tabcolsep}{1.5pt}\renewcommand{\arraystretch}{1.06}
\begin{tabularx}{\linewidth}{@{}>{\color{black!65}}c>{\raggedright\arraybackslash}Xc@{}}
1 & OpenAI o3 DR & 45.75\,/\,25.12 \\
1 & Perplexity DR & 37.63\,/\,29.21 \\
1 & Qwen-3-max DR & 40.00\,/\,22.22 \\
1 & Grok Deep Search & 35.75\,/\,20.73 \\
1 & Gemini-2.5-pro DR & 59.88\,/\,30.76 \\
1 & Gemini-3-pro DR & 67.12\,/\,28.83 \\
3 & Langchain (GPT-5) & 59.88\,/\,20.96 \\
\end{tabularx}
\par\vspace{4pt}
\priorbenchmarkpanel{\href{https://arxiv.org/pdf/2601.09688v1}{(e) DeepResearchEval}}{Table 3}
{\fontsize{7.5}{8.5}\selectfont\itshape Covera. $\uparrow$\par}\vspace{2pt}
\setlength{\tabcolsep}{1.5pt}\renewcommand{\arraystretch}{1.06}
\begin{tabularx}{\linewidth}{@{}>{\color{black!65}}c>{\raggedright\arraybackslash}Xc@{}}
1 & OpenAI DR & 8.6 \\
1 & Perplexity DR & 8.2 \\
1 & Grok4 DR & 8.5 \\
1 & Qwen-3-235B-A22B DR & 8.0 \\
1 & Gemini-2.5-Pro DR & 9.2 \\
\end{tabularx}
\par\vspace{4pt}
\end{minipage}\hfill%
\begin{minipage}[t]{0.485\textwidth}\vspace{0pt}\fontsize{8}{9.4}\selectfont
\noindent\textbf{Setting / Model or system}\hfill\textbf{Score(s)}\par\vspace{2pt}\hrule\vspace{4pt}
\priorbenchmarkpanel{\href{https://arxiv.org/pdf/2603.00582v2}{(f) Super Research}}{Table 1}
{\fontsize{7.5}{8.5}\selectfont\itshape Coverage $\uparrow$\par}\vspace{2pt}
\setlength{\tabcolsep}{1.5pt}\renewcommand{\arraystretch}{1.06}
\begin{tabularx}{\linewidth}{@{}>{\color{black!65}}c>{\raggedright\arraybackslash}Xc@{}}
1 & o3 DR & 23.44 \\
1 & Sonar DR & 28.26 \\
1 & Gemini DR & 33.15 \\
2 & Grok-4-1-fast & 30.09 \\
3 & Deepseek-r1 & 24.21 \\
3 & Minimax-m2.1 & 20.91 \\
\end{tabularx}
\par\vspace{4pt}
\priorbenchmarkpanel{\href{https://arxiv.org/pdf/2608.02163v1}{(g) From Simple QA to Deep Research}}{Table 4}
{\fontsize{7.5}{8.5}\selectfont\itshape Agent $q_h\uparrow$ / $q_o\uparrow$ / $q_a\uparrow$\par}\vspace{2pt}
\setlength{\tabcolsep}{1.5pt}\renewcommand{\arraystretch}{1.06}
\begin{tabularx}{\linewidth}{@{}>{\color{black!65}}c>{\raggedright\arraybackslash}Xc@{}}
3 & GPT-5.6 Terra & 0.86\,/\,0.67\,/\,0.70 \\
3 & GLM-5.2 & 0.80\,/\,0.68\,/\,0.69 \\
3 & DeepSeek-V4-Pro & 0.83\,/\,0.68\,/\,0.73 \\
3 & Qwen3.7-Max & 0.83\,/\,0.67\,/\,0.71 \\
3 & MiniMax-M2.5 & 0.67\,/\,0.42\,/\,0.55 \\
\end{tabularx}
\par\vspace{4pt}
\priorbenchmarkpanel{\href{https://arxiv.org/pdf/2607.25151v1}{(h) HiEviDR}}{Table 3}
{\fontsize{7.5}{8.5}\selectfont\itshape Wiki Report $\uparrow$ / Arxiv Report $\uparrow$\par}\vspace{2pt}
\setlength{\tabcolsep}{1.5pt}\renewcommand{\arraystretch}{1.06}
\begin{tabularx}{\linewidth}{@{}>{\color{black!65}}c>{\raggedright\arraybackslash}Xc@{}}
3\textsubscript{R} & Gemma4-31B-it & 18.28\,/\,19.26 \\
3\textsubscript{R} & GPT-5.6-sol & 18.88\,/\,19.69 \\
3\textsubscript{R} & Grok-4.5 & 18.66\,/\,19.43 \\
3\textsubscript{D} & Gemma4-31B-it & 19.04\,/\,19.58 \\
3\textsubscript{D} & GPT-5.6-sol & 19.33\,/\,19.73 \\
3\textsubscript{D} & Grok-4.5 & 19.00\,/\,18.90 \\
\end{tabularx}
\par\vspace{4pt}
\priorbenchmarkpanel{\href{https://arxiv.org/pdf/2601.10504v3}{(i) DR-Arena}}{Table 3 / Figure 6}
{\fontsize{7.5}{8.5}\selectfont\itshape DR-Arena Elo $\uparrow$ / Depth-type share / Width-type share\par}\vspace{2pt}
\setlength{\tabcolsep}{1.5pt}\renewcommand{\arraystretch}{1.06}
\begin{tabularx}{\linewidth}{@{}>{\color{black!65}}c>{\raggedright\arraybackslash}Xc@{}}
2 & GPT-5.1-Search & 1084\,/\,27\%\,/\,26\% \\
2 & Gemini-2.5-Pro-Grounding & 1054\,/\,33\%\,/\,24\% \\
2 & o3-Search & 1041\,/\,19\%\,/\,30\% \\
2 & Grok-4-Search & 958\,/\,30\%\,/\,25\% \\
2 & Claude-Opus-4.1-Search & 921\,/\,31\%\,/\,25\% \\
2 & Perplexity-Sonar-Pro-High & 942\,/\,37\%\,/\,26\% \\
\end{tabularx}
\par\vspace{4pt}
\priorbenchmarkpanel{\href{https://proceedings.iclr.cc/paper_files/paper/2026/file/998f8d8ff2697da2eae5f87143668754-Paper-Conference.pdf}{(j) LiveResearchBench}}{Table 1}
{\fontsize{7.5}{8.5}\selectfont\itshape Coverage \& Comprehensiveness $\uparrow$\par}\vspace{2pt}
\setlength{\tabcolsep}{1.5pt}\renewcommand{\arraystretch}{1.06}
\begin{tabularx}{\linewidth}{@{}>{\color{black!65}}c>{\raggedright\arraybackslash}Xc@{}}
1 & OpenAI o3 DR & 85.0 \\
1 & Perplexity Sonar DR & 65.5 \\
1 & Grok-4 DR & 86.3 \\
1 & Gemini DR & 75.8 \\
2 & GPT-5 & 83.4 \\
2 & Gemini 2.5 Pro & 73.1 \\
2 & Claude 4.1 Opus & 50.8 \\
2 & Perplexity Sonar Reasoning Pro & 46.7 \\
3 & Deerflow+ (w. GPT-5) & 61.6 \\
3 & Open DR (w. GPT-5) & 65.3 \\
\end{tabularx}
\par\vspace{4pt}
\end{minipage}
\endgroup
\end{table}

Cross-benchmark comparisons require accounting for differences in experimental settings, model versions, and metric definitions. Studies may use different retrieval resources, tool interfaces, and execution budgets; evaluations conducted at different times often cover different versions within the same model family; and even similarly named metrics may differ in their reference content, scoring criteria, and aggregation rules. We therefore compare relative performance trends among related models and interpret agreement and variation in light of these conditions.

For model pairs with relatively clear correspondences, our results agree with several prior studies. In Table~\ref{tab:main-results}, o3 Deep Research achieves higher overall coverage than Perplexity Sonar Deep Research. The corresponding OpenAI--Perplexity product-family ordering also appears in the selected dimensions of DeepResearch Bench~\citep{du2026deepresearch}, DeepResearch Bench II~\citep{li2026deepresearch}, ResearchRubrics, Wiki Live Challenge's Broad metric~\citep{wang2026wiki}, DeepResearchEval~\citep{wang2026deepresearcheval}, and LiveResearchBench~\citep{wang2026liveresearchbench}. The advantage of o3 Deep Research over Qwen Deep Research is also consistent with the corresponding product-family results in DeepResearch Bench II, Wiki Live Challenge, and DeepResearchEval. Among native-search models, LiveResearchBench orders GPT-5 above Gemini 2.5 Pro and Sonar Reasoning Pro, matching the overall-coverage order of the corresponding model families in Table~\ref{tab:main-results}. Although versions and protocols differ, these recurring relative advantages provide useful external context for our results.

Despite these recurring trends, some metrics and model pairs exhibit different relative orderings, including Wiki Live Challenge's Cov.\ Wiki, Super Research's Coverage~\citep{dong2026super}, and the GLM-5.2--DeepSeek-V4-Pro comparison in From Simple QA to Deep Research~\citep{wang2026simple}. Taken together, these comparisons show that our evaluation recovers several relative performance trends reported in prior benchmarks, providing complementary evidence for its validity.

\clearpage
\subsection{Validating the design of evaluation metrics}
\label{app:metric-validation}
\begingroup

\subsubsection*{Comparison protocol}
We compare our final metrics with expert assessments of 72 reports from six agents across 12 cases, each from a different domain. The six agents comprise Perplexity Sonar Deep Research; GPT-5.6 Sol, Gemini 3.1 Pro Preview, and Sonar Reasoning Pro with native search; and GLM 5.2 and MiniMax M3 under the local research framework. For each case, the same three annotators independently rate all six reports on breadth, depth, and overall quality using a 1--10 integer scale, and their mean ratings serve as the human reference. Annotation details are provided in Appendix~\ref{app:report-level-annotation}.

All schemes use identical reports, reference trees, and binary rubric judgments. We compare our final metrics with six flat and structural coverage baselines (two each for overall quality, breadth, and depth), and conduct four component ablations to examine the contributions of key design choices in our final metrics.

We compute Kendall's \(\tau_b\) and Spearman's \(\rho_s\) between scheme scores and mean expert ratings in the corresponding dimension within each case, then average the correlations equally across the 12 cases. Tie rate is the percentage of report pairs with equal metric scores within each case, averaged across cases.

\subsubsection*{Baseline schemes}
For each case, we follow Appendix~\ref{app:coverage-metric-details}'s notation: reference tree \(T=(V,E)\), root \(s\), selected research directions \(\mathcal A=\operatorname{Ch}(s)\), and binary incoming-edge scores \(x(v)\). Let \(E_\ell\) contain edges whose target nodes have hop distance \(\ell\in\{1,2,3\}\) from \(s\), with coverage \(C_\ell=|E_\ell|^{-1}\sum_{(u,v)\in E_\ell}x(v)\). For \(b\in\mathcal A\), \(E_b\) contains the edge entering \(b\) and all edges in its subtree; \(E_{b,\ell}=E_b\cap E_\ell\). All baseline scores are case-level, with denominators fixed by the reference tree.

\Needspace{4\baselineskip}
\noindent\textbf{S1: Flat coverage.} All rubric-bearing edges receive equal weight:

\[
S_c^{\mathrm{flat}}=\frac{1}{|E|}\sum_{(u,v)\in E}x(v).
\]

\Needspace{4\baselineskip}
\noindent\textbf{S2: Level-balanced coverage.} Coverage at hop distances \(1,2,3\) receives equal weight:

\[
S_c^{\mathrm{level}}=\frac{C_1+C_2+C_3}{3}.
\]

\Needspace{4\baselineskip}
\noindent\textbf{B1: Branch coverage.} A research direction (branch) is covered if any of its edges is satisfied:

\[
B_c^{\mathrm{branch}}
=\frac{1}{|\mathcal A|}\sum_{b\in\mathcal A}
\mathbf{1}\!\left[\sum_{(u,v)\in E_b}x(v)>0\right].
\]

\Needspace{4\baselineskip}
\noindent\textbf{B2: Multi-level direction coverage.} The fraction of covered research directions is averaged over hop distances \(1,2,3\):

\[
B_c^{\mathrm{multi}}
=\frac{1}{3|\mathcal A|}\sum_{\ell=1}^{3}\sum_{b\in\mathcal A}
\mathbf{1}\!\left[\sum_{(u,v)\in E_{b,\ell}}x(v)>0\right].
\]

\Needspace{4\baselineskip}
\noindent\textbf{D1: Terminal-level coverage.} Only third-hop coverage is counted: \(D_c^{\mathrm{terminal}}=C_3\).

\Needspace{4\baselineskip}
\noindent\textbf{D2: Within-path pair coverage.} For actual three-hop root-to-leaf paths \(\mathcal P_3\), each representing a knowledge chain, average the co-coverage of all three edge pairs:

\[
D_c^{\mathrm{pair}}
=\frac{1}{|\mathcal P_3|}\sum_{p\in\mathcal P_3}
\frac{x(v_{p1})x(v_{p2})+x(v_{p1})x(v_{p3})+x(v_{p2})x(v_{p3})}{3}.
\]

Here \(v_{p\ell}\) is the node at hop distance \(\ell\) from \(s\) on path \(p\). Paths receive equal weight; shared ancestors are counted per path, and shorter paths are excluded.

\subsubsection*{Ablation schemes}
Each ablation in Table~\ref{tab:metric-ablation-schemes} changes one parameter of our final metrics (F0; Appendix~\ref{app:coverage-metric-details}), using an endpoint value to disable a component and retaining the other defaults: \(\beta=0.60\), \(\alpha=0.25\), \(\eta=0.75\), and \(\rho=0.50\). As in Appendix~\ref{app:coverage-metric-details}, \(d(v)=\operatorname{RankAgg}_{\rho}((D(w))_{w\in\operatorname{Ch}(v)})\); child scores are recomputed recursively for each variant.

\begin{appendixtable}[H]{\textwidth}
\caption{Ablation schemes and parameter changes.}
\label{tab:metric-ablation-schemes}
\begingroup
\setlength{\tabcolsep}{3pt}
\begin{tabularx}{\linewidth}{@{}l>{\raggedright\arraybackslash}p{0.27\linewidth}c l>{\raggedright\arraybackslash}X@{}}
\toprule
ID & Scoring scheme & \shortstack{Parameter\\change} & \shortstack[l]{Evaluated\\metric} & Explanation \\
\midrule
A-B1 & w/o descendant fusion & \(\beta:0.60\to1\) & Breadth & Reduces case-level breadth to \(C_1\). \\
A-D1 & w/o continuity discount & \(\eta:0.75\to1\) & Depth & Removes the continuity discount. \\
A-D2 & w/o descendant credit beyond gaps & \(\eta:0.75\to0\) & Depth & Blocks descendant credit at uncovered relations. \\
A-D3 & w/o direct relation contribution & \(\alpha:0.25\to0\) & Depth & Removes direct non-leaf credit but retains the continuity factor. \\
\bottomrule
\end{tabularx}
\endgroup
\end{appendixtable}

All variants preserve actual-leaf initialization \(B(v)=D(v)=x(v)\) and case-level aggregation rules.

\subsubsection*{Complete results}

\begin{appendixtable}[H]{\textwidth}
\caption{\textbf{Agreement of scoring schemes with expert assessments.} Kendall's \(\tau_b\) and Spearman's \(\rho_s\) are computed within each case and averaged across 12 cases. Tie rate is the percentage of report pairs with equal metric scores within each case, averaged across cases. F0 denotes final metrics; A-prefixed IDs denote ablations.}
\label{tab:metric-validation-complete}
\begin{tabular}{@{}lllrrr@{}}
\toprule
Metric & ID & Scoring scheme & $\tau_b\uparrow$ & $\rho_s\uparrow$ & Tie rate (\%) \\
\midrule
Overall & F0 & Final metrics & \textbf{0.751} & \textbf{0.833} & 0.56 \\
 & S2 & Level-balanced coverage & 0.615 & 0.718 & 0.56 \\
 & S1 & Flat coverage & 0.508 & 0.597 & 4.44 \\
\midrule
Breadth & F0 & Final metrics & \textbf{0.696} & \textbf{0.777} & 0.56 \\
 & B1 & Branch coverage & 0.533 & 0.592 & 26.11 \\
 & B2 & Multi-level direction coverage & 0.463 & 0.545 & 8.89 \\
 & A-B1 & w/o descendant fusion & 0.643 & 0.714 & 16.11 \\
\midrule
Depth & F0 & Final metrics & \textbf{0.727} & \textbf{0.813} & 0.56 \\
 & D2 & Within-path pair coverage & 0.577 & 0.664 & 2.22 \\
 & D1 & Terminal-level coverage & 0.488 & 0.575 & 9.44 \\
 & A-D1 & w/o continuity discount & 0.668 & 0.761 & 0.56 \\
 & A-D2 & w/o descendant credit beyond gaps & 0.653 & 0.755 & 1.11 \\
 & A-D3 & w/o direct relation contribution & 0.583 & 0.678 & 3.33 \\
\bottomrule
\end{tabular}
\end{appendixtable}

Across the evaluated schemes (Table~\ref{tab:metric-validation-complete}), F0 ranks highest under both correlations in all three dimensions. Removing descendant fusion lowers breadth \(\tau_b\) from 0.696 to 0.643; removing the continuity discount lowers depth \(\tau_b\) from 0.727 to 0.668. The other depth ablations also reduce agreement. Branch coverage and first-hop-only coverage have tie rates of 26.11\% and 16.11\%, respectively, compared with 0.56\% for our final metrics, demonstrating that our final metrics distinguish substantially more report pairs than these coarser coverage schemes.
\par
\endgroup

\Needspace{4\baselineskip}
\subsection{Textual Organization of Covered Knowledge Chains}
\label{app:chain-textual-organization}
\begingroup

To address the concern about whether consecutive knowledge relations in the reference tree are also coherently presented in the report, we examine the textual proximity of their supporting evidence and how related knowledge is connected across sections.

\newsavebox{\chainlocalitytablebox}
\sbox{\chainlocalitytablebox}{%
\begin{tabular}{@{}lc@{}}
\toprule
Covered chain & Same-section rate (\%) \\
\midrule
1 hop: two consecutive rubrics & 87.12 \\
2 hops: three consecutive rubrics & 87.92 \\
\bottomrule
\end{tabular}%
}
\par\addvspace{\baselineskip}
\begingroup
\begin{wraptable}{r}{\wd\chainlocalitytablebox}
\centering
\caption{Evidence locality for consecutive satisfied rubrics. Same-section rate measures the proportion of covered chains whose supporting evidence lies in the same main body section, averaged equally across reports.}
\label{tab:chain-evidence-locality}
\usebox{\chainlocalitytablebox}
\end{wraptable}

To this end, we analyzed 500 sampled reports. For groups of consecutive rubrics originally judged satisfied, we used Qwen3.8 Flash to locate supporting evidence sentences and checked whether they appeared in the same main body section. A main body section is a thematic division of the report, identified from its heading hierarchy.

In this subset, evidence for both chain lengths is predominantly located within the same main body section (Table~\ref{tab:chain-evidence-locality}). We further inspected examples where the supporting evidence appeared in different sections and found that knowledge points in a coherent logical chain do not necessarily have to appear consecutively in the text. Reports may devote separate sections to different explanatory steps, analytical dimensions, or historical stages, or introduce a phenomenon in one section before tracing its origins in another. In such cases, mechanisms, supporting evidence, sources, or applicability conditions can connect explanations across sections and sustain meaningful progression. Appendix~\ref{app:example-showcase} illustrates how evidence from different report sections can support a coherent knowledge chain.
\finishwraptable
\endgroup
\par\addvspace{\baselineskip}
\par
\endgroup

\Needspace{8\baselineskip}
\section{Detailed Results and Analysis}
\label{app:detailed-results}

\subsection{Breadth and Depth across Domains}
\label{app:domain-coverage-profiles}

\begin{figure}[H]
\centering
\includegraphics[width=\textwidth]{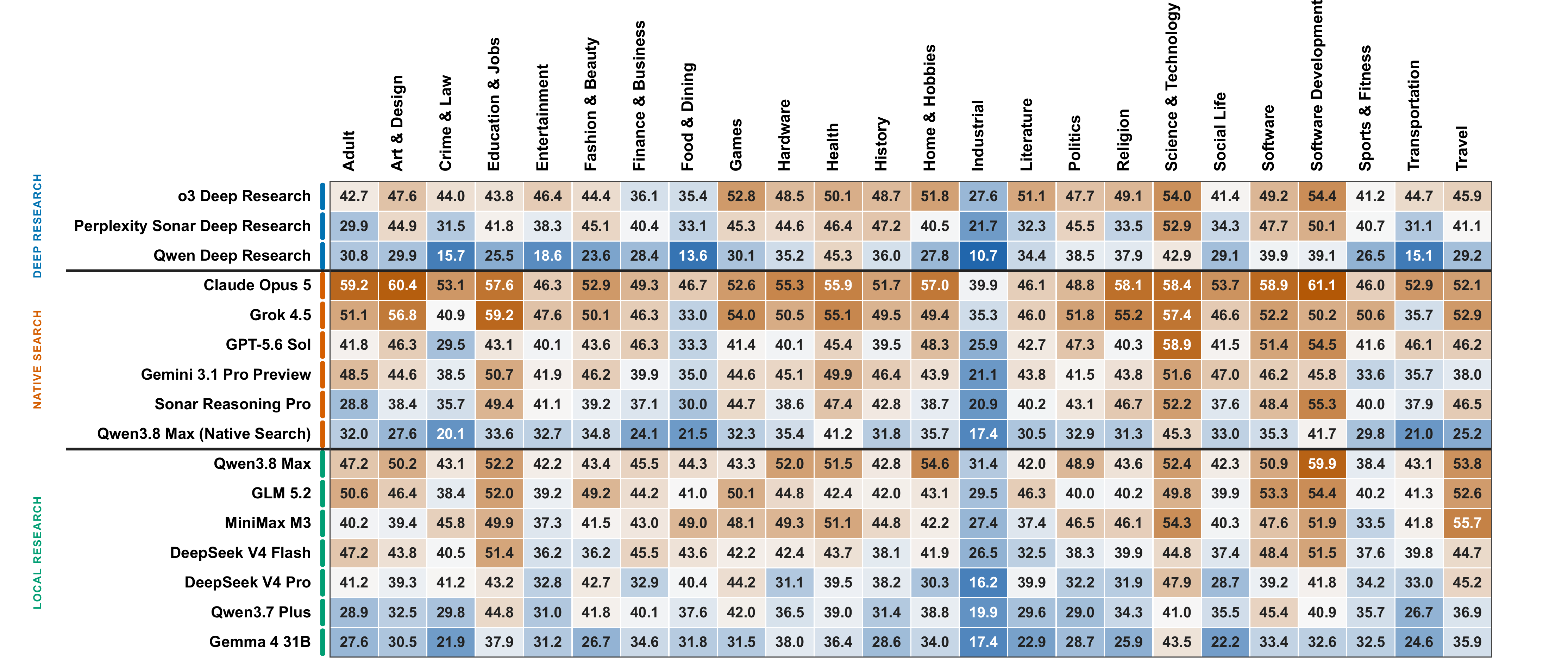}

\caption{\textbf{Breadth across domains.}}
\label{fig:domain-breadth}
\end{figure}

\begin{figure}[H]
\centering
\includegraphics[width=\textwidth]{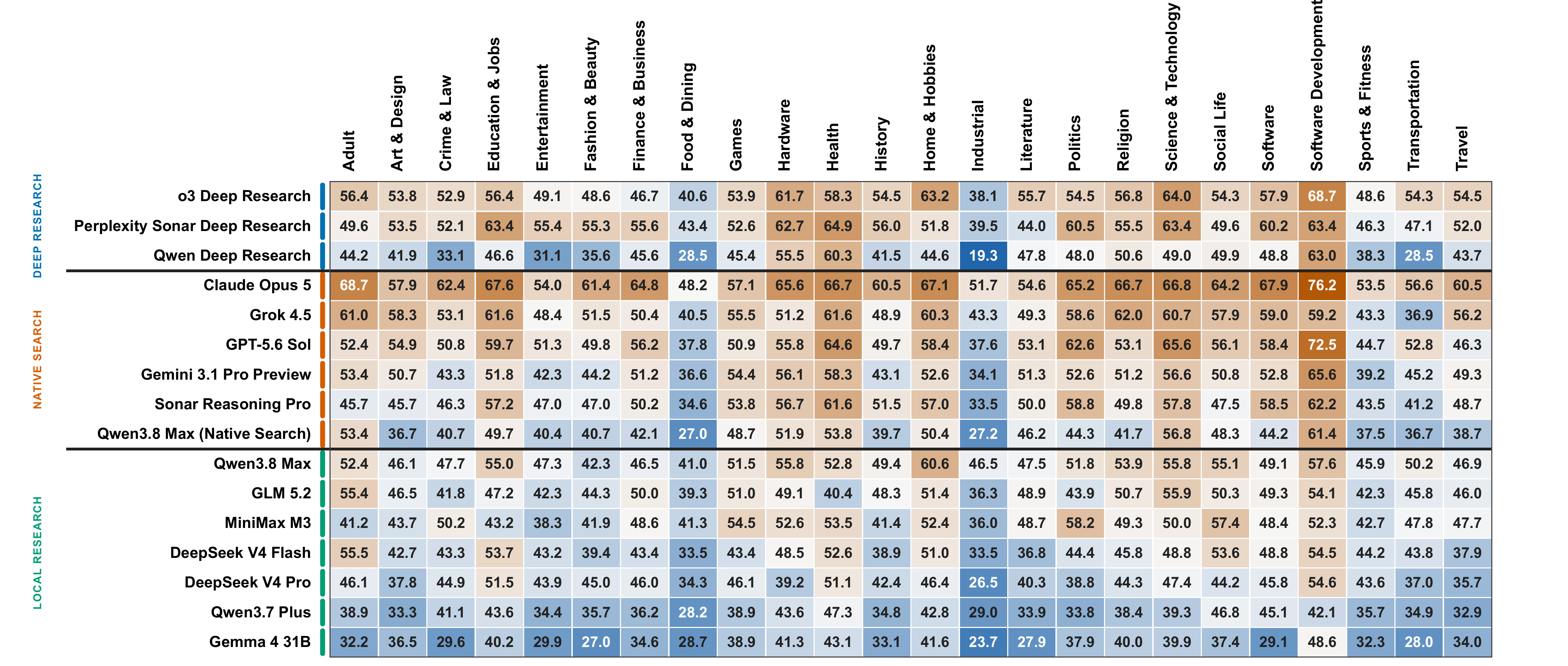}

\caption{\textbf{Depth across domains.}}
\label{fig:domain-depth}
\end{figure}

Figures~\ref{fig:domain-breadth} and~\ref{fig:domain-depth} complement the overall coverage results in Section~\ref{sec:domain-coverage-profiles} and Figure~\ref{fig:domain-overall-coverage} by showing breadth and depth separately across the 24 domains.

\subsection{Research Process and Trajectory Analysis}
\label{app:trajectory-local}

\par
\begin{figure}[H]
\centering
\includegraphics[width=\textwidth]{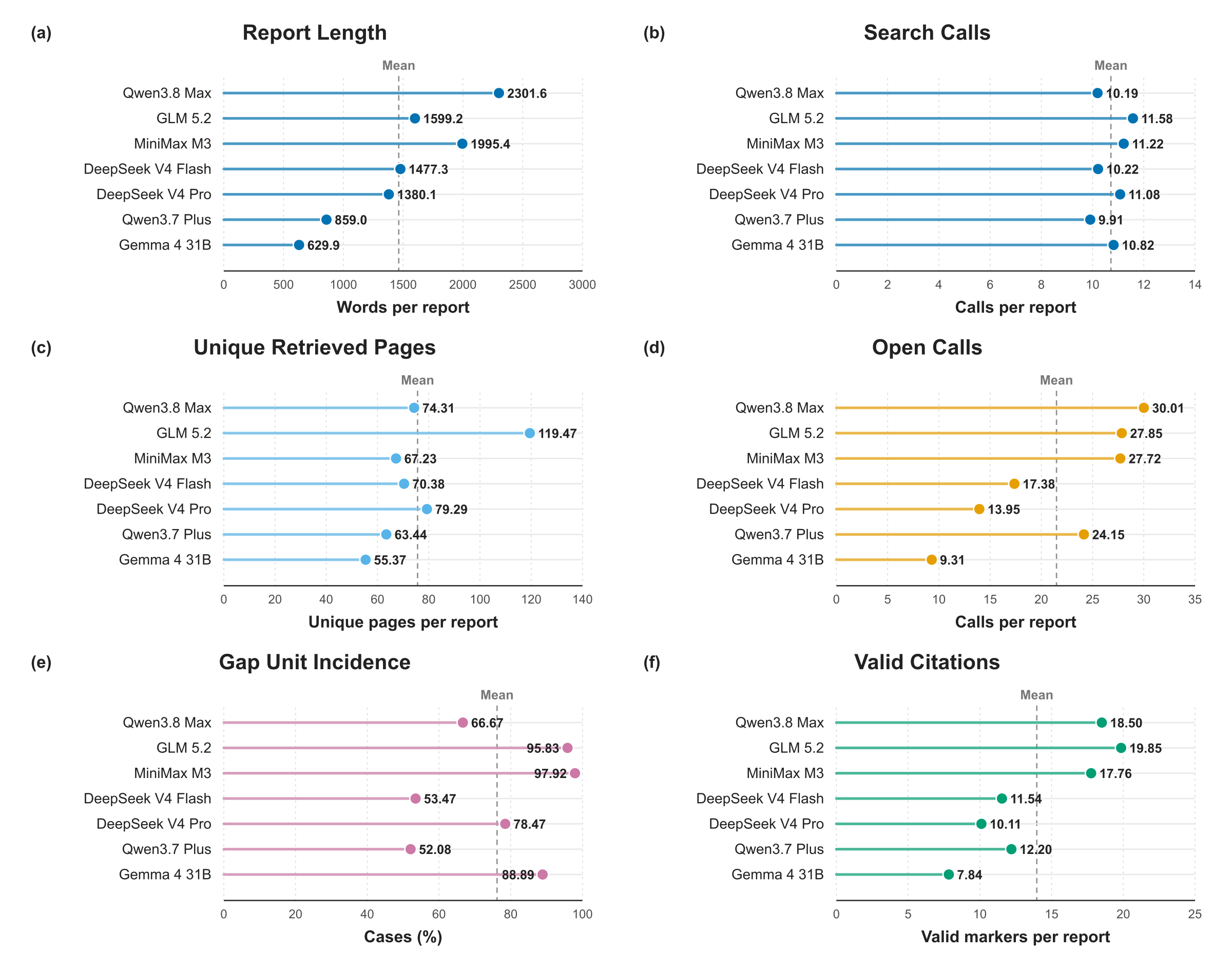}
\caption{Research-process profiles under the local research framework: (a) final-report length, (b) issued Search calls, (c) unique retrieved Wikipedia pages, (d) issued Open calls, (e) Gap Unit incidence, and (f) source-ID-valid citation markers. Dashed lines mark the mean of the seven displayed backbone-LLM means.}
\label{fig:trajectory-local-observables}
\end{figure}

\addvspace{\baselineskip}
\begingroup
\begin{wraptable}{r}{0.52\textwidth}
\centering
\caption{Mean final-report length across the 16 evaluated configurations.}
\label{tab:report-length-all-configurations}
\small
\setlength{\tabcolsep}{6pt}
\begin{tabularx}{\linewidth}{>{\raggedright\arraybackslash}Xr}
\toprule
\textbf{Configuration} & \textbf{Mean words} \\
\midrule
\rowcolor{drsystemshade}\multicolumn{2}{c}{\textbf{Deep Research Systems}} \\
o3 Deep Research & 5,298.0 \\
Perplexity Sonar Deep Research & 8,987.4 \\
Qwen Deep Research & 4,075.5 \\
\midrule
\rowcolor{nativesearchshade}\multicolumn{2}{c}{\textbf{Native Search LLMs}} \\
Claude Opus 5 & 5,273.0 \\
Grok 4.5 & 1,731.5 \\
GPT-5.6 Sol & 4,446.8 \\
Gemini 3.1 Pro Preview & 1,523.0 \\
Sonar Reasoning Pro & 3,048.6 \\
Qwen3.8 Max & 3,090.7 \\
\midrule
\rowcolor{localframeworkshade}\multicolumn{2}{c}{\textbf{\shortstack{Backbone LLMs under the\\local research framework}}} \\
Qwen3.8 Max & 2,301.6 \\
MiniMax M3 & 1,995.4 \\
GLM 5.2 & 1,599.2 \\
DeepSeek V4 Flash & 1,477.3 \\
DeepSeek V4 Pro & 1,380.1 \\
Qwen3.7 Plus & 859.0 \\
Gemma 4 31B & 629.9 \\
\bottomrule
\end{tabularx}
\end{wraptable}
\textbf{Report length across all configurations.} Table~\ref{tab:report-length-all-configurations} summarizes the mean final-report length of all 16 evaluated configurations, each averaged over the same 144 cases.

\textbf{Process profiles under the local research framework.} To examine the research processes behind the final reports, we next focus on the seven backbone LLMs under the local research framework, for which complete trajectories are available under a unified setting.
All seven backbone LLMs share the same Wikipedia snapshot, Search/Open interface, research workflow, and action budget. We analyze the complete $144\times7=1{,}008$ trajectories and jointly report overall coverage, breadth, and depth together with six process observables: final-report length, issued Search calls, unique retrieved Wikipedia pages, issued Open calls, Gap Unit incidence, and source-ID-valid citation markers. Table~\ref{tab:trajectory-local-overview} provides the exact means, and Figure~\ref{fig:trajectory-local-observables} visualizes the process profiles.
\finishwraptable
\endgroup
\par\addvspace{\baselineskip}

Even under this unified setting, the backbones exhibit distinct research processes. Their Search-call counts are similar, but retrieved pages, Open calls, Research Unit usage, valid citations, and final-report length vary substantially. For example, Gap Unit usage is only weakly associated with coverage performance, suggesting that strong performance depends more on effective planning than on simply expanding the number of Research Units.

\begin{appendixtable}[H]{\textwidth}
\small
\caption{\textbf{Coverage and process profiles under the local research framework.} Gap Unit is the percentage of cases that trigger an additional Research Unit.}
\label{tab:trajectory-local-overview}
\resizebox{\textwidth}{!}{%
\begin{tabular}{lccccccccc}
\toprule
 & \multicolumn{3}{c}{\textbf{Coverage (\%)}} & \multicolumn{6}{c}{\textbf{Process observables}} \\
\cmidrule(lr){2-4}\cmidrule(lr){5-10}

\textbf{Backbone} & \textbf{overall $\uparrow$} & \textbf{breadth $\uparrow$} & \textbf{depth $\uparrow$} & \textbf{Report words} & \textbf{Search} & \textbf{Retrieved pages} & \textbf{Open} & \textbf{Gap Unit (\%)} & \textbf{Valid citations} \\
\midrule
Qwen3.8 Max & \textbf{47.94} & \textbf{46.62} & \textbf{50.37} & 2,301.6 & 10.19 & 74.31 & 30.01 & 66.67 & 18.50 \\
GLM 5.2 & 45.34 & 44.61 & 47.09 & 1,599.2 & 11.58 & 119.47 & 27.85 & 95.83 & 19.85 \\
MiniMax M3 & 45.34 & 44.34 & 47.55 & 1,995.4 & 11.22 & 67.23 & 27.72 & 97.92 & 17.76 \\
DeepSeek V4 Flash & 42.63 & 41.42 & 45.04 & 1,477.3 & 10.22 & 70.38 & 17.38 & 53.47 & 11.54 \\
DeepSeek V4 Pro & 39.22 & 36.96 & 43.03 & 1,380.1 & 11.08 & 79.29 & 13.95 & 78.47 & 10.11 \\
Qwen3.7 Plus & 36.10 & 35.37 & 37.94 & 859.0 & 9.91 & 63.44 & 24.15 & 52.08 & 12.20 \\
Gemma 4 31B & 32.00 & 30.43 & 34.82 & 629.9 & 10.82 & 55.37 & 9.31 & 88.89 & 7.84 \\
\bottomrule
\end{tabular}%
}

\end{appendixtable}

\Needspace{4\baselineskip}
\subsection{Sensitivity Analysis under a Common Target Length}
\label{app:target-length-sensitivity}
\begingroup
\paragraph{Motivation.}
More comprehensive knowledge coverage can require more space, as additional substantive facts and relations may satisfy more reference rubrics. ResearchRubrics~\citep{sharma2026researchrubrics} notes this relationship while acknowledging potential verbosity bias; LiveResearchBench~\citep{wang2026liveresearchbench} uses generous output budgets. Imposing identical report lengths adds compression and content-selection constraints, changing the evaluation target. Our primary setting therefore allows systems to develop reports within their configured research budgets. However, to assess sensitivity to output-length constraints, we also examine coverage performance under a common report-length budget. We additionally reran all seven local backbones on 24 cases, one from each of the 24 domains, retaining research settings while targeting 1,200 words.

\begin{appendixtable}[t]{\textwidth}
\caption{Original and target-length results averaged over the same 24 cases. The report-length target is 1,200 words.}
\label{tab:target-length-comparison}
\small
\setlength{\tabcolsep}{6pt}
\begin{tabular}{@{}lrrrr@{}}
\toprule
\textbf{Backbone} & \shortstack{\textbf{Original}\\\textbf{Overall (\%)}} & \shortstack{\textbf{Target-length}\\\textbf{Overall (\%)}} & \shortstack{\textbf{Original}\\\textbf{words}} & \shortstack{\textbf{Target-length}\\\textbf{words}} \\
\midrule
Qwen3.8 Max & 49.11 & 43.31 & 2,251.9 & 1,198.4 \\
MiniMax M3 & 45.13 & 41.94 & 1,899.2 & 1,349.7 \\
GLM 5.2 & 45.16 & 44.30 & 1,622.0 & 1,496.4 \\
DeepSeek V4 Flash & 43.20 & 38.19 & 1,448.0 & 1,224.7 \\
DeepSeek V4 Pro & 36.86 & 38.07 & 1,307.8 & 1,239.0 \\
Qwen3.7 Plus & 35.82 & 33.41 & 878.3 & 1,163.4 \\
Gemma 4 31B & 32.09 & 34.01 & 642.7 & 1,035.0 \\
\bottomrule
\end{tabular}
\end{appendixtable}

\paragraph{Results.}
Table~\ref{tab:target-length-comparison} shows that differences in knowledge coverage persist under a common length target, with some changes in the relative ordering of the models. The same three models remain at the top, although their ordering changes. For example, Qwen3.8 Max and DeepSeek V4 Pro have similar lengths (1,198 vs. 1,239 words) but differ by 5.24 Overall percentage points.

\paragraph{Length adherence.}
Although we explicitly specified a common length target, adherence varied across models: MiniMax and GLM tended to exceed it, whereas Gemma tended to fall short. Nevertheless, average report lengths became substantially closer, providing a more comparable basis for this supplementary evaluation.
\par
\endgroup
\FloatBarrier

\subsection{Retrieval-to-report Coverage and Knowledge Retention}
\label{app:retrieval-report-retention}
\begingroup
Most deep research agents likely follow a multi-stage workflow, retrieving broadly before selecting material for the final report. Knowledge omitted during writing may reflect their own preferences regarding which knowledge to include. We therefore apply our evaluation method to examine different models' breadth and depth preferences during retrieval and writing, together with their retention of retrieved knowledge.

To this end, we evaluated retrieval and report coverage on 24 cases for the seven backbone LLMs evaluated under the local research framework. At the retrieval stage, we collected the retrieved materials and measured the model's knowledge coverage at this stage in terms of Breadth and Depth. At the writing stage, we evaluated the final reports using the same method. Table~\ref{tab:retrieval-report-retention} compares breadth and depth at the retrieval and report stages, and Table~\ref{tab:knowledge-retention-by-hop} reports rubric-level retention overall and by hop distance.

\begin{appendixtable}[H]{\textwidth}
\caption{Retrieval-to-report coverage and final-report performance. Arrows indicate retrieval $\rightarrow$ report; percentages in parentheses denote report-to-retrieval ratios of mean scores. Overall scores refer to the final reports. Models are ordered by their final-report Overall scores in descending order.}
\label{tab:retrieval-report-retention}
\setlength{\tabcolsep}{6pt}
\renewcommand{\arraystretch}{1.08}
\begin{tabular}{@{}lccc@{}}
\toprule
\multicolumn{1}{@{}c}{Model} & \multicolumn{1}{c}{Breadth} & \multicolumn{1}{c}{Depth} & \multicolumn{1}{c@{}}{Overall score} \\
\midrule
Qwen3.8 Max & 49.36 $\rightarrow$ 44.35 (89.8\%) & 54.64 $\rightarrow$ 50.16 (91.8\%) & \textbf{46.73} \\
GLM 5.2 & 55.04 $\rightarrow$ 45.65 (82.9\%) & 57.80 $\rightarrow$ 46.40 (80.3\%) & 45.56 \\
MiniMax M3 & 53.46 $\rightarrow$ 42.77 (80.0\%) & 52.91 $\rightarrow$ 45.68 (86.3\%) & 43.71 \\
DeepSeek V4 Flash & 49.38 $\rightarrow$ 44.39 (89.9\%) & 54.75 $\rightarrow$ 44.20 (80.7\%) & 43.61 \\
DeepSeek V4 Pro & 43.33 $\rightarrow$ 34.85 (80.4\%) & 47.59 $\rightarrow$ 43.99 (92.4\%) & 38.54 \\
Qwen3.7 Plus & 52.87 $\rightarrow$ 33.75 (63.8\%) & 53.65 $\rightarrow$ 37.69 (70.2\%) & 35.23 \\
Gemma 4 31B & 43.66 $\rightarrow$ 30.04 (68.8\%) & 42.18 $\rightarrow$ 34.18 (81.0\%) & 31.53 \\
\bottomrule
\end{tabular}
\end{appendixtable}

\begin{appendixtable}[H]{\textwidth}
\caption{Knowledge retention overall and by hop distance. Retention measures the proportion of fully covered retrieval rubrics retained in the final reports.}
\label{tab:knowledge-retention-by-hop}
\setlength{\tabcolsep}{8pt}
\renewcommand{\arraystretch}{1.08}
\begin{tabular}{@{}lcccc@{}}
\toprule
\multicolumn{1}{@{}c}{\multirow{2}{*}{Model}} & \multicolumn{4}{c@{}}{Knowledge retention (\%)} \\
\cmidrule(lr){2-5}
 & \multicolumn{1}{c}{All} & \multicolumn{1}{c}{1-hop} & \multicolumn{1}{c}{2-hop} & \multicolumn{1}{c@{}}{3-hop} \\
\midrule
Qwen3.8 Max & 65.6 & 88.1 & 63.0 & 45.8 \\
GLM 5.2 & 52.0 & 84.1 & 49.6 & 30.1 \\
MiniMax M3 & 56.7 & 79.0 & 55.2 & 35.4 \\
DeepSeek V4 Flash & 56.5 & 89.8 & 49.6 & 32.8 \\
DeepSeek V4 Pro & 53.7 & 74.7 & 52.0 & 34.0 \\
Qwen3.7 Plus & 36.4 & 66.3 & 28.8 & 16.0 \\
Gemma 4 31B & 50.7 & 66.5 & 48.4 & 22.9 \\
\bottomrule
\end{tabular}
\end{appendixtable}

Overall, 47.0\% of fully covered retrieval rubric instances are no longer fully covered in reports, and knowledge retention decreases as hop distance increases. Models also differ in what they preserve: DeepSeek Pro preserves more of its Depth than Breadth score (92.4\% vs. 80.4\%), suggesting more focused coverage along selected chains. Flash better preserves Breadth (89.9\% vs. 80.7\% for Depth), consistent with retaining topical range while compressing deeper content. Qwen3.8 Max maintains high score retention in both dimensions (89.8\%/91.8\%) and retains the most retrieved knowledge (65.6\%), showing stronger coverage preservation during synthesis.

Evaluating retrieval and reporting separately distinguishes retrieval gaps from omissions of already retrieved knowledge in the final report. This offers valuable guidance for improving retrieval and knowledge retention during synthesis.
\par
\endgroup
\FloatBarrier

\subsection{Supplementary Evaluation of Open Deep Research}
\label{app:odr-supplementary-evaluation}
\begingroup
We additionally evaluate an open-source deep research harness to broaden the range of research systems evaluated by our benchmark. We evaluate LangChain Open Deep Research (ODR) with DeepSeek V4 Flash, using Tavily for web search, on 24 sampled cases. Table~\ref{tab:odr-original16-matched24} reports the scores of the original 16 agents on the same 24 cases.

As shown in Table~\ref{tab:odr-supplementary-evaluation}, ODR achieves Overall and Depth scores 11.67 and 16.81 percentage points higher than the local configuration, respectively, also exceeding o3 Deep Research on this subset. Its larger exploration budget, iterative supervision, and evidence synthesis may contribute to these gains. These results illustrate the value of evaluating complete research configurations alongside backbone LLMs.  Broader evaluation across additional harnesses and backbone models under matched research budgets remains an important direction for future work.

\begin{appendixtable}[H]{\textwidth}
\caption{Supplementary evaluation of LangChain Open Deep Research. Both ODR and the local research framework use DeepSeek V4 Flash. Results and research activity are averaged over the same 24 cases. Coverage scores are reported as percentages. \textsuperscript{\(\dagger\)} Model-call records use configuration-specific counting conventions; approximately 92\% of ODR responses are used for webpage summarization. ``---'' indicates unavailable internal counts.}
\label{tab:odr-supplementary-evaluation}
\setlength{\tabcolsep}{4pt}
\begin{tabularx}{\linewidth}{@{}>{\raggedright\arraybackslash}Xrrrrrr@{}}
\toprule
\multirow{2}{*}{Configuration} & \multicolumn{1}{c}{\multirow{2}{*}{Overall}} & \multicolumn{1}{c}{\multirow{2}{*}{Breadth}} & \multicolumn{1}{c}{\multirow{2}{*}{Depth}} & \multicolumn{1}{c}{Successful} & \multicolumn{1}{c}{Model-call} & \multicolumn{1}{c}{Report} \\
 & & & & \multicolumn{1}{c}{search queries} & \multicolumn{1}{c}{records\textsuperscript{\(\dagger\)}} & \multicolumn{1}{c}{words} \\
\midrule
LangChain ODR & 57.11 & 51.98 & 63.29 & 133.67 & 693.29 & 7,130 \\
Local research framework & 45.43 & 45.49 & 46.48 & 9.88 & 28.04 & 1,437 \\
o3 Deep Research & 52.15 & 47.54 & 57.78 & --- & --- & 5,344 \\
\bottomrule
\end{tabularx}
\end{appendixtable}
\FloatBarrier

\begin{appendixtable}[H]{\textwidth}
\begingroup
\caption{Results for the original 16 agents on the same 24 cases used in Table~\ref{tab:odr-supplementary-evaluation}. All values are percentages.}
\label{tab:odr-original16-matched24}
\small
\setlength{\tabcolsep}{6pt}
\begin{tabularx}{\linewidth}{>{\raggedright\arraybackslash}Xccc}
\toprule
\textbf{Deep Research Agent} & \textbf{Overall Coverage} & \textbf{Breadth} & \textbf{Depth} \\
\midrule
\rowcolor{drsystemshade}\multicolumn{4}{c}{\textbf{(1) Deep Research Systems}} \\
o3 Deep Research & 52.15 & 47.54 & 57.78 \\
Perplexity Sonar Deep Research & 47.14 & 41.19 & 54.54 \\
Qwen Deep Research & 38.41 & 32.01 & 47.09 \\
\midrule
\rowcolor{nativesearchshade}\multicolumn{4}{c}{\textbf{(2) Native Search LLMs}} \\
Claude Opus 5 & \textbf{58.70} & 54.84 & \textbf{63.21} \\
Grok 4.5 & 56.82 & \textbf{54.94} & 59.11 \\
GPT-5.6 Sol & 51.14 & 46.04 & 57.77 \\
Gemini 3.1 Pro Preview & 45.90 & 41.54 & 51.85 \\
Sonar Reasoning Pro & 46.44 & 42.20 & 52.40 \\
Qwen3.8 Max & 36.36 & 30.74 & 44.03 \\
\midrule
\rowcolor{localframeworkshade}\multicolumn{4}{c}{\textbf{(3) Backbone LLMs under the local research framework}} \\
Qwen3.8 Max & 49.66 & 48.21 & 52.45 \\
MiniMax M3 & 46.15 & 44.50 & 49.21 \\
GLM 5.2 & 45.32 & 43.01 & 48.52 \\
DeepSeek V4 Flash & 45.43 & 45.49 & 46.48 \\
DeepSeek V4 Pro & 38.62 & 35.28 & 43.86 \\
Qwen3.7 Plus & 35.78 & 33.99 & 38.66 \\
Gemma 4 31B & 33.32 & 31.65 & 36.07 \\
\bottomrule
\end{tabularx}
\endgroup
\end{appendixtable}
\FloatBarrier
\par

\endgroup
\FloatBarrier

\section{Case Study}
\label{app:case-studies}

We use three diagnostic cases to connect aggregate scores to concrete report behavior. The first requires similar overall coverage with a crossed breadth--depth profile; the second examines reports that follow largely different chains of related knowledge within the same research space; and the third is restricted to the unified local research framework and contrasts higher research activity with lower coverage performance.

\subsection{Similar Overall Coverage Can Hide Different Research Strategies}
\label{app:case-equal-overall}

\paragraph{Query.}
\begin{quote}
\footnotesize\itshape
I'm trying to understand the career of the British entertainment mogul Lew Grade, particularly his transition from television into big-budget international film production during the 1970s and early 1980s. I'd like a well-researched overview that explains how his instincts and financing strategies produced both remarkable successes and costly failures, and how these outcomes shaped his reputation and financial standing over time. Rather than just listing his projects, I'm interested in the bigger picture: why he pursued ambitious, prestige, and family-oriented entertainment for global audiences, how his partnerships with major directors and creative figures worked in practice, and what balance of artistic risk and commercial calculation drove his decisions. I'd also like to understand how a combination of overreach and shifting business pressures eventually undermined his control of his own company and led to a corporate takeover. Please connect his personal management style to the broader changes in how film and television were financed and distributed during his era, and assess how his legacy has been judged.
\end{quote}

\begin{figure}[H]
\centering
\includegraphics[width=\textwidth]{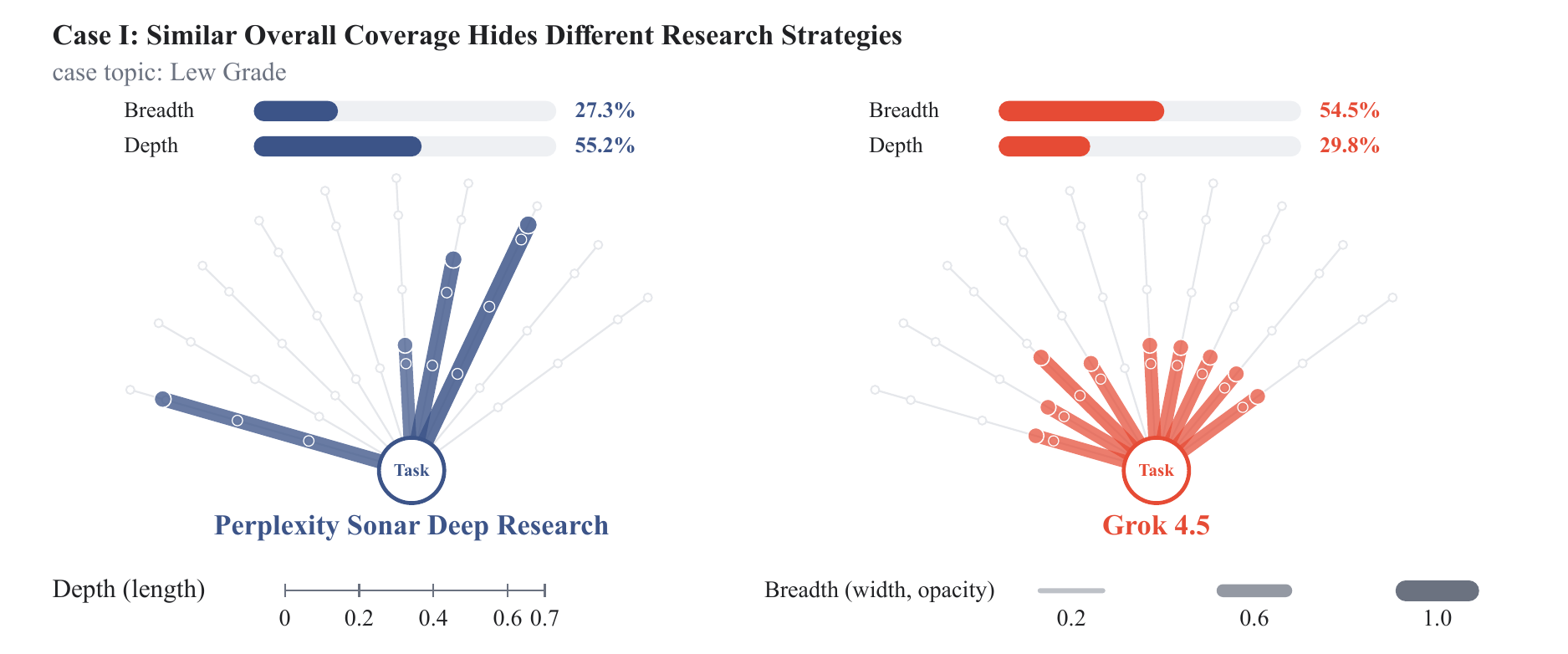}

\caption{\textbf{Similar overall coverage can conceal different research strategies.} Bars show aggregate breadth and depth (\%). Gray rays mark ten shared research directions. Colored ray width and opacity encode branch breadth, while ray length encodes branch depth.}
\label{fig:case-equal-overall}
\end{figure}

Perplexity Sonar Deep Research and Grok 4.5 obtain similar overall coverage on this case (38.83 vs.\ 40.31), despite exhibiting sharply inverted coverage profiles. Perplexity Sonar Deep Research covers major research directions less comprehensively but follows chains of related knowledge farther within selected directions (breadth 27.33, depth 55.19). By contrast, Grok 4.5 exhibits the opposite pattern (breadth 54.48, depth 29.82). Figure~\ref{fig:case-equal-overall} shows how these contrasting profiles can yield similar overall coverage scores.

This case illustrates a general limitation of scalar report evaluation: similar aggregate performance does not imply similar research behavior. One report covers major research directions more comprehensively, while the other follows chains of related knowledge farther within selected directions.

\begin{figure}[H]
\centering
\includegraphics[width=\textwidth]{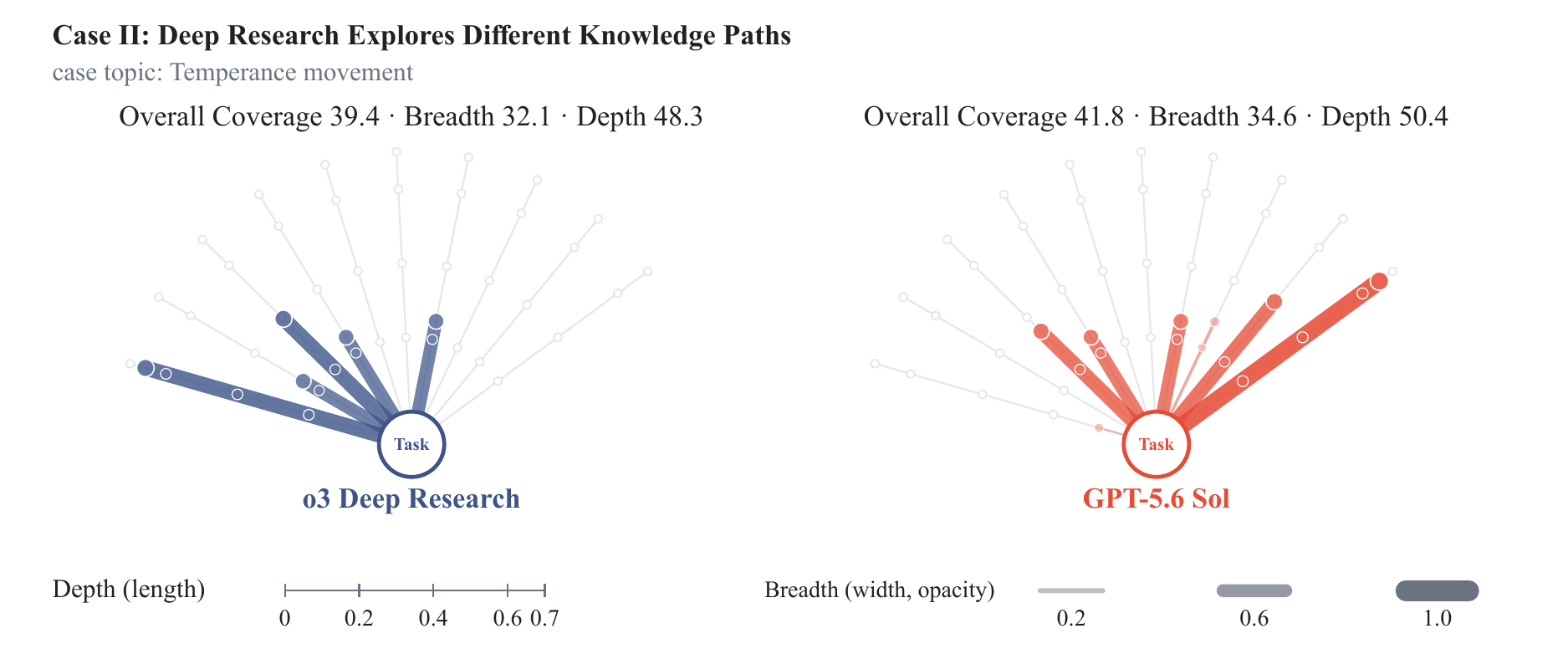}

\caption{\textbf{Deep research can explore different knowledge paths.} Values above the models report overall coverage, breadth, and depth (\%). Matched gray rays mark the same research directions. Colored ray width and opacity encode branch breadth, while ray length encodes branch depth.}
\label{fig:case-route-diversity}
\end{figure}

\subsection{Deep Research Can Explore Different Knowledge Paths}
\label{app:case-route-diversity}

\paragraph{Query.}
\begin{quote}
\footnotesize\itshape
I'm trying to understand the temperance movement not just as a campaign against alcohol, but as a broad historical phenomenon that grew out of specific social conditions and then spread and changed in surprising ways. I'd like a research-style report that traces how early alarm over heavy drinking in one society, combined with new medical and moral arguments, laid the groundwork for organized reform. I'm especially curious about how different religious traditions shaped competing versions of the movement, giving it distinct doctrinal flavors and leaders, and how this influenced its appeal across communities. Beyond its origins in Britain and America, I'd like to understand how the movement traveled to colonial and settler societies, where it became entangled with other causes like women's political rights and, strikingly, with anti-colonial and independence struggles and religious revivals. Please explain how these motivations interacted, why the same idea took on such different meanings in different places, and what tensions or contradictions emerged as it evolved, without simply listing every organization or figure involved.
\end{quote}

Reports on the same question may emphasize different directions and knowledge chains even when their aggregate scores are similar. On this case, o3 Deep Research and GPT-5.6 Sol achieve breadth scores of 32.11 and 34.64, depth scores of 48.26 and 50.43, and overall coverage scores of 39.37 and 41.79, respectively. Despite these similar aggregate results, their branch-level coverage profiles differ substantially. Figure~\ref{fig:case-route-diversity} summarizes both reports across the same ten research directions, making their differing emphases visible.

This case illustrates how reports on the same question can emphasize different parts of the underlying knowledge space. Branch-level analysis makes these differences visible by showing each report's progression along knowledge chains within individual directions.

\subsection{More Research Activity Does Not Guarantee Higher Coverage}
\label{app:case-activity-coverage}

\paragraph{Query.}
\begin{quote}
\footnotesize\itshape

I'd like a well-researched cultural history of Horn \& Hardart, the automat chain, that goes beyond describing how the vending-style restaurants worked and instead examines why this company became such a durable fixture in American popular imagination. I'm especially curious about how the company built its public presence through sponsorship of broadcast entertainment in the early days of radio, including how those programs helped launch performers who later became famous, and how that kind of promotional and media involvement shaped its brand. Second, I'd like to understand why writers, songwriters, filmmakers, and stage productions kept referencing or recreating its dining rooms across decades and what the automat came to symbolize about class, city life, and nostalgia in songs, comedies, and films. Finally, please connect its rise and eventual decline to broader shifts in urban dining, entertainment, and changing tastes, so the report reads as an interpretive account of its lasting legacy rather than just a timeline of locations and menu items.
\end{quote}

\Needspace{18\baselineskip}
Table~\ref{tab:case-activity-horn-hardart} summarizes the research activity, report length, and coverage scores of GLM 5.2 and DeepSeek V4 Flash on this case. Under the same frozen corpus, tool interface, workflow, and action budgets, GLM 5.2 makes more Search and Open calls, retrieves more pages, and uses more tokens, with part of its effort devoted to revisiting shared sources and expanding contextual discussion. Its report is also longer, yet DeepSeek V4 Flash achieves higher breadth, depth, and overall coverage. This case shows that greater research activity and longer reports do not necessarily imply higher knowledge coverage, highlighting the importance of effective planning and evidence selection.

\begin{appendixtable}[H]{\textwidth}
\centering
\footnotesize
\setlength{\tabcolsep}{4pt}
\caption{\textbf{Research activity and coverage on Horn \& Hardart S03.} Pages denotes unique Wikipedia pages returned during research.}
\label{tab:case-activity-horn-hardart}
\begin{tabular}{@{}lcccccccc@{}}
\toprule

Agent & Search & Open & Pages & Tokens (k) & Words & \shortstack{breadth\\(\%)} & \shortstack{depth\\(\%)} & \shortstack{overall\\coverage (\%)} \\
\midrule
GLM 5.2 & 12 & 34 & 112 & 302.9 & 2,214 & 26.40 & 32.28 & 29.19 \\
DeepSeek V4 Flash & 9 & 28 & 43 & 150.4 & 1,372 & 45.42 & 45.65 & 45.54 \\
\bottomrule
\end{tabular}
\end{appendixtable}

\FloatBarrier
\Needspace{8\baselineskip}
\begingroup
\subsection{Example Showcase}
\label{app:example-showcase}
We present a report generated by o3 Deep Research for the query below, together with a representative subset of rubrics drawn from the corresponding local subgraph for analysis. We first show the query and the selected portion of the rubric tree (Figure~\ref{fig:example-tree}), followed by the rubric texts and their satisfaction status (Table~\ref{tab:example-rubrics}). We then present the full report with highlighted evidence and analyze its coverage of the selected rubrics.

\paragraph{Query.}
\begin{quote}
\footnotesize\itshape
I'm trying to understand the origins of the roguelike video game genre and would love a well-researched history that goes beyond just listing famous titles. I'm especially curious about how the genre actually came together in the late 1970s and early 1980s: what earlier interactive and exploration-style games inspired its first creators, how the hardware and display limitations of the era (things like the text-only machines and character-based screens that early developers had to work with) shaped its distinctive look and design conventions, and how tabletop role-playing traditions fed into its dungeon-crawling and narrative structure. I'd also like to understand who the key people were behind its formative early works and how the personal-computer platforms and distribution methods of that time affected which games reached players. Finally, please explain how these different technical constraints, cultural influences, and design choices interacted to define the genre's core identity, and why certain defining features became so persistent even as computing itself changed dramatically afterward.
\end{quote}
\endgroup
\newpage

\begingroup
\setlength{\intextsep}{6pt}
\begin{figure}[H]
\centering
\includegraphics[width=0.64\linewidth]{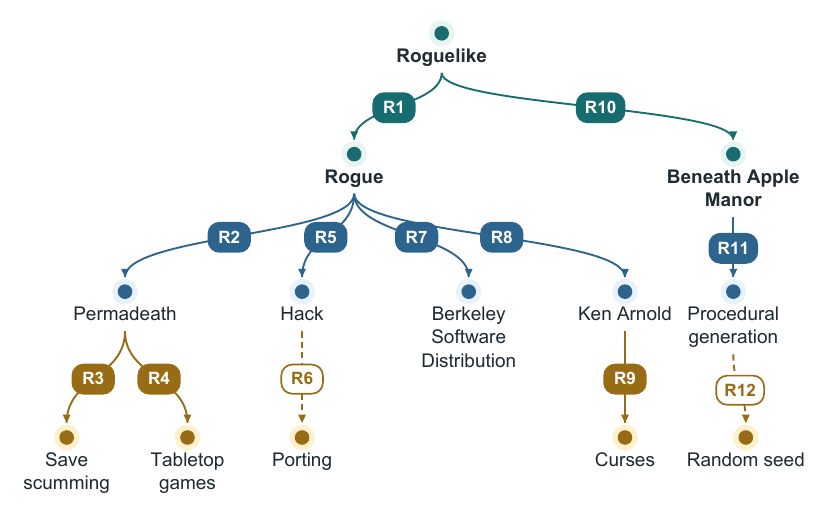}
\caption{Representative rubric relations from the local subgraph with ``Roguelike'' as the root entity.}
\label{fig:example-tree}
\end{figure}
\begin{table}[H]
\centering
\hypersetup{linkcolor=blue}
\caption{Selected rubrics and their satisfaction status in the generated report.}
\label{tab:example-rubrics}
\small
\setlength{\tabcolsep}{3.2pt}
\renewcommand{\arraystretch}{1.0}
\begin{tabularx}{\linewidth}{p{.31in}X>{\centering\arraybackslash}p{.55in}}
\toprule
\textbf{ID} & \textbf{Relation and Rubric Text} & \textbf{Satisfied} \\
\midrule
\rowcolor{exTableStripe}
\hypertarget{ex:rub:R1}{}\textbf{\hyperlink{ex:ev:6}{R1}} & \textbf{Roguelike \ExArrow{} Rogue}\newline The report explicitly states or clearly paraphrases that the 1980 game Rogue is considered the forerunner and namesake of the roguelike genre. & \ding{51} \\[2.5pt]
\hypertarget{ex:rub:R2}{}\textbf{\hyperlink{ex:ev:3}{R2}} & \textbf{Rogue \ExArrow{} Permadeath}\newline The report explicitly states or clearly paraphrases that Rogue implements permadeath, meaning a character that loses all health is dead. & \ding{51} \\[2.5pt]
\rowcolor{exTableStripe}
\hypertarget{ex:rub:R3}{}\textbf{\hyperlink{ex:ev:4}{R3}} & \textbf{Permadeath \ExArrow{} Save scumming}\newline The report explicitly states or clearly paraphrases that players may circumvent permadeath by backing up and restoring savefiles, a tactic called 'save scumming' that is considered cheating. & \ding{51} \\[2.5pt]
\hypertarget{ex:rub:R4}{}\textbf{\hyperlink{ex:ev:5}{R4}} & \textbf{Permadeath \ExArrow{} Tabletop games}\newline The report explicitly states or clearly paraphrases that permadeath is a mechanic used in tabletop games as well as video games. & \ding{51} \\[2.5pt]
\rowcolor{exTableStripe}
\hypertarget{ex:rub:R5}{}\textbf{\hyperlink{ex:ev:16}{R5}} & \textbf{Rogue \ExArrow{} Hack}\newline The report explicitly states or clearly paraphrases that Rogue inspired the development of Hack. & \ding{51} \\[2.5pt]
\hypertarget{ex:rub:R6}{}\textbf{R6} & \textbf{Hack \ExArrow{} Porting}\newline The report explicitly states or clearly paraphrases that Hack was ported to other platforms, with some later ports using graphical tiles instead of ASCII characters. & \ding{55} \\[2.5pt]
\rowcolor{exTableStripe}
\hypertarget{ex:rub:R7}{}\textbf{\hyperlink{ex:ev:10}{R7}} & \textbf{Rogue \ExArrow{} Berkeley Software Distribution}\newline The report explicitly states or clearly paraphrases that Rogue became popular in part due to its inclusion in the Berkeley Software Distribution (BSD). & \ding{51} \\[2.5pt]
\hypertarget{ex:rub:R8}{}\textbf{\hyperlink{ex:ev:8}{R8}} & \textbf{Rogue \ExArrow{} Ken Arnold}\newline The report explicitly states or clearly paraphrases that Ken Arnold made later contributions to Rogue. & \ding{51} \\[2.5pt]
\rowcolor{exTableStripe}
\hypertarget{ex:rub:R9}{}\textbf{\hyperlink{ex:ev:1}{R9}} & \textbf{Ken Arnold \ExArrow{} Curses}\newline The report explicitly states or clearly paraphrases that Ken Arnold wrote the first curses library, released with BSD UNIX and used for games such as Rogue. & \ding{51} \\[2.5pt]
\hypertarget{ex:rub:R10}{}\textbf{\hyperlink{ex:ev:15}{R10}} & \textbf{Roguelike \ExArrow{} Beneath Apple Manor}\newline The report explicitly states or clearly paraphrases that Beneath Apple Manor predates Rogue within the context of the roguelike genre's history. & \ding{51} \\[2.5pt]
\rowcolor{exTableStripe}
\hypertarget{ex:rub:R11}{}\textbf{\hyperlink{ex:ev:2}{R11}} & \textbf{Beneath Apple Manor \ExArrow{} Procedural generation}\newline The report explicitly states or clearly paraphrases that Beneath Apple Manor was one of the first video games to use procedural generation. & \ding{51} \\[2.5pt]
\hypertarget{ex:rub:R12}{}\textbf{R12} & \textbf{Procedural generation \ExArrow{} Random seed}\newline The report explicitly states or clearly paraphrases that procedural generation used predefined seed values (map seeds) to generate very large game worlds. & \ding{55} \\[2.5pt]
\bottomrule
\end{tabularx}

\end{table}
\endgroup
\clearpage
\begingroup\setlength{\ExReportUnit}{1bp}
\centering
\begin{tikzpicture}[x=\ExReportUnit,y=-\ExReportUnit]
\node[anchor=north west,inner sep=0,outer sep=0] at (0,0){\includegraphics[page=1,width=396bp]{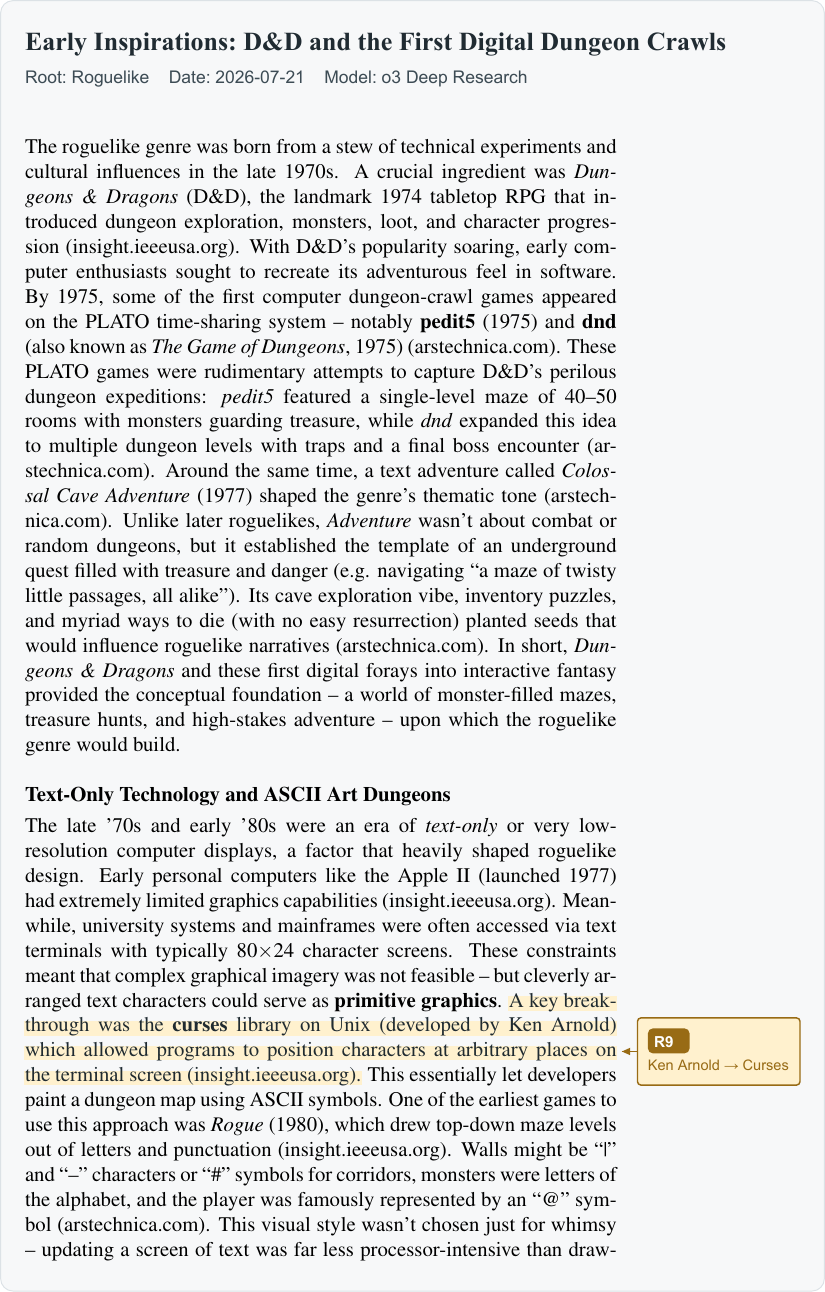}};
\node[inner sep=0,outer sep=0] at (12.0000,470.8273){\hypertarget{ex:ev:1}{}};
\node[anchor=north west,inner sep=0,outer sep=0] at (33.87700,113.98801){\href{https://insight.ieeeusa.org/articles/going-rogue-a-brief-history-of-the-computerized-dungeon-crawl/\#:~:text=Possibly\%20more\%20than\%20any\%20other,Players\%20could\%20purchase}{\phantom{\rule{76.52200\ExReportUnit}{10.90399\ExReportUnit}}}};
\node[anchor=north west,inner sep=0,outer sep=0] at (198.00000,161.80801){\href{https://arstechnica.com/gaming/2020/03/ascii-art-permadeath-the-history-of-roguelike-games/\#:~:text=Although\%20they\%20had\%20no\%20direct,boss\%20fight\%20at\%20the\%20end}{\phantom{\rule{66.45001\ExReportUnit}{10.78497\ExReportUnit}}}};
\node[anchor=north west,inner sep=0,outer sep=0] at (284.14499,209.62900){\href{https://arstechnica.com/gaming/2020/03/ascii-art-permadeath-the-history-of-roguelike-games/\#:~:text=Although\%20they\%20had\%20no\%20direct,boss\%20fight\%20at\%20the\%20end}{\phantom{\rule{12.85101\ExReportUnit}{10.90399\ExReportUnit}}}};
\node[anchor=north west,inner sep=0,outer sep=0] at (11.00400,221.58401){\href{https://arstechnica.com/gaming/2020/03/ascii-art-permadeath-the-history-of-roguelike-games/\#:~:text=Although\%20they\%20had\%20no\%20direct,boss\%20fight\%20at\%20the\%20end}{\phantom{\rule{58.70900\ExReportUnit}{10.53000\ExReportUnit}}}};
\node[anchor=north west,inner sep=0,outer sep=0] at (263.47198,233.53900){\href{https://arstechnica.com/gaming/2020/03/ascii-art-permadeath-the-history-of-roguelike-games/\#:~:text=Will\%20Crowther\%20and\%20Don\%20Woods\%E2\%80\%99,winning\%20players\%20to\%20start\%20again}{\phantom{\rule{33.52402\ExReportUnit}{10.90399\ExReportUnit}}}};
\node[anchor=north west,inner sep=0,outer sep=0] at (11.00400,245.49399){\href{https://arstechnica.com/gaming/2020/03/ascii-art-permadeath-the-history-of-roguelike-games/\#:~:text=Will\%20Crowther\%20and\%20Don\%20Woods\%E2\%80\%99,winning\%20players\%20to\%20start\%20again}{\phantom{\rule{38.23600\ExReportUnit}{10.90402\ExReportUnit}}}};
\node[anchor=north west,inner sep=0,outer sep=0] at (163.52000,305.26999){\href{https://arstechnica.com/gaming/2020/03/ascii-art-permadeath-the-history-of-roguelike-games/\#:~:text=Will\%20Crowther\%20and\%20Don\%20Woods\%E2\%80\%99,winning\%20players\%20to\%20start\%20again}{\phantom{\rule{66.45000\ExReportUnit}{10.90402\ExReportUnit}}}};
\node[anchor=north west,inner sep=0,outer sep=0] at (185.74001,427.81100){\href{https://insight.ieeeusa.org/articles/going-rogue-a-brief-history-of-the-computerized-dungeon-crawl/\#:~:text=Beneath\%20Apple\%20Manor\%E2\%80\%99s\%20initial\%20graphics,by\%20Michael\%20Toy\%20and\%20Glenn}{\phantom{\rule{76.52299\ExReportUnit}{10.90399\ExReportUnit}}}};
\node[anchor=north west,inner sep=0,outer sep=0] at (92.13900,511.49701){\href{https://insight.ieeeusa.org/articles/going-rogue-a-brief-history-of-the-computerized-dungeon-crawl/\#:~:text=emulate\%20PLATO\%E2\%80\%99s\%20graphical\%20capabilities\%2C\%20so,Adventure\%2C\%20Rogue\%20has\%20the\%20player}{\phantom{\rule{76.52200\ExReportUnit}{10.90399\ExReportUnit}}}};
\node[anchor=north west,inner sep=0,outer sep=0] at (135.66499,547.36200){\href{https://insight.ieeeusa.org/articles/going-rogue-a-brief-history-of-the-computerized-dungeon-crawl/\#:~:text=applications\%20was\%20gaming\%2C\%20and\%20Rogue,and\%20contributed\%20to\%20its\%20development}{\phantom{\rule{76.52200\ExReportUnit}{10.90399\ExReportUnit}}}};
\node[anchor=north west,inner sep=0,outer sep=0] at (29.80100,583.22803){\href{https://arstechnica.com/gaming/2020/03/ascii-art-permadeath-the-history-of-roguelike-games/\#:~:text=They\%20distributed\%20executable\%20versions\%20of,else\%E2\%80\%94you\%20used\%20up\%20a\%20turn}{\phantom{\rule{66.45000\ExReportUnit}{10.90399\ExReportUnit}}}};
\node[anchor=north west,inner sep=0,outer sep=0] at (311.00000,493.61855){\hyperlink{ex:rub:R9}{\phantom{\rule{20.00000\ExReportUnit}{12.00000\ExReportUnit}}}};
\end{tikzpicture}\par\endgroup\clearpage
\begingroup\setlength{\ExReportUnit}{1bp}
\centering
\begin{tikzpicture}[x=\ExReportUnit,y=-\ExReportUnit]
\node[anchor=north west,inner sep=0,outer sep=0] at (0,0){\includegraphics[page=2,width=396bp]{figures/appendix_roguelike_report.pdf}};
\node[inner sep=0,outer sep=0] at (12.0000,368.8969){\hypertarget{ex:ev:2}{}};
\node[anchor=north west,inner sep=0,outer sep=0] at (14.32100,30.05701){\href{https://insight.ieeeusa.org/articles/going-rogue-a-brief-history-of-the-computerized-dungeon-crawl/\#:~:text=Directly\%20inspired\%20by\%20Dungeons\%20and,and\%20contributed\%20to\%20its\%20development}{\phantom{\rule{76.52200\ExReportUnit}{11.01398\ExReportUnit}}}};
\node[anchor=north west,inner sep=0,outer sep=0] at (33.64700,77.98798){\href{https://www.killscreen.com/brief-history-roguelike/\#:~:text=Thirty\%20years\%20the\%20senior\%20of,how\%20avid\%20the\%20fanbase\%20is}{\phantom{\rule{83.79500\ExReportUnit}{10.90405\ExReportUnit}}}};
\node[anchor=north west,inner sep=0,outer sep=0] at (177.49300,203.51700){\href{https://insight.ieeeusa.org/articles/going-rogue-a-brief-history-of-the-computerized-dungeon-crawl/\#:~:text=applications\%20was\%20gaming\%2C\%20and\%20Rogue,and\%20contributed\%20to\%20its\%20development}{\phantom{\rule{77.44901\ExReportUnit}{10.78400\ExReportUnit}}}};
\node[anchor=north west,inner sep=0,outer sep=0] at (262.60599,203.51700){\href{https://arstechnica.com/gaming/2020/03/ascii-art-permadeath-the-history-of-roguelike-games/\#:~:text=They\%20distributed\%20executable\%20versions\%20of,else\%E2\%80\%94you\%20used\%20up\%20a\%20turn}{\phantom{\rule{34.39001\ExReportUnit}{10.78400\ExReportUnit}}}};
\node[anchor=north west,inner sep=0,outer sep=0] at (11.00400,215.47198){\href{https://arstechnica.com/gaming/2020/03/ascii-art-permadeath-the-history-of-roguelike-games/\#:~:text=They\%20distributed\%20executable\%20versions\%20of,else\%E2\%80\%94you\%20used\%20up\%20a\%20turn}{\phantom{\rule{38.23600\ExReportUnit}{10.49600\ExReportUnit}}}};
\node[anchor=north west,inner sep=0,outer sep=0] at (263.47198,290.19199){\href{https://arstechnica.com/gaming/2020/03/ascii-art-permadeath-the-history-of-roguelike-games/\#:~:text=with\%20riches\%20but\%20also\%20with,winning\%20players\%20to\%20start\%20again}{\phantom{\rule{33.52402\ExReportUnit}{10.90402\ExReportUnit}}}};
\node[anchor=north west,inner sep=0,outer sep=0] at (11.00400,302.03699){\href{https://arstechnica.com/gaming/2020/03/ascii-art-permadeath-the-history-of-roguelike-games/\#:~:text=with\%20riches\%20but\%20also\%20with,winning\%20players\%20to\%20start\%20again}{\phantom{\rule{38.23600\ExReportUnit}{10.88000\ExReportUnit}}}};
\node[anchor=north west,inner sep=0,outer sep=0] at (14.32100,421.69901){\href{https://arstechnica.com/gaming/2020/03/ascii-art-permadeath-the-history-of-roguelike-games/\#:~:text=Beneath\%20Apple\%20Manor\%20,hidden\%20somewhere\%20beneath\%20the\%20manor}{\phantom{\rule{66.45000\ExReportUnit}{10.90399\ExReportUnit}}}};
\node[anchor=north west,inner sep=0,outer sep=0] at (94.31900,457.56400){\href{https://insight.ieeeusa.org/articles/going-rogue-a-brief-history-of-the-computerized-dungeon-crawl/\#:~:text=Both\%20Rogue\%20and\%20Beneath\%20Apple,enhanced\%20the\%20game\%E2\%80\%99s\%20replay\%20value}{\phantom{\rule{76.52200\ExReportUnit}{10.90402\ExReportUnit}}}};
\node[anchor=north west,inner sep=0,outer sep=0] at (101.96500,481.47501){\href{https://arstechnica.com/gaming/2020/03/ascii-art-permadeath-the-history-of-roguelike-games/\#:~:text=Hack\%20\%281980,Craddock}{\phantom{\rule{66.45000\ExReportUnit}{10.90298\ExReportUnit}}}};
\node[anchor=north west,inner sep=0,outer sep=0] at (175.37601,481.47501){\href{https://insight.ieeeusa.org/articles/going-rogue-a-brief-history-of-the-computerized-dungeon-crawl/\#:~:text=Both\%20Rogue\%20and\%20Beneath\%20Apple,enhanced\%20the\%20game\%E2\%80\%99s\%20replay\%20value}{\phantom{\rule{76.52199\ExReportUnit}{10.90298\ExReportUnit}}}};
\node[anchor=north west,inner sep=0,outer sep=0] at (311.00000,383.66468){\hyperlink{ex:rub:R11}{\phantom{\rule{24.00000\ExReportUnit}{12.00000\ExReportUnit}}}};
\end{tikzpicture}\par\endgroup\clearpage
\begingroup\setlength{\ExReportUnit}{1bp}
\centering
\begin{tikzpicture}[x=\ExReportUnit,y=-\ExReportUnit]
\node[anchor=north west,inner sep=0,outer sep=0] at (0,0){\includegraphics[page=3,width=396bp]{figures/appendix_roguelike_report.pdf}};
\node[inner sep=0,outer sep=0] at (12.0000,61.0511){\hypertarget{ex:ev:3}{}};
\node[inner sep=0,outer sep=0] at (12.0000,97.0936){\hypertarget{ex:ev:4}{}};
\node[inner sep=0,outer sep=0] at (12.0000,132.9585){\hypertarget{ex:ev:5}{}};
\node[inner sep=0,outer sep=0] at (12.0000,323.8183){\hypertarget{ex:ev:6}{}};
\node[inner sep=0,outer sep=0] at (12.0000,495.8033){\hypertarget{ex:ev:7}{}};
\node[anchor=north west,inner sep=0,outer sep=0] at (102.85200,101.89801){\href{https://arstechnica.com/gaming/2020/03/ascii-art-permadeath-the-history-of-roguelike-games/\#:~:text=traps\%20or\%20grabbing\%20an\%20item,else\%E2\%80\%94you\%20used\%20up\%20a\%20turn}{\phantom{\rule{66.45000\ExReportUnit}{10.90399\ExReportUnit}}}};
\node[anchor=north west,inner sep=0,outer sep=0] at (50.79700,137.76300){\href{https://arstechnica.com/gaming/2020/03/ascii-art-permadeath-the-history-of-roguelike-games/\#:~:text=death\%2C\%20which\%20came\%20easily\%20even,be\%20erased\%20immediately\%20on\%20reload}{\phantom{\rule{66.45000\ExReportUnit}{10.90399\ExReportUnit}}}};
\node[anchor=north west,inner sep=0,outer sep=0] at (214.66600,293.18100){\href{https://insight.ieeeusa.org/articles/going-rogue-a-brief-history-of-the-computerized-dungeon-crawl/\#:~:text=In\%20an\%20attempt\%20to\%20define,a\%20graphical\%20or\%20ASCII\%20tileset}{\phantom{\rule{76.52199\ExReportUnit}{10.90399\ExReportUnit}}}};
\node[anchor=north west,inner sep=0,outer sep=0] at (14.32100,424.26401){\href{https://arstechnica.com/gaming/2020/03/ascii-art-permadeath-the-history-of-roguelike-games/\#:~:text=In\%201980\%2C\%20UC\%20Santa\%20Cruz,galaxy\%20before\%20time\%20runs\%20out}{\phantom{\rule{66.45000\ExReportUnit}{10.90399\ExReportUnit}}}};
\node[anchor=north west,inner sep=0,outer sep=0] at (148.40601,448.17401){\href{https://arstechnica.com/gaming/2020/03/ascii-art-permadeath-the-history-of-roguelike-games/\#:~:text=In\%201980\%2C\%20UC\%20Santa\%20Cruz,galaxy\%20before\%20time\%20runs\%20out}{\phantom{\rule{66.45000\ExReportUnit}{10.90399\ExReportUnit}}}};
\node[anchor=north west,inner sep=0,outer sep=0] at (283.93500,512.56299){\href{https://insight.ieeeusa.org/articles/going-rogue-a-brief-history-of-the-computerized-dungeon-crawl/\#:~:text=emulate\%20PLATO\%E2\%80\%99s\%20graphical\%20capabilities\%2C\%20so,Adventure\%2C\%20Rogue\%20has\%20the\%20player}{\phantom{\rule{13.06100\ExReportUnit}{10.90399\ExReportUnit}}}};
\node[anchor=north west,inner sep=0,outer sep=0] at (11.00400,524.51801){\href{https://insight.ieeeusa.org/articles/going-rogue-a-brief-history-of-the-computerized-dungeon-crawl/\#:~:text=emulate\%20PLATO\%E2\%80\%99s\%20graphical\%20capabilities\%2C\%20so,Adventure\%2C\%20Rogue\%20has\%20the\%20player}{\phantom{\rule{68.77100\ExReportUnit}{10.90399\ExReportUnit}}}};
\node[anchor=north west,inner sep=0,outer sep=0] at (263.47198,548.42798){\href{https://arstechnica.com/gaming/2020/03/ascii-art-permadeath-the-history-of-roguelike-games/\#:~:text=With\%20help\%20from\%20a\%20Unix,the\%20risk\%20of\%20unintended\%20harm}{\phantom{\rule{33.52402\ExReportUnit}{10.90405\ExReportUnit}}}};
\node[anchor=north west,inner sep=0,outer sep=0] at (11.00400,560.38397){\href{https://arstechnica.com/gaming/2020/03/ascii-art-permadeath-the-history-of-roguelike-games/\#:~:text=With\%20help\%20from\%20a\%20Unix,the\%20risk\%20of\%20unintended\%20harm}{\phantom{\rule{38.23600\ExReportUnit}{10.90405\ExReportUnit}}}};
\node[anchor=north west,inner sep=0,outer sep=0] at (14.32100,584.29401){\href{https://arstechnica.com/gaming/2020/03/ascii-art-permadeath-the-history-of-roguelike-games/\#:~:text=textmode\%20paint\%20tool\%2C\%20the\%20pair,the\%20risk\%20of\%20unintended\%20harm}{\phantom{\rule{66.45000\ExReportUnit}{10.90399\ExReportUnit}}}};
\node[anchor=north west,inner sep=0,outer sep=0] at (311.00000,72.01937){\hyperlink{ex:rub:R2}{\phantom{\rule{20.00000\ExReportUnit}{12.00000\ExReportUnit}}}};
\node[anchor=north west,inner sep=0,outer sep=0] at (311.00000,110.21937){\hyperlink{ex:rub:R3}{\phantom{\rule{20.00000\ExReportUnit}{12.00000\ExReportUnit}}}};
\node[anchor=north west,inner sep=0,outer sep=0] at (311.00000,156.61937){\hyperlink{ex:rub:R4}{\phantom{\rule{20.00000\ExReportUnit}{12.00000\ExReportUnit}}}};
\node[anchor=north west,inner sep=0,outer sep=0] at (311.00000,334.60915){\hyperlink{ex:rub:R1}{\phantom{\rule{20.00000\ExReportUnit}{12.00000\ExReportUnit}}}};
\node[anchor=north west,inner sep=0,outer sep=0] at (311.00000,506.63959){\hyperlink{ex:rub:R9}{\phantom{\rule{20.00000\ExReportUnit}{12.00000\ExReportUnit}}}};
\end{tikzpicture}\par\endgroup\clearpage
\begingroup\setlength{\ExReportUnit}{1bp}
\centering
\begin{tikzpicture}[x=\ExReportUnit,y=-\ExReportUnit]
\node[anchor=north west,inner sep=0,outer sep=0] at (0,0){\includegraphics[page=4,width=396bp]{figures/appendix_roguelike_report.pdf}};
\node[inner sep=0,outer sep=0] at (12.0000,174.8015){\hypertarget{ex:ev:8}{}};
\node[inner sep=0,outer sep=0] at (12.0000,210.6675){\hypertarget{ex:ev:9}{}};
\node[inner sep=0,outer sep=0] at (12.0000,234.4010){\hypertarget{ex:ev:10}{}};
\node[inner sep=0,outer sep=0] at (12.0000,270.4434){\hypertarget{ex:ev:11}{}};
\node[inner sep=0,outer sep=0] at (12.0000,306.3083){\hypertarget{ex:ev:12}{}};
\node[inner sep=0,outer sep=0] at (12.0000,354.1293){\hypertarget{ex:ev:13}{}};
\node[inner sep=0,outer sep=0] at (12.0000,389.9953){\hypertarget{ex:ev:14}{}};
\node[inner sep=0,outer sep=0] at (12.0000,512.5353){\hypertarget{ex:ev:15}{}};
\node[anchor=north west,inner sep=0,outer sep=0] at (224.73801,18.21198){\href{https://arstechnica.com/gaming/2020/03/ascii-art-permadeath-the-history-of-roguelike-games/\#:~:text=textmode\%20paint\%20tool\%2C\%20the\%20pair,the\%20risk\%20of\%20unintended\%20harm}{\phantom{\rule{66.44998\ExReportUnit}{10.90405\ExReportUnit}}}};
\node[anchor=north west,inner sep=0,outer sep=0] at (50.67400,66.03198){\href{https://arstechnica.com/gaming/2020/03/ascii-art-permadeath-the-history-of-roguelike-games/\#:~:text=They\%20distributed\%20executable\%20versions\%20of,else\%E2\%80\%94you\%20used\%20up\%20a\%20turn}{\phantom{\rule{66.45000\ExReportUnit}{10.90399\ExReportUnit}}}};
\node[anchor=north west,inner sep=0,outer sep=0] at (82.00800,89.94299){\href{https://arstechnica.com/gaming/2020/03/ascii-art-permadeath-the-history-of-roguelike-games/\#:~:text=who\%20could\%20move\%20no\%20faster,else\%E2\%80\%94you\%20used\%20up\%20a\%20turn}{\phantom{\rule{66.44999\ExReportUnit}{10.90399\ExReportUnit}}}};
\node[anchor=north west,inner sep=0,outer sep=0] at (34.47800,125.80801){\href{https://arstechnica.com/gaming/2020/03/ascii-art-permadeath-the-history-of-roguelike-games/\#:~:text=University\%20of\%20California\%20campuses,else\%E2\%80\%94you\%20used\%20up\%20a\%20turn}{\phantom{\rule{66.45000\ExReportUnit}{10.90399\ExReportUnit}}}};
\node[anchor=north west,inner sep=0,outer sep=0] at (210.16600,227.42700){\href{https://arstechnica.com/gaming/2020/03/ascii-art-permadeath-the-history-of-roguelike-games/\#:~:text=When\%20Toy\%20transferred\%20to\%20UC,companies\%20all\%20around\%20the\%20world}{\phantom{\rule{66.45000\ExReportUnit}{10.90399\ExReportUnit}}}};
\node[anchor=north west,inner sep=0,outer sep=0] at (283.93500,227.42700){\href{https://insight.ieeeusa.org/articles/going-rogue-a-brief-history-of-the-computerized-dungeon-crawl/\#:~:text=applications\%20was\%20gaming\%2C\%20and\%20Rogue,and\%20contributed\%20to\%20its\%20development}{\phantom{\rule{13.06100\ExReportUnit}{10.90399\ExReportUnit}}}};
\node[anchor=north west,inner sep=0,outer sep=0] at (11.00400,239.38199){\href{https://insight.ieeeusa.org/articles/going-rogue-a-brief-history-of-the-computerized-dungeon-crawl/\#:~:text=applications\%20was\%20gaming\%2C\%20and\%20Rogue,and\%20contributed\%20to\%20its\%20development}{\phantom{\rule{68.77100\ExReportUnit}{10.90402\ExReportUnit}}}};
\node[anchor=north west,inner sep=0,outer sep=0] at (138.16200,263.29300){\href{https://arstechnica.com/gaming/2020/03/ascii-art-permadeath-the-history-of-roguelike-games/\#:~:text=When\%20Toy\%20transferred\%20to\%20UC,companies\%20all\%20around\%20the\%20world}{\phantom{\rule{66.44899\ExReportUnit}{10.90399\ExReportUnit}}}};
\node[anchor=north west,inner sep=0,outer sep=0] at (214.66600,263.29300){\href{https://insight.ieeeusa.org/articles/going-rogue-a-brief-history-of-the-computerized-dungeon-crawl/\#:~:text=Rogue\%20was\%20extremely\%20popular\%20on,beyond\%20its\%20commercial\%20PC\%20ports}{\phantom{\rule{76.52199\ExReportUnit}{10.90399\ExReportUnit}}}};
\node[anchor=north west,inner sep=0,outer sep=0] at (37.86700,358.93399){\href{https://insight.ieeeusa.org/articles/going-rogue-a-brief-history-of-the-computerized-dungeon-crawl/\#:~:text=Rogue\%20was\%20extremely\%20popular\%20on,beyond\%20its\%20commercial\%20PC\%20ports}{\phantom{\rule{76.52300\ExReportUnit}{10.90402\ExReportUnit}}}};
\node[anchor=north west,inner sep=0,outer sep=0] at (198.55200,418.71002){\href{https://arstechnica.com/gaming/2020/03/ascii-art-permadeath-the-history-of-roguelike-games/\#:~:text=Its\%20only\%20problem\%20was\%2C\%20in,game\%20never\%20really\%20caught\%20on}{\phantom{\rule{66.45001\ExReportUnit}{10.90399\ExReportUnit}}}};
\node[anchor=north west,inner sep=0,outer sep=0] at (263.47198,553.20599){\href{https://arstechnica.com/gaming/2020/03/ascii-art-permadeath-the-history-of-roguelike-games/\#:~:text=Beneath\%20Apple\%20Manor\%20,hidden\%20somewhere\%20beneath\%20the\%20manor}{\phantom{\rule{33.52402\ExReportUnit}{10.90399\ExReportUnit}}}};
\node[anchor=north west,inner sep=0,outer sep=0] at (11.00400,565.16101){\href{https://arstechnica.com/gaming/2020/03/ascii-art-permadeath-the-history-of-roguelike-games/\#:~:text=Beneath\%20Apple\%20Manor\%20,hidden\%20somewhere\%20beneath\%20the\%20manor}{\phantom{\rule{38.23600\ExReportUnit}{10.90399\ExReportUnit}}}};
\node[anchor=north west,inner sep=0,outer sep=0] at (311.00000,221.50375){\hyperlink{ex:rub:R8}{\phantom{\rule{20.00000\ExReportUnit}{12.00000\ExReportUnit}}}};
\node[anchor=north west,inner sep=0,outer sep=0] at (311.00000,259.70375){\hyperlink{ex:rub:R9}{\phantom{\rule{20.00000\ExReportUnit}{12.00000\ExReportUnit}}}};
\node[anchor=north west,inner sep=0,outer sep=0] at (311.00000,325.00064){\hyperlink{ex:rub:R7}{\phantom{\rule{20.00000\ExReportUnit}{12.00000\ExReportUnit}}}};
\node[anchor=north west,inner sep=0,outer sep=0] at (311.00000,400.78615){\hyperlink{ex:rub:R1}{\phantom{\rule{20.00000\ExReportUnit}{12.00000\ExReportUnit}}}};
\node[anchor=north west,inner sep=0,outer sep=0] at (311.00000,531.22661){\hyperlink{ex:rub:R10}{\phantom{\rule{24.00000\ExReportUnit}{12.00000\ExReportUnit}}}};
\end{tikzpicture}\par\endgroup\clearpage
\begingroup\setlength{\ExReportUnit}{1bp}
\centering
\begin{tikzpicture}[x=\ExReportUnit,y=-\ExReportUnit]
\node[anchor=north west,inner sep=0,outer sep=0] at (0,0){\includegraphics[page=5,width=396bp]{figures/appendix_roguelike_report.pdf}};
\node[inner sep=0,outer sep=0] at (12.0000,91.1156){\hypertarget{ex:ev:16}{}};
\node[inner sep=0,outer sep=0] at (12.0000,91.1156){\hypertarget{ex:ev:17}{}};
\node[inner sep=0,outer sep=0] at (12.0000,115.0265){\hypertarget{ex:ev:18}{}};
\node[anchor=north west,inner sep=0,outer sep=0] at (35.03400,30.16699){\href{https://arstechnica.com/gaming/2020/03/ascii-art-permadeath-the-history-of-roguelike-games/\#:~:text=Its\%20only\%20problem\%20was\%2C\%20in,game\%20never\%20really\%20caught\%20on}{\phantom{\rule{66.45000\ExReportUnit}{10.90399\ExReportUnit}}}};
\node[anchor=north west,inner sep=0,outer sep=0] at (191.86800,66.03198){\href{https://arstechnica.com/gaming/2020/03/ascii-art-permadeath-the-history-of-roguelike-games/\#:~:text=Its\%20only\%20problem\%20was\%2C\%20in,game\%20never\%20really\%20caught\%20on}{\phantom{\rule{66.45000\ExReportUnit}{10.90399\ExReportUnit}}}};
\node[anchor=north west,inner sep=0,outer sep=0] at (73.10400,119.72101){\href{https://arstechnica.com/gaming/2020/03/ascii-art-permadeath-the-history-of-roguelike-games/\#:~:text=The\%20birth\%20of\%20the\%20roguelike}{\phantom{\rule{66.45000\ExReportUnit}{10.87900\ExReportUnit}}}};
\node[anchor=north west,inner sep=0,outer sep=0] at (224.73801,143.74100){\href{https://arstechnica.com/gaming/2020/03/ascii-art-permadeath-the-history-of-roguelike-games/\#:~:text=Hack\%20\%281980,Craddock}{\phantom{\rule{66.44998\ExReportUnit}{10.78500\ExReportUnit}}}};
\node[anchor=north west,inner sep=0,outer sep=0] at (263.47198,167.65100){\href{https://arstechnica.com/gaming/2020/03/ascii-art-permadeath-the-history-of-roguelike-games/\#:~:text=Hack\%20\%281980,Craddock}{\phantom{\rule{33.52402\ExReportUnit}{10.90399\ExReportUnit}}}};
\node[anchor=north west,inner sep=0,outer sep=0] at (11.00400,179.60699){\href{https://arstechnica.com/gaming/2020/03/ascii-art-permadeath-the-history-of-roguelike-games/\#:~:text=Hack\%20\%281980,Craddock}{\phantom{\rule{38.23600\ExReportUnit}{10.90399\ExReportUnit}}}};
\node[anchor=north west,inner sep=0,outer sep=0] at (37.97200,215.47198){\href{https://arstechnica.com/gaming/2020/03/ascii-art-permadeath-the-history-of-roguelike-games/\#:~:text=he\%20no\%20longer\%20had\%20access,Craddock}{\phantom{\rule{66.45000\ExReportUnit}{10.90402\ExReportUnit}}}};
\node[anchor=north west,inner sep=0,outer sep=0] at (263.47198,239.38199){\href{https://arstechnica.com/gaming/2020/03/ascii-art-permadeath-the-history-of-roguelike-games/\#:~:text=It\%20had\%20elaborate\%20maze\%2Fdungeon,provided\%20his\%20copyright\%20remained\%20intact}{\phantom{\rule{33.52402\ExReportUnit}{10.90402\ExReportUnit}}}};
\node[anchor=north west,inner sep=0,outer sep=0] at (11.00400,251.33801){\href{https://arstechnica.com/gaming/2020/03/ascii-art-permadeath-the-history-of-roguelike-games/\#:~:text=It\%20had\%20elaborate\%20maze\%2Fdungeon,provided\%20his\%20copyright\%20remained\%20intact}{\phantom{\rule{38.23600\ExReportUnit}{10.90399\ExReportUnit}}}};
\node[anchor=north west,inner sep=0,outer sep=0] at (22.11900,287.09299){\href{https://arstechnica.com/gaming/2020/03/ascii-art-permadeath-the-history-of-roguelike-games/\#:~:text=It\%20had\%20elaborate\%20maze\%2Fdungeon,provided\%20his\%20copyright\%20remained\%20intact}{\phantom{\rule{66.45000\ExReportUnit}{11.01401\ExReportUnit}}}};
\node[anchor=north west,inner sep=0,outer sep=0] at (284.14499,335.02399){\href{https://arstechnica.com/gaming/2020/03/ascii-art-permadeath-the-history-of-roguelike-games/\#:~:text=Two\%20years\%20later\%2C\%20in\%20December,CWI\%20or\%20fans\%20on\%20Usenet}{\phantom{\rule{12.85101\ExReportUnit}{10.53000\ExReportUnit}}}};
\node[anchor=north west,inner sep=0,outer sep=0] at (11.00400,346.97900){\href{https://arstechnica.com/gaming/2020/03/ascii-art-permadeath-the-history-of-roguelike-games/\#:~:text=Two\%20years\%20later\%2C\%20in\%20December,CWI\%20or\%20fans\%20on\%20Usenet}{\phantom{\rule{58.70900\ExReportUnit}{10.90399\ExReportUnit}}}};
\node[anchor=north west,inner sep=0,outer sep=0] at (76.61100,346.97900){\href{https://arstechnica.com/gaming/2020/03/ascii-art-permadeath-the-history-of-roguelike-games/\#:~:text=to\%20previously\%20visited\%20dungeon\%20levels\%2C,lost\%20interest\%20and\%20stopped\%20development}{\phantom{\rule{66.45000\ExReportUnit}{10.90399\ExReportUnit}}}};
\node[anchor=north west,inner sep=0,outer sep=0] at (139.05099,436.64301){\href{https://arstechnica.com/gaming/2020/03/ascii-art-permadeath-the-history-of-roguelike-games/\#:~:text=Meanwhile\%2C\%20unbeknownst\%20to\%20Brouwer\%20and,deepest\%20recesses\%20of\%20the\%20dungeon}{\phantom{\rule{66.45001\ExReportUnit}{10.90399\ExReportUnit}}}};
\node[anchor=north west,inner sep=0,outer sep=0] at (263.47198,472.50800){\href{https://arstechnica.com/gaming/2020/03/ascii-art-permadeath-the-history-of-roguelike-games/\#:~:text=Meanwhile\%2C\%20unbeknownst\%20to\%20Brouwer\%20and,deepest\%20recesses\%20of\%20the\%20dungeon}{\phantom{\rule{33.52402\ExReportUnit}{10.90399\ExReportUnit}}}};
\node[anchor=north west,inner sep=0,outer sep=0] at (11.00400,484.46301){\href{https://arstechnica.com/gaming/2020/03/ascii-art-permadeath-the-history-of-roguelike-games/\#:~:text=Meanwhile\%2C\%20unbeknownst\%20to\%20Brouwer\%20and,deepest\%20recesses\%20of\%20the\%20dungeon}{\phantom{\rule{38.23600\ExReportUnit}{10.90399\ExReportUnit}}}};
\node[anchor=north west,inner sep=0,outer sep=0] at (103.64400,592.06000){\href{https://www.killscreen.com/brief-history-roguelike/\#:~:text=NetHack\%20took\%20the\%20code\%20and,fic}{\phantom{\rule{83.79500\ExReportUnit}{10.90399\ExReportUnit}}}};
\node[anchor=north west,inner sep=0,outer sep=0] at (311.00000,113.86235){\hyperlink{ex:rub:R5}{\phantom{\rule{20.00000\ExReportUnit}{12.00000\ExReportUnit}}}};
\end{tikzpicture}\par\endgroup\clearpage
\begingroup\setlength{\ExReportUnit}{1bp}
\centering
\begin{tikzpicture}[x=\ExReportUnit,y=-\ExReportUnit]
\node[anchor=north west,inner sep=0,outer sep=0] at (0,0){\includegraphics[page=6,width=396bp]{figures/appendix_roguelike_report.pdf}};
\node[anchor=north west,inner sep=0,outer sep=0] at (263.47198,340.57800){\href{https://arstechnica.com/gaming/2020/03/ascii-art-permadeath-the-history-of-roguelike-games/\#:~:text=Will\%20Crowther\%20and\%20Don\%20Woods\%E2\%80\%99,winning\%20players\%20to\%20start\%20again}{\phantom{\rule{33.52402\ExReportUnit}{10.90399\ExReportUnit}}}};
\node[anchor=north west,inner sep=0,outer sep=0] at (11.00400,352.53299){\href{https://arstechnica.com/gaming/2020/03/ascii-art-permadeath-the-history-of-roguelike-games/\#:~:text=Will\%20Crowther\%20and\%20Don\%20Woods\%E2\%80\%99,winning\%20players\%20to\%20start\%20again}{\phantom{\rule{38.23600\ExReportUnit}{10.90402\ExReportUnit}}}};
\node[anchor=north west,inner sep=0,outer sep=0] at (56.72900,352.53299){\href{https://insight.ieeeusa.org/articles/going-rogue-a-brief-history-of-the-computerized-dungeon-crawl/\#:~:text=applications\%20was\%20gaming\%2C\%20and\%20Rogue,and\%20contributed\%20to\%20its\%20development}{\phantom{\rule{76.52300\ExReportUnit}{10.90402\ExReportUnit}}}};
\end{tikzpicture}\par\endgroup\clearpage
\begingroup\setlength{\ExReportUnit}{1bp}
\centering
\begin{tikzpicture}[x=\ExReportUnit,y=-\ExReportUnit]
\node[anchor=north west,inner sep=0,outer sep=0] at (0,0){\includegraphics[page=7,width=396bp]{figures/appendix_roguelike_report.pdf}};
\node[anchor=north west,inner sep=0,outer sep=0] at (14.32100,54.07703){\href{https://insight.ieeeusa.org/articles/going-rogue-a-brief-history-of-the-computerized-dungeon-crawl/\#:~:text=In\%20an\%20attempt\%20to\%20define,a\%20graphical\%20or\%20ASCII\%20tileset}{\phantom{\rule{76.52200\ExReportUnit}{10.90399\ExReportUnit}}}};
\node[anchor=north west,inner sep=0,outer sep=0] at (52.83300,179.60699){\href{https://arstechnica.com/gaming/2020/03/ascii-art-permadeath-the-history-of-roguelike-games/\#:~:text=Although\%20they\%20had\%20no\%20direct,boss\%20fight\%20at\%20the\%20end}{\phantom{\rule{66.45000\ExReportUnit}{10.53000\ExReportUnit}}}};
\node[anchor=north west,inner sep=0,outer sep=0] at (126.62200,179.60699){\href{https://arstechnica.com/gaming/2020/03/ascii-art-permadeath-the-history-of-roguelike-games/\#:~:text=In\%201980\%2C\%20UC\%20Santa\%20Cruz,galaxy\%20before\%20time\%20runs\%20out}{\phantom{\rule{66.45000\ExReportUnit}{10.53000\ExReportUnit}}}};
\node[anchor=north west,inner sep=0,outer sep=0] at (200.41100,179.60699){\href{https://arstechnica.com/gaming/2020/03/ascii-art-permadeath-the-history-of-roguelike-games/\#:~:text=When\%20Toy\%20transferred\%20to\%20UC,companies\%20all\%20around\%20the\%20world}{\phantom{\rule{66.45000\ExReportUnit}{10.53000\ExReportUnit}}}};
\node[anchor=north west,inner sep=0,outer sep=0] at (46.55300,191.56201){\href{https://insight.ieeeusa.org/articles/going-rogue-a-brief-history-of-the-computerized-dungeon-crawl/\#:~:text=Beneath\%20Apple\%20Manor\%E2\%80\%99s\%20initial\%20graphics,by\%20Michael\%20Toy\%20and\%20Glenn}{\phantom{\rule{76.52200\ExReportUnit}{10.90399\ExReportUnit}}}};
\node[anchor=north west,inner sep=0,outer sep=0] at (130.60899,191.56201){\href{https://insight.ieeeusa.org/articles/going-rogue-a-brief-history-of-the-computerized-dungeon-crawl/\#:~:text=applications\%20was\%20gaming\%2C\%20and\%20Rogue,and\%20contributed\%20to\%20its\%20development}{\phantom{\rule{76.52301\ExReportUnit}{10.90399\ExReportUnit}}}};
\node[anchor=north west,inner sep=0,outer sep=0] at (214.66600,191.56201){\href{https://insight.ieeeusa.org/articles/going-rogue-a-brief-history-of-the-computerized-dungeon-crawl/\#:~:text=Rogue\%20was\%20extremely\%20popular\%20on,beyond\%20its\%20commercial\%20PC\%20ports}{\phantom{\rule{76.52199\ExReportUnit}{10.90399\ExReportUnit}}}};
\node[anchor=north west,inner sep=0,outer sep=0] at (58.83900,203.51700){\href{https://arstechnica.com/gaming/2020/03/ascii-art-permadeath-the-history-of-roguelike-games/\#:~:text=Hack\%20\%281980,Craddock}{\phantom{\rule{66.45000\ExReportUnit}{10.53000\ExReportUnit}}}};
\node[anchor=north west,inner sep=0,outer sep=0] at (132.82300,203.51700){\href{https://arstechnica.com/gaming/2020/03/ascii-art-permadeath-the-history-of-roguelike-games/\#:~:text=Will\%20Crowther\%20and\%20Don\%20Woods\%E2\%80\%99,winning\%20players\%20to\%20start\%20again}{\phantom{\rule{66.45000\ExReportUnit}{10.53000\ExReportUnit}}}};
\node[anchor=north west,inner sep=0,outer sep=0] at (227.22900,263.29300){\href{https://arstechnica.com/gaming/2020/03/ascii-art-permadeath-the-history-of-roguelike-games/\#:~:text=Its\%20only\%20problem\%20was\%2C\%20in,game\%20never\%20really\%20caught\%20on}{\phantom{\rule{66.44998\ExReportUnit}{10.90399\ExReportUnit}}}};
\node[anchor=north west,inner sep=0,outer sep=0] at (14.32100,275.24799){\href{https://insight.ieeeusa.org/articles/going-rogue-a-brief-history-of-the-computerized-dungeon-crawl/\#:~:text=Rogue\%20was\%20extremely\%20popular\%20on,beyond\%20its\%20commercial\%20PC\%20ports}{\phantom{\rule{76.52200\ExReportUnit}{10.90402\ExReportUnit}}}};
\node[anchor=north west,inner sep=0,outer sep=0] at (180.68600,299.15799){\href{https://insight.ieeeusa.org/articles/going-rogue-a-brief-history-of-the-computerized-dungeon-crawl/\#:~:text=In\%20an\%20attempt\%20to\%20define,a\%20graphical\%20or\%20ASCII\%20tileset}{\phantom{\rule{76.52200\ExReportUnit}{10.90402\ExReportUnit}}}};
\node[anchor=north west,inner sep=0,outer sep=0] at (373.04199,345.66199){\href{https://insight.ieeeusa.org/articles/going-rogue-a-brief-history-of-the-computerized-dungeon-crawl/\#:~:text=Possibly\%20more\%20than\%20any\%20other,Players\%20could\%20purchase}{\phantom{\rule{11.95401\ExReportUnit}{10.01300\ExReportUnit}}}};
\node[anchor=north west,inner sep=0,outer sep=0] at (25.79800,356.62100){\href{https://insight.ieeeusa.org/articles/going-rogue-a-brief-history-of-the-computerized-dungeon-crawl/\#:~:text=Possibly\%20more\%20than\%20any\%20other,Players\%20could\%20purchase}{\phantom{\rule{62.09400\ExReportUnit}{10.01300\ExReportUnit}}}};
\node[anchor=north west,inner sep=0,outer sep=0] at (289.38901,369.57202){\href{https://arstechnica.com/gaming/2020/03/ascii-art-permadeath-the-history-of-roguelike-games/\#:~:text=Although\%20they\%20had\%20no\%20direct,boss\%20fight\%20at\%20the\%20end}{\phantom{\rule{60.00400\ExReportUnit}{10.01300\ExReportUnit}}}};
\node[anchor=north west,inner sep=0,outer sep=0] at (289.38901,382.52399){\href{https://arstechnica.com/gaming/2020/03/ascii-art-permadeath-the-history-of-roguelike-games/\#:~:text=Will\%20Crowther\%20and\%20Don\%20Woods\%E2\%80\%99,winning\%20players\%20to\%20start\%20again}{\phantom{\rule{60.00400\ExReportUnit}{10.01202\ExReportUnit}}}};
\node[anchor=north west,inner sep=0,outer sep=0] at (373.04199,395.47501){\href{https://insight.ieeeusa.org/articles/going-rogue-a-brief-history-of-the-computerized-dungeon-crawl/\#:~:text=Beneath\%20Apple\%20Manor\%E2\%80\%99s\%20initial\%20graphics,by\%20Michael\%20Toy\%20and\%20Glenn}{\phantom{\rule{11.95401\ExReportUnit}{10.01300\ExReportUnit}}}};
\node[anchor=north west,inner sep=0,outer sep=0] at (25.79800,406.43402){\href{https://insight.ieeeusa.org/articles/going-rogue-a-brief-history-of-the-computerized-dungeon-crawl/\#:~:text=Beneath\%20Apple\%20Manor\%E2\%80\%99s\%20initial\%20graphics,by\%20Michael\%20Toy\%20and\%20Glenn}{\phantom{\rule{62.09400\ExReportUnit}{10.01300\ExReportUnit}}}};
\node[anchor=north west,inner sep=0,outer sep=0] at (373.04199,419.38501){\href{https://insight.ieeeusa.org/articles/going-rogue-a-brief-history-of-the-computerized-dungeon-crawl/\#:~:text=emulate\%20PLATO\%E2\%80\%99s\%20graphical\%20capabilities\%2C\%20so,Adventure\%2C\%20Rogue\%20has\%20the\%20player}{\phantom{\rule{11.95401\ExReportUnit}{10.01300\ExReportUnit}}}};
\node[anchor=north west,inner sep=0,outer sep=0] at (25.79800,430.34399){\href{https://insight.ieeeusa.org/articles/going-rogue-a-brief-history-of-the-computerized-dungeon-crawl/\#:~:text=emulate\%20PLATO\%E2\%80\%99s\%20graphical\%20capabilities\%2C\%20so,Adventure\%2C\%20Rogue\%20has\%20the\%20player}{\phantom{\rule{62.09400\ExReportUnit}{10.01300\ExReportUnit}}}};
\node[anchor=north west,inner sep=0,outer sep=0] at (373.04199,443.29602){\href{https://insight.ieeeusa.org/articles/going-rogue-a-brief-history-of-the-computerized-dungeon-crawl/\#:~:text=applications\%20was\%20gaming\%2C\%20and\%20Rogue,and\%20contributed\%20to\%20its\%20development}{\phantom{\rule{11.95401\ExReportUnit}{10.01300\ExReportUnit}}}};
\node[anchor=north west,inner sep=0,outer sep=0] at (25.79800,454.25500){\href{https://insight.ieeeusa.org/articles/going-rogue-a-brief-history-of-the-computerized-dungeon-crawl/\#:~:text=applications\%20was\%20gaming\%2C\%20and\%20Rogue,and\%20contributed\%20to\%20its\%20development}{\phantom{\rule{62.09400\ExReportUnit}{10.01199\ExReportUnit}}}};
\node[anchor=north west,inner sep=0,outer sep=0] at (289.38901,467.20599){\href{https://arstechnica.com/gaming/2020/03/ascii-art-permadeath-the-history-of-roguelike-games/\#:~:text=They\%20distributed\%20executable\%20versions\%20of,else\%E2\%80\%94you\%20used\%20up\%20a\%20turn}{\phantom{\rule{60.00400\ExReportUnit}{10.01300\ExReportUnit}}}};
\node[anchor=north west,inner sep=0,outer sep=0] at (373.04199,480.15698){\href{https://insight.ieeeusa.org/articles/going-rogue-a-brief-history-of-the-computerized-dungeon-crawl/\#:~:text=Directly\%20inspired\%20by\%20Dungeons\%20and,and\%20contributed\%20to\%20its\%20development}{\phantom{\rule{11.95401\ExReportUnit}{10.01300\ExReportUnit}}}};
\node[anchor=north west,inner sep=0,outer sep=0] at (25.79800,491.11600){\href{https://insight.ieeeusa.org/articles/going-rogue-a-brief-history-of-the-computerized-dungeon-crawl/\#:~:text=Directly\%20inspired\%20by\%20Dungeons\%20and,and\%20contributed\%20to\%20its\%20development}{\phantom{\rule{62.09400\ExReportUnit}{10.01300\ExReportUnit}}}};
\node[anchor=north west,inner sep=0,outer sep=0] at (235.61800,504.06799){\href{https://www.killscreen.com/brief-history-roguelike/\#:~:text=Thirty\%20years\%20the\%20senior\%20of,how\%20avid\%20the\%20fanbase\%20is}{\phantom{\rule{75.61501\ExReportUnit}{10.01300\ExReportUnit}}}};
\node[anchor=north west,inner sep=0,outer sep=0] at (289.38901,517.01898){\href{https://arstechnica.com/gaming/2020/03/ascii-art-permadeath-the-history-of-roguelike-games/\#:~:text=with\%20riches\%20but\%20also\%20with,winning\%20players\%20to\%20start\%20again}{\phantom{\rule{60.00400\ExReportUnit}{10.01300\ExReportUnit}}}};
\node[anchor=north west,inner sep=0,outer sep=0] at (289.38901,529.97101){\href{https://arstechnica.com/gaming/2020/03/ascii-art-permadeath-the-history-of-roguelike-games/\#:~:text=Beneath\%20Apple\%20Manor\%20,hidden\%20somewhere\%20beneath\%20the\%20manor}{\phantom{\rule{60.00400\ExReportUnit}{10.01202\ExReportUnit}}}};
\node[anchor=north west,inner sep=0,outer sep=0] at (373.04199,542.92200){\href{https://insight.ieeeusa.org/articles/going-rogue-a-brief-history-of-the-computerized-dungeon-crawl/\#:~:text=Both\%20Rogue\%20and\%20Beneath\%20Apple,enhanced\%20the\%20game\%E2\%80\%99s\%20replay\%20value}{\phantom{\rule{11.95401\ExReportUnit}{10.01300\ExReportUnit}}}};
\node[anchor=north west,inner sep=0,outer sep=0] at (25.79800,553.88098){\href{https://insight.ieeeusa.org/articles/going-rogue-a-brief-history-of-the-computerized-dungeon-crawl/\#:~:text=Both\%20Rogue\%20and\%20Beneath\%20Apple,enhanced\%20the\%20game\%E2\%80\%99s\%20replay\%20value}{\phantom{\rule{62.09400\ExReportUnit}{10.01300\ExReportUnit}}}};
\node[anchor=north west,inner sep=0,outer sep=0] at (289.38901,566.83203){\href{https://arstechnica.com/gaming/2020/03/ascii-art-permadeath-the-history-of-roguelike-games/\#:~:text=Hack\%20\%281980,Craddock}{\phantom{\rule{60.00400\ExReportUnit}{10.01294\ExReportUnit}}}};
\node[anchor=north west,inner sep=0,outer sep=0] at (289.38901,579.78400){\href{https://arstechnica.com/gaming/2020/03/ascii-art-permadeath-the-history-of-roguelike-games/\#:~:text=traps\%20or\%20grabbing\%20an\%20item,else\%E2\%80\%94you\%20used\%20up\%20a\%20turn}{\phantom{\rule{60.00400\ExReportUnit}{10.01300\ExReportUnit}}}};
\node[anchor=north west,inner sep=0,outer sep=0] at (289.38901,592.73499){\href{https://arstechnica.com/gaming/2020/03/ascii-art-permadeath-the-history-of-roguelike-games/\#:~:text=death\%2C\%20which\%20came\%20easily\%20even,be\%20erased\%20immediately\%20on\%20reload}{\phantom{\rule{60.00400\ExReportUnit}{10.01300\ExReportUnit}}}};
\end{tikzpicture}\par\endgroup\clearpage
\begingroup\setlength{\ExReportUnit}{1bp}
\centering
\begin{tikzpicture}[x=\ExReportUnit,y=-\ExReportUnit]
\node[anchor=north west,inner sep=0,outer sep=0] at (0,0){\includegraphics[page=8,width=396bp]{figures/appendix_roguelike_report.pdf}};
\node[anchor=north west,inner sep=0,outer sep=0] at (373.04199,18.88702){\href{https://insight.ieeeusa.org/articles/going-rogue-a-brief-history-of-the-computerized-dungeon-crawl/\#:~:text=In\%20an\%20attempt\%20to\%20define,a\%20graphical\%20or\%20ASCII\%20tileset}{\phantom{\rule{11.95401\ExReportUnit}{10.01300\ExReportUnit}}}};
\node[anchor=north west,inner sep=0,outer sep=0] at (25.79800,29.84601){\href{https://insight.ieeeusa.org/articles/going-rogue-a-brief-history-of-the-computerized-dungeon-crawl/\#:~:text=In\%20an\%20attempt\%20to\%20define,a\%20graphical\%20or\%20ASCII\%20tileset}{\phantom{\rule{62.09400\ExReportUnit}{10.01300\ExReportUnit}}}};
\node[anchor=north west,inner sep=0,outer sep=0] at (289.38901,42.79797){\href{https://arstechnica.com/gaming/2020/03/ascii-art-permadeath-the-history-of-roguelike-games/\#:~:text=In\%201980\%2C\%20UC\%20Santa\%20Cruz,galaxy\%20before\%20time\%20runs\%20out}{\phantom{\rule{60.00400\ExReportUnit}{10.01202\ExReportUnit}}}};
\node[anchor=north west,inner sep=0,outer sep=0] at (289.38901,55.74902){\href{https://arstechnica.com/gaming/2020/03/ascii-art-permadeath-the-history-of-roguelike-games/\#:~:text=With\%20help\%20from\%20a\%20Unix,the\%20risk\%20of\%20unintended\%20harm}{\phantom{\rule{60.00400\ExReportUnit}{10.01300\ExReportUnit}}}};
\node[anchor=north west,inner sep=0,outer sep=0] at (289.38901,68.70001){\href{https://arstechnica.com/gaming/2020/03/ascii-art-permadeath-the-history-of-roguelike-games/\#:~:text=textmode\%20paint\%20tool\%2C\%20the\%20pair,the\%20risk\%20of\%20unintended\%20harm}{\phantom{\rule{60.00400\ExReportUnit}{10.01300\ExReportUnit}}}};
\node[anchor=north west,inner sep=0,outer sep=0] at (289.38901,81.65198){\href{https://arstechnica.com/gaming/2020/03/ascii-art-permadeath-the-history-of-roguelike-games/\#:~:text=who\%20could\%20move\%20no\%20faster,else\%E2\%80\%94you\%20used\%20up\%20a\%20turn}{\phantom{\rule{60.00400\ExReportUnit}{10.01300\ExReportUnit}}}};
\node[anchor=north west,inner sep=0,outer sep=0] at (289.38901,94.60303){\href{https://arstechnica.com/gaming/2020/03/ascii-art-permadeath-the-history-of-roguelike-games/\#:~:text=University\%20of\%20California\%20campuses,else\%E2\%80\%94you\%20used\%20up\%20a\%20turn}{\phantom{\rule{60.00400\ExReportUnit}{10.01300\ExReportUnit}}}};
\node[anchor=north west,inner sep=0,outer sep=0] at (289.38901,107.55499){\href{https://arstechnica.com/gaming/2020/03/ascii-art-permadeath-the-history-of-roguelike-games/\#:~:text=When\%20Toy\%20transferred\%20to\%20UC,companies\%20all\%20around\%20the\%20world}{\phantom{\rule{60.00400\ExReportUnit}{10.01300\ExReportUnit}}}};
\node[anchor=north west,inner sep=0,outer sep=0] at (373.04199,120.50601){\href{https://insight.ieeeusa.org/articles/going-rogue-a-brief-history-of-the-computerized-dungeon-crawl/\#:~:text=Rogue\%20was\%20extremely\%20popular\%20on,beyond\%20its\%20commercial\%20PC\%20ports}{\phantom{\rule{11.95401\ExReportUnit}{10.01300\ExReportUnit}}}};
\node[anchor=north west,inner sep=0,outer sep=0] at (25.79800,131.46500){\href{https://insight.ieeeusa.org/articles/going-rogue-a-brief-history-of-the-computerized-dungeon-crawl/\#:~:text=Rogue\%20was\%20extremely\%20popular\%20on,beyond\%20its\%20commercial\%20PC\%20ports}{\phantom{\rule{62.09400\ExReportUnit}{10.01300\ExReportUnit}}}};
\node[anchor=north west,inner sep=0,outer sep=0] at (289.38901,144.41699){\href{https://arstechnica.com/gaming/2020/03/ascii-art-permadeath-the-history-of-roguelike-games/\#:~:text=Its\%20only\%20problem\%20was\%2C\%20in,game\%20never\%20really\%20caught\%20on}{\phantom{\rule{60.00400\ExReportUnit}{10.01199\ExReportUnit}}}};
\node[anchor=north west,inner sep=0,outer sep=0] at (289.38901,157.36801){\href{https://arstechnica.com/gaming/2020/03/ascii-art-permadeath-the-history-of-roguelike-games/\#:~:text=The\%20birth\%20of\%20the\%20roguelike}{\phantom{\rule{60.00400\ExReportUnit}{10.01300\ExReportUnit}}}};
\node[anchor=north west,inner sep=0,outer sep=0] at (289.38901,170.31900){\href{https://arstechnica.com/gaming/2020/03/ascii-art-permadeath-the-history-of-roguelike-games/\#:~:text=he\%20no\%20longer\%20had\%20access,Craddock}{\phantom{\rule{60.00400\ExReportUnit}{10.01300\ExReportUnit}}}};
\node[anchor=north west,inner sep=0,outer sep=0] at (289.38901,183.27100){\href{https://arstechnica.com/gaming/2020/03/ascii-art-permadeath-the-history-of-roguelike-games/\#:~:text=It\%20had\%20elaborate\%20maze\%2Fdungeon,provided\%20his\%20copyright\%20remained\%20intact}{\phantom{\rule{60.00400\ExReportUnit}{10.01300\ExReportUnit}}}};
\node[anchor=north west,inner sep=0,outer sep=0] at (289.38901,196.22198){\href{https://arstechnica.com/gaming/2020/03/ascii-art-permadeath-the-history-of-roguelike-games/\#:~:text=Two\%20years\%20later\%2C\%20in\%20December,CWI\%20or\%20fans\%20on\%20Usenet}{\phantom{\rule{60.00400\ExReportUnit}{10.01300\ExReportUnit}}}};
\node[anchor=north west,inner sep=0,outer sep=0] at (289.38901,209.17401){\href{https://arstechnica.com/gaming/2020/03/ascii-art-permadeath-the-history-of-roguelike-games/\#:~:text=to\%20previously\%20visited\%20dungeon\%20levels\%2C,lost\%20interest\%20and\%20stopped\%20development}{\phantom{\rule{60.00400\ExReportUnit}{10.01199\ExReportUnit}}}};
\node[anchor=north west,inner sep=0,outer sep=0] at (289.38901,222.12500){\href{https://arstechnica.com/gaming/2020/03/ascii-art-permadeath-the-history-of-roguelike-games/\#:~:text=Meanwhile\%2C\%20unbeknownst\%20to\%20Brouwer\%20and,deepest\%20recesses\%20of\%20the\%20dungeon}{\phantom{\rule{60.00400\ExReportUnit}{10.01300\ExReportUnit}}}};
\node[anchor=north west,inner sep=0,outer sep=0] at (235.61800,235.07700){\href{https://www.killscreen.com/brief-history-roguelike/\#:~:text=NetHack\%20took\%20the\%20code\%20and,fic}{\phantom{\rule{75.61501\ExReportUnit}{10.01199\ExReportUnit}}}};
\end{tikzpicture}\par\endgroup\smallskip

\begingroup
\noindent \textbf{Depth and continuity along knowledge chains.} This example illustrates depth through sustained coverage along related knowledge chains and continuity through coverage of their intermediate relations. Along R1\ensuremath{\rightarrow}R2\ensuremath{\rightarrow}R3 and R1\ensuremath{\rightarrow}R2\ensuremath{\rightarrow}R4, the report identifies Rogue as the genre's forerunner and namesake (R1), explains its permanent-death mechanic (R2), and develops this account through practical enforcement and cultural connections. Specifically, R3 describes copying save files to circumvent permanent death and the authors' use of save-file deletion to enforce it, while R4 connects permanent character death to tabletop role-playing traditions. All intermediate relations on these paths are covered, and R2--R4 have distinct supporting sentences within the same paragraph. The example thus makes both deeper coverage and continuity along the reference chains directly inspectable.\par\medskip

\noindent \textbf{Logical coherence across sections.} Consecutive parent--child rubrics in the reference tree can be satisfied by evidence from different report sections while still corresponding to logically connected knowledge. For R1\ensuremath{\rightarrow}R5, R1 is satisfied in the section that describes how Rogue was developed and widely distributed, explaining how it became the genre's defining model. R5 is satisfied in the subsequent section that discusses early roguelike variants and their developers, describing how Hack's creator attempted to recreate Rogue from memory. The discussion progresses from establishing Rogue's foundational role to showing how its design influenced a specific successor, preserving the continuous reasoning across the section boundary and thus demonstrating meaningful depth. Similarly, R10 and R11 connect Beneath Apple Manor's position before Rogue with its early use of procedural generation, bringing together complementary historical and technical information from separate sections.\par\medskip

\noindent \textbf{Informative omissions in otherwise covered branches.} R6 and R12 identify specific gaps despite coverage of their parent relations. R6 asks about Hack's cross-platform ports and their use of graphical tiles, directly addressing the query's interest in display constraints and platform changes. The report's discussion of graphical tiles in Beneath Apple Manor does not satisfy this Hack-specific requirement. R12 asks about predefined seeds for generating large worlds; describing randomized dungeons alone does not explain this mechanism. R12 offers an optional technical extension to the discussion of memory-constrained design. Although unmet in this report, these rubrics remain valuable for the task, helping distinguish reports by the specificity and completeness of their knowledge coverage.\par\medskip
\endgroup
\FloatBarrier
\clearpage

\section{Local Research Framework}
\label{app:local-research-framework}

\begin{figure}[H]
\centering
\includegraphics[width=\textwidth]{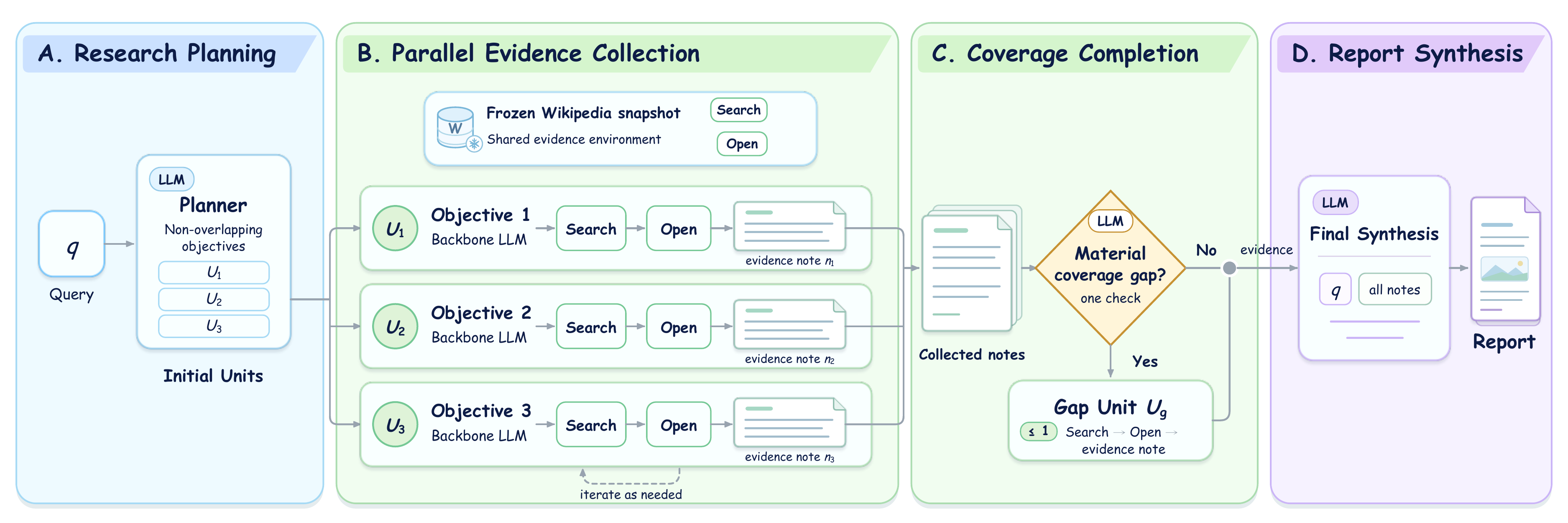}
\caption{\textbf{Local research framework workflow.} The backbone LLM decomposes the research query into three Research Units, which execute concurrently against the shared frozen Wikipedia snapshot through \textsc{Search} and \textsc{Open}. Their evidence notes undergo one coverage-gap check; when a material gap is found, at most one Gap Unit collects supplemental evidence before the original query and all collected notes are synthesized into the final report.}
\label{fig:local-research-framework-overview}
\end{figure}

The local research framework provides a common retrieval and orchestration setting for evaluating backbone LLMs. It standardizes the evidence corpus, tool interface, research stages, and interaction budget while leaving planning decisions, tool-use choices, and report generation to each backbone. Figure~\ref{fig:local-research-framework-overview} summarizes the resulting workflow.

\subsection{Retrieval Environment and Tool Interface}
\label{app:local-research-tools}

All evaluations under the local research framework use the fixed English Wikipedia snapshot \texttt{enwiki-\allowbreak20260701-\allowbreak v1}. The report generator receives the benchmark query. Provider-native web search is disabled, so every backbone searches the same time-invariant evidence space through the framework's local tools.

The framework exposes two operations over this snapshot. \textsc{Search} and \textsc{Open} serve distinct roles: \textsc{Search} identifies potentially relevant Wikipedia pages, whereas \textsc{Open} reads their contents. Given a source identifier and a cursor position, \textsc{Open} returns a bounded text span. If additional text remains, it also returns a \texttt{next\_cursor} that can be used to retrieve the subsequent span. All backbone LLMs use the same tool schemas, although they independently decide what to search for and which sources to read.

\Needspace{17\baselineskip}
\subsection{Research Workflow and Formal Configuration}
\label{app:local-research-workflow}

\newsavebox{\localconfigurationtablebox}
\sbox{\localconfigurationtablebox}{%
\footnotesize
\setlength{\tabcolsep}{2pt}%
\begin{tabular}{@{}lc@{}}
\toprule
Parameter & Value \\
\midrule
Wikipedia snapshot & \texttt{enwiki-20260701-v1} \\
Retrieval tools & \textsc{Search} and \textsc{Open} \\
Provider-native web search & Disabled \\
Initial Research Units & 3 \\
Concurrent Research Units & 3 \\
Gap Units & At most 1 \\
Total Research Units & At most 4 \\
Model rounds per Unit & 7 \\
Tool calls per Unit & At most 15 \\
\textsc{Search} calls per Unit & At most 3 \\
\textsc{Open} calls per Unit & At most 12 \\
\textsc{Open} calls per model turn & At most 4 \\
Retrieved content per \textsc{Open} call & At most 6,000 characters \\
\bottomrule
\end{tabular}%
}
\begingroup
\setlength{\columnsep}{11pt}
\begin{wraptable}[15]{r}{\wd\localconfigurationtablebox}
\centering
\caption{\textbf{Key local research framework configuration.} The same settings are used for all seven backbone LLMs evaluated under the framework.}
\label{tab:local-research-configuration}
\usebox{\localconfigurationtablebox}
\end{wraptable}

Table~\ref{tab:local-research-configuration} summarizes the configuration of our local research framework used for the seven backbone LLMs in the main experiments.

Given a benchmark query, a planner first decomposes the task into exactly three non-overlapping Research Units, each defined by a short title and a bounded objective. The three initial Units execute concurrently. Within each Unit, the backbone uses \textsc{Search} and \textsc{Open} to investigate its assigned objective under fixed round and tool budgets, then returns concise evidence notes rather than a complete report. This decomposition preserves a common research structure while allowing each backbone to decide which queries to issue and which pages to read.

After the initial Units finish, the framework performs one coverage-gap check over the query and collected evidence notes. If the check identifies a material missing direction that can be investigated within the frozen Wikipedia snapshot, it creates one supplemental Gap Unit. The workflow therefore executes three Units by default and at most four Units in total, providing a bounded second stage of evidence collection without changing the protocol across backbone LLMs.

Finally, the synthesis stage receives the original query and the evidence notes produced by the Research Units and combines them into a coherent Markdown report. The resulting report is subsequently scored with the same rubric set and judge protocol used for all evaluated agents.

\finishwraptable
\endgroup
\Needspace{6\baselineskip}
\section{Evaluation Timeframes for Evaluated Deep Research Agents}
\label{app:evaluation-timeframes}

The outputs evaluated in this paper were collected during the timeframes listed in Table~\ref{tab:evaluation-timeframes}. These dates indicate when the reports used in our experiments were generated.

\begin{appendixtable}[H]{\textwidth}
\centering
\footnotesize
\setlength{\tabcolsep}{2pt}
\caption{Data collection timeframes for the evaluated deep research agents. All dates refer to 2026.}
\label{tab:evaluation-timeframes}
\begin{tabular}{@{}lcc@{}}
\toprule
Agent Category & Evaluated Model & \shortstack{Data Collection\\Date Range} \\
\midrule
\multirow{3}{*}{Deep Research Systems}
  & o3 Deep Research & July 21--22 \\
  & Perplexity Sonar Deep Research & July 21--24 \\
  & Qwen Deep Research & July 21--24 \\
\addlinespace[3pt]
\multirow{6}{*}{Native Search LLMs}
  & Claude Opus 5 & August 12--14 \\
  & Grok 4.5 & August 12--14 \\
  & GPT-5.6 Sol & August 12--14 \\
  & Gemini 3.1 Pro Preview & August 12--14 \\
  & Sonar Reasoning Pro & August 12--14 \\
  & Qwen3.8 Max (Native Search) & August 12--14 \\
\addlinespace[3pt]
\multirow{7}{*}{local research framework}
  & Qwen3.8 Max & August 12--14 \\
  & GLM 5.2 & August 12--14 \\
  & MiniMax M3 & August 12--14 \\
  & DeepSeek V4 Flash & August 12--14 \\
  & DeepSeek V4 Pro & August 12--14 \\
  & Qwen3.7 Plus & August 12--14 \\
  & Gemma 4 31B & August 12--14 \\
\bottomrule
\end{tabular}
\end{appendixtable}

\Needspace{20\baselineskip}
\section{Cost Accounting}
\label{app:model-call-costs}

We report recorded model-call costs in USD for benchmark construction, report generation, and judge evaluation.

\Needspace{14\baselineskip}
\subsection{Benchmark Construction}

\newsavebox{\constructioncosttablebox}
\sbox{\constructioncosttablebox}{%
\footnotesize
\setlength{\tabcolsep}{4pt}%
\begin{tabular}{@{}lcc@{}}
\toprule
Stage & Mean/root & Total (48 roots) \\
\midrule
Graph construction & 26.9463 & 1,293.42 \\
Subgraph sampling & 0.1124 & 5.39 \\
Query generation & 0.3505 & 16.82 \\
Rubric generation & 6.1453 & 294.97 \\
\midrule
\textbf{Total} & \textbf{33.5544} & \textbf{1,610.61} \\
\bottomrule
\end{tabular}%
}
\begin{wraptable}{r}{\wd\constructioncosttablebox}
\centering
\caption{Benchmark construction cost (USD).}
\label{tab:benchmark-construction-costs}
\usebox{\constructioncosttablebox}
\end{wraptable}

We initially constructed local graphs for 240 candidate root entities and sampled three cases from each, yielding 720 candidate cases. Based on the benchmark-size analysis in Appendix~\ref{app:sample-size-stability}, we retained 48 roots and their 144 cases for the formal benchmark. All four construction stages use Claude Opus 4.8 for their model calls. Table~\ref{tab:benchmark-construction-costs} reports the construction costs for this retained set.

\finishwraptable
\subsection{Report Generation}

For each of the three Deep Research Systems, we generated reports for all 720 candidate cases. Table~\ref{tab:report-generation-costs} reports the mean cost per accepted report and normalizes every agent to the final 144-case evaluation scope.

\begin{appendixtable}[H]{\textwidth}
\centering
\footnotesize
\setlength{\tabcolsep}{3pt}
\caption{Model-call cost of report generation for 16 agents (USD).}
\label{tab:report-generation-costs}
\begin{tabular}{@{}llcc@{}}
\toprule
Track & Model & Mean/case & Total (144 cases) \\
\midrule
\multirow{3}{*}{Deep Research Systems}
  & o3 Deep Research & 2.1302 & 306.75 \\
  & Perplexity Sonar Deep Research & 0.5571 & 80.22 \\
  & Qwen Deep Research & 0.2069 & 29.80 \\
\addlinespace[3pt]
\multirow{6}{*}{Native Search LLMs}
  & Claude Opus 5 & 1.8110 & 260.79 \\
  & GPT-5.6 Sol & 1.2967 & 186.72 \\
  & Gemini 3.1 Pro Preview & 0.1703 & 24.52 \\
  & Grok 4.5 & 0.2274 & 32.75 \\
  & Qwen3.8 Max & 0.0873 & 12.58 \\
  & Sonar Reasoning Pro & 0.0510 & 7.34 \\
\addlinespace[3pt]
\multirow{7}{*}{local research framework}
  & DeepSeek V4 Flash & 0.0125 & 1.79 \\
  & DeepSeek V4 Pro & 0.1336 & 19.24 \\
  & Gemma 4 31B & 0.0699 & 10.06 \\
  & GLM 5.2 & 0.0795 & 11.45 \\
  & MiniMax M3 & 0.0620 & 8.92 \\
  & Qwen3.7 Plus & 0.0925 & 13.32 \\
  & Qwen3.8 Max & 0.4377 & 63.03 \\
\midrule
\multicolumn{2}{@{}l}{\textbf{All 16 deep research agents}} & \textbf{7.4255} & \textbf{1,069.27} \\
\bottomrule
\end{tabular}
\end{appendixtable}

\FloatBarrier
\Needspace{7\baselineskip}
\subsection{Judge Evaluation}

The formal evaluation contains $144\times16=2{,}304$ report--case evaluations. We use Gemini 2.5 Pro as the judge model for all evaluations. The recorded model-call cost is USD~0.2731 per report on average and USD~629.11 in total.

\FloatBarrier
\Needspace{8\baselineskip}
\section{Human Annotation Protocol}
\label{app:human-annotation}

\subsection{Rubric-level annotation}
\label{app:rubric-level-annotation}
\paragraph{Annotators and annotation procedure.}
The annotations support the rubric-level human--judge comparison in Section~\ref{sec:human-judge-consistency}. Ten PhD-level annotators participated in the human evaluation. The annotation sample comprised 3,695 report--rubric pairs from 12 reports, with each report representing a different benchmark domain. Under the annotation protocol, each pair receives two independent initial labels. Annotators do not have access to one another's labels or the LLM judges' verdicts. A third annotator adjudicates disagreements using the relevant report passages and the same annotation criteria, producing the final human reference.

\begin{figure}[htbp]
\centering
\includegraphics[width=\textwidth]{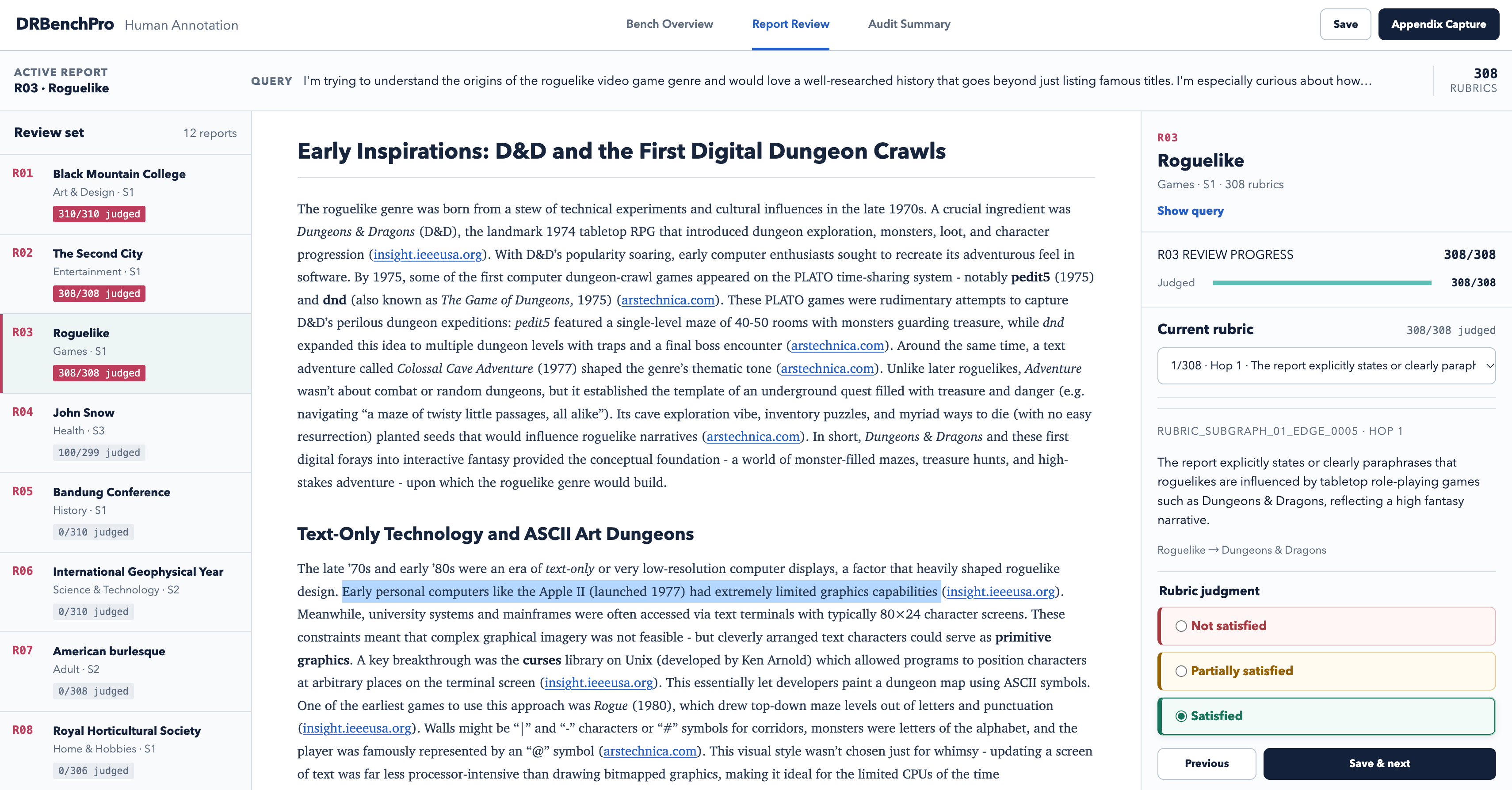}
\caption{\textbf{Human annotation interface.} Annotators assign \emph{Satisfied}, \emph{Partially Satisfied}, or \emph{Not Satisfied} to each report--rubric pair.}
\label{fig:human-annotation-interface}
\end{figure}

\paragraph{Annotation criteria.}
Each report--rubric pair received one of three labels: \emph{Satisfied}, \emph{Partially Satisfied}, or \emph{Not Satisfied}. \emph{Satisfied} indicates that the report explicitly states or clearly paraphrases the rubric's required knowledge, including its essential relation and qualifications. \emph{Partially Satisfied} indicates that the report covers a substantive part of the requirement but omits an essential detail or qualification. \emph{Not Satisfied} indicates that the required knowledge is absent; merely mentioning a related entity or discussing the general topic does not constitute coverage. Judgments concern content present in the report, without filling gaps from external knowledge. Figure~\ref{fig:human-annotation-interface} illustrates the annotation interface during the report review process.

\paragraph{Agreement and judge comparison.}
Initial inter-annotator agreement was Cohen's \(\kappa = 0.841\), computed before adjudication over all 3,695 report--rubric pairs using the original three-category labels. The resulting human-reviewed labels serve as the reference for comparison with three LLM judges. We compare the judges' original three-category verdicts with this reference. For the binary comparison, \emph{Satisfied} is mapped to 1 (covered), while \emph{Partially Satisfied} and \emph{Not Satisfied} are mapped to 0 (uncovered), for both human and judge labels. Macro-F1 is computed separately for each report and then averaged equally across the 12 reports.

\FloatBarrier
\Needspace{9\baselineskip}
\subsection{Report-level expert assessment}
\label{app:report-level-annotation}
The report-level assessment in Section~\ref{sec:human-judge-consistency} comprises 72 reports from six deep research agents across 12 cases from distinct domains. Each case includes six reports answering the same research query. Ten PhD-level annotators participate in this assessment. For each case, the same three annotators from this pool independently rate all six reports on breadth, depth, and overall quality using a 1--10 integer scale, and their mean ratings serve as the human reference. Annotators assess reports without access to agent identities, automatically computed scores, or one another's ratings. Differences among annotators are reflected in the mean rating.

\Needspace{10\baselineskip}
\section{Prompt Templates}
\label{app:prompt-templates}

We release the prompts used for benchmark construction and for evaluation to enable reproducibility and independent analysis.

\subsection{Benchmark Construction Prompts}
\label{app:benchmark-construction-prompts}

\subsubsection{Root Entity Neighbor Selection}
\label{app:prompt-root-full-page-selection}

This prompt selects the children of the root from candidate entities extracted across its complete Wikipedia page.

\begin{AIbox}{System Prompt}
You are selecting a set of directly related first-hop entities from the candidate links found in the full page of a core Wikipedia topic. These entities will become the main directions for later deep-research tasks, so you should balance direct relationship to the core topic, aspect coverage, and later composability.

Return ONLY JSON that follows the required schema. The JSON object MUST contain exactly two top-level keys: `selected_candidate_ids` and `selected_neighbors`.

Rules:
1. Candidate scope and count: choose entities only from the provided `candidate_neighbors`. The selection scope is the full-article candidate pool, including both lead-section and body-section candidates. If there are at least `target_neighbor_count` available unique candidates, return `target_neighbor_count` unique candidates. Return fewer only when the candidate list itself contains fewer available unique candidates than the target.
2. Selection objective: choose candidates that best support future root-centered deep-research tasks. Value here means research usefulness, aspect coverage, and composability for broad knowledge-coverage evaluation.
3. Coverage, combination, and duplicate filtering: prefer candidates that cover different research aspects of the root, and avoid concentrating on one narrow direction. Exclude highly similar titles, aliases, repeated directions, and obvious-common-knowledge or low-information candidates such as incidental mentions, dates, generic list pages, or weak parenthetical references. Form a differentiated branch set suitable for systematic DeepResearch.
4. Output structure: put the canonical selection in `selected_candidate_ids`. For each id in `selected_candidate_ids`, include exactly one object with the same `candidate_id` in `selected_neighbors`; each selected neighbor object MUST contain only `candidate_id` and `relation_explanation`.
5. Relationship explanation: `relation_explanation` MUST be one concise sentence explaining the relationship between the root and the selected entity, and it MUST be based only on that candidate's `anchor_text` and `context_text`.
\end{AIbox}

\begin{AIbox}{User Prompt}
Select first-hop root expansion entities for the following Wikipedia root.

root_title:
{root_title}

root_summary:
{root_summary}

target_neighbor_count:
{target_neighbor_count}

candidate_neighbors:
{candidate_neighbors}

Return ONLY JSON.
\end{AIbox}

\subsubsection{Non-root Entity Neighbor Selection}
\label{app:prompt-nonroot-neighbor-selection}

This prompt selects bounded outgoing neighbors for expanding a non-root entity.

\begin{AIbox}{System Prompt}
Given a source Wikipedia page and the candidate links from its full article, select the outgoing entities that are best suited as next-layer expansion topics. These entities should be central and directly related to the source topic, and they should provide valuable evidence entry points for the later research scope.

Return ONLY JSON that follows the required schema.

Rules:
1. Select from the full source-article candidate pool, including both lead-section and body-section candidates. You may choose only from the provided `candidate_neighbors`.
2. Choose the candidates that are most semantically central, most directly related to the source entity, and most useful for expanding understanding of it.
3. If the candidate count is at least `target_neighbor_count`, return exactly `target_neighbor_count` unique candidates.
4. Return only `selected_candidate_ids`.
\end{AIbox}

\begin{AIbox}{User Prompt}
Select the best outgoing neighbors for the following source node.

source_title:
{source_title}

source_depth:
{source_depth}

source_summary:
{source_summary}

target_neighbor_count:
{target_neighbor_count}

candidate_neighbors:
{candidate_neighbors}

Return ONLY JSON.
\end{AIbox}

\subsubsection{Subgraph Sampling}
\label{app:prompt-depth1-branch-selection}

This prompt groups the children of the root to select branches that form coherent and complementary subgraphs for downstream task construction.

\begin{AIbox}{System Prompt}
You are selecting and combining candidate directions that are directly connected to a core Wikipedia topic, forming several subtopic sets suitable for deep research. Each set should feel like a natural research scope: its internal directions should be related and complementary, able to support later research-question framing, evidence organization, and evaluation of whether a report covers key knowledge relationships.

Return ONLY JSON that follows the required schema.

Rules:
1. When the candidate pool allows it, construct exactly `target_subgraphs` groups; when the candidate pool allows it, each group should contain exactly `depth_1_branches_per_subgraph` unique candidates. Choose only from the provided depth-1 candidate list.
2. Each group should form a coherent research scope: the selected first-hop branches should have semantic, historical, institutional, causal, comparative, or conceptual connections, and should be able to support a natural deep-research report. Do not simply choose the first candidates. Prefer branches whose descendant tree is more likely to provide effective supporting evidence.
3. Branches within a group should be complementary. Avoid homogeneous branches, near duplicates, or branches that can be represented by the same angle. Unless the seed topic naturally connects them, do not combine unrelated incidental mentions.
4. Different groups should overlap as little as possible. If overlap is unavoidable, prefer reusing candidates that have the most direct relationship to the seed/root, can connect multiple research angles, and whose descendant tree is more likely to provide effective supporting evidence.
5. Use `selected_candidate_ids` as the canonical output.
\end{AIbox}

\begin{AIbox}{User Prompt}
Select depth-1 branch groups for subgraph construction.

seed_entity:
{seed_entity}

root_summary:
{root_summary}

target_subgraphs:
{target_subgraphs}

depth_1_branches_per_subgraph:
{depth_1_branches_per_subgraph}

candidate_count:
{candidate_count}

depth_1_candidates:
{depth_1_candidates}

Return ONLY JSON.
\end{AIbox}

\subsubsection{Research Query Generation}
\label{app:prompt-query-generation-anti-leakage}

This prompt converts the hidden sampled scope into a research query centered on the root entity while preventing entity-list leakage.

\begin{AIbox}{System Prompt}
You convert a set of hidden entity-relation clues into a research question that an outside user might ask.

Your task is to write ONE research-style query centered on the seed/root entity. This query will be given to a downstream DeepResearch model to generate a report, and coverage will be evaluated using rubrics corresponding to the hidden clues. `subgraph_nodes` is hidden construction context: use it only to understand the overall topical composition and depth profile. The final query must still present the root/seed entity as the explicit central topic.

Return ONLY JSON that follows the required schema. The JSON object MUST contain exactly one top-level key: `query`.

Rules:
1. Central topic: the final `query` MUST clearly center on `seed_entity`; the seed/root entity name may appear. The query should use the full hidden context to form a research task, rather than only asking about the root article itself.
2. Anti-leakage: the final `query` MUST NOT directly mention any entity name in `forbidden_entity_names`, any exact title from `depth_1_branch_inputs`, or any exact title from `subgraph_nodes` except the seed/root entity. Also avoid semantic leakage: do not rewrite hidden candidate entities, node titles, summaries, or relation explanations into downstream-facing concrete topics, mechanisms, institutions, places, events, works, methods, people, effects, or checklists. Deeper hidden entities may not all appear in the forbidden list, but the query should still not read like an entity checklist.
3. Writing style: imitate a thoughtful but non-expert user asking for research help. The query should be specific and purposeful, but should not make the user appear to know hidden entities, internal technical terms, hidden graph structure, or specific field methods. Use a natural-language paragraph.
4. Source of specificity: the query should usually be one coherent paragraph, around 100-200 words in English or a similar level of detail in the target language. It must not be too short or too generic, and it must not become specific by enumerating hidden candidate directions. Specificity should come from root-centered research framing, such as causal relationships, historical development, changes in evidence and interpretation, interactions among different forces, or how internal practices and external pressures shaped development.
\end{AIbox}

\begin{AIbox}{User Prompt}
Generate a query from the hidden construction context below.

query_gen_mode:
{query_gen_mode}

graph_metadata:
{graph_metadata}

seed_entity:
{seed_entity}

subgraph_overview:
{subgraph_overview}

root_node:
{root_node}

subgraph_nodes:
{subgraph_nodes}

depth_1_branch_inputs:
{depth_1_branch_inputs}

forbidden_entity_names:
{forbidden_entity_names}

Important:
- The hidden node set is rooted at `seed_entity`.
- The root node is depth 0.
- Each `subgraph_nodes` item contains a node id, Wikipedia title, and root-hop depth.
- A smaller depth means fewer knowledge hops from the root and usually a closer structural relationship to it.
- The final query should be a research task based on the full hidden context while staying centered on the root entity.
- Node titles are hidden construction evidence and must not become a topic checklist in the final query.

Return ONLY JSON.
\end{AIbox}

\subsubsection{Rubric Generation}
\label{app:prompt-edge-rubric-generation}

This prompt converts each edge in a case subgraph into an evaluation rubric based on its associated evidence.

\begin{AIbox}{System Prompt}
You will receive a batch of relationship records, each made from a source entity and a target entity. Each record includes evidence from Wikipedia page summaries and relationship context. Your task is to write one verifiable knowledge-coverage requirement for each record, so that a later research report can be checked for whether it clearly covers the specific knowledge relationship expressed by that record.

For each relationship record, use only the materials provided by that record: source page summary, target page summary, and relation evidence sentence / paragraph. Do not use internal structural information such as node depth, path position, centrality, or subgraph topology.

Return ONLY JSON that follows the required schema.

Rules:
1. Generate exactly one rubric object for each provided `rubric_id`.
2. Each rubric must be faithful to that relationship record's own source/target/evidence materials, and must be written as a knowledge coverage requirement for report content, not as a writing-quality judgment.
3. The rubric must be based on the provided page summaries and relationship evidence. Do not invent claims beyond the summaries and relation evidence. If the evidence is weak, keep the rubric narrow and concrete. If the evidence sentence supports a clear relationship, prefer explicit relation coverage. A report may satisfy the same knowledge content through a clear paraphrase.
4. Use the exact key names from the output schema. Each rubric may return only `rubric_id` and `rubric_text`.
\end{AIbox}

\begin{AIbox}{User Prompt}
For each relationship record in the following list, generate one rubric object based on the provided Wikipedia page evidence.

edge_rubric_inputs:
{edge_rubric_inputs}

Return ONLY JSON.
\end{AIbox}

\subsection{Evaluation Prompt}
\label{app:formal-evaluation-prompts}

\subsubsection{Judge Prompt}
\label{app:prompt-rubric-judge}

\begin{AIbox}{System Prompt}
You need to evaluate whether a report covers a set of predefined scoring points.

The input will include a user research question, several rubric objects, and the report to be evaluated. Each rubric object represents one specific knowledge point to check, and `rubric_text` is the only scoring criterion. Your task is to judge, item by item, whether the report satisfies the corresponding `rubric_text`, and to provide a brief reason for each judgment.

Return only JSON that conforms to the required schema.

Rules:
1. Score each rubric object using only its `rubric_text` and a three-tier scoring scheme: `1` means satisfied, `0.5` means partially satisfied, and `0` means not satisfied. Base your judgment only on what is explicitly present in the report. Do not use external knowledge, common-sense completion, or subjective inference.
2. Assign `1` when the report explicitly states or clearly paraphrases the knowledge required by the rubric, covering the key facts, relationships, qualifiers, or numerical details required by the rubric. Reasonable paraphrases are acceptable, but the rubric must be substantively satisfied.
3. Assign `0.5` when the report mentions the core object, relationship, or fact required by the rubric and covers at least one substantive key element, but misses other key elements, qualifiers, dates, numerical values, causal links, or necessary context. General discussion of a related topic, mere mention of related terms, or content that does not cover any specific knowledge required by the rubric is not enough for `0.5`.
4. Assign `0` when the report does not mention the specific knowledge required by the rubric, only discusses a related topic generally, or makes a factual statement that contradicts the rubric. If the report's core factual claim contradicts the rubric, assign `0` rather than `0.5`, even if related content is mentioned.
5. For every input rubric object, return exactly one object in `rubric_results`, and preserve the original `rubric_id` exactly. Do not omit, merge, rewrite, or add rubrics.
6. Each result must contain only `rubric_id`, `score`, and `reason`. The `reason` should briefly explain why the rubric is satisfied, partially satisfied, or not satisfied, and should indicate which key elements the report covered, which key elements it missed, or which fact contradicts the rubric.
7. Keep each `reason` within 50 words.
\end{AIbox}

\begin{AIbox}{User Prompt}
Please formally score a report against a provided rubric set.

query:
{query}

The report to evaluate is provided in the next message. The rubric_objects to score are provided after the report. When asked to respond, return only JSON. Each rubric result must contain only `rubric_id`, `score`, and `reason`.

# Report To Evaluate
{report_text}

rubric_objects:
{rubric_objects_json}

Now evaluate the report against exactly the rubric object(s) listed above.
Use only the provided rubric ids.
score must be one of:
- 1: The report explicitly states or clearly paraphrases the knowledge required by the rubric and covers the key elements
- 0.5: The report covers some substantive key elements of the rubric, but misses other necessary elements or qualifiers
- 0: The report does not mention the specific knowledge, is only generally related, or factually contradicts the rubric

Each reason must be concise: within 50 words.

Return ONLY JSON.
\end{AIbox}

\end{document}